\documentclass[sigconf,nonacm]{aamas}

\usepackage{balance}
\usepackage{pgfplots}
\usepgfplotslibrary{groupplots}
\usetikzlibrary{arrows.meta, shapes.geometric, positioning, calc}
\usepackage{multirow}
\usepackage{array}

\title{Copies or Sources? Measuring How LLM Aggregators Count Restated Evidence in Multi-Agent Systems}

\setcopyright{none}
\renewcommand\footnotetextcopyrightpermission[1]{}
\authorsaddresses{}
\author{Jianxin Gao}
\affiliation{\institution{China Agricultural University}}
\email{jxgao@cau.edu.cn}
\author{Runze Li}
\affiliation{\institution{Jilin University}}
\email{rzli5524@mails.jlu.edu.cn}
\author{Tianyi Yu}
\affiliation{\institution{Tianjin University of Finance and Economics}}
\email{tiantyu318@stu.tjufe.edu.cn}
\author{Liangwei Ren}
\affiliation{\institution{China Agricultural University}}
\email{lwren@cau.edu.cn}
\author{Bohan Chen}
\affiliation{\institution{China Agricultural University}}
\email{chenbohan@cau.edu.cn}
\author{Zining Wang}
\affiliation{\institution{Tianjin University of Science and Technology}}
\email{25101314@mail.tust.edu.cn}
\renewcommand{\shortauthors}{Jianxin Gao et al.}
\makeatletter
\def\@mkauthors{%
  \global\setbox\mktitle@bx=\vbox{%
    \unvbox\mktitle@bx\par\medskip
    \centering
    {\fontsize{12}{14.5}\selectfont
      Jianxin Gao\textsuperscript{1,*,$\dagger$}\quad
      Runze Li\textsuperscript{2,*}\quad
      Tianyi Yu\textsuperscript{3,*}\quad
      Liangwei Ren\textsuperscript{1}\quad
      Bohan Chen\textsuperscript{1}\quad
      Zining Wang\textsuperscript{4}\par}
    \smallskip
    {\fontsize{10}{12.5}\selectfont
      \textsuperscript{1}China Agricultural University\quad
      \textsuperscript{2}Jilin University\par
      \textsuperscript{3}Tianjin University of Finance and Economics\quad
      \textsuperscript{4}Tianjin University of Science and Technology\par}
    \smallskip
    {\fontsize{9}{11}\selectfont
      \href{mailto:jxgao@cau.edu.cn}{jxgao@cau.edu.cn}\quad
      \href{mailto:rzli5524@mails.jlu.edu.cn}{rzli5524@mails.jlu.edu.cn}\quad
      \href{mailto:tiantyu318@stu.tjufe.edu.cn}{tiantyu318@stu.tjufe.edu.cn}\par
      \href{mailto:lwren@cau.edu.cn}{lwren@cau.edu.cn}\quad
      \href{mailto:chenbohan@cau.edu.cn}{chenbohan@cau.edu.cn}\quad
      \href{mailto:25101314@mail.tust.edu.cn}{25101314@mail.tust.edu.cn}\par
      \textsuperscript{*}Equal contribution.\quad
      \textsuperscript{$\dagger$}Corresponding author.\par}
    \bigskip\kern-3.11065pt}}
\makeatother
\AtBeginDocument{%
  \fancypagestyle{standardpagestyle}{%
    \fancyhf{}\fancyfoot[C]{\footnotesize\thepage}}
  \fancypagestyle{firstpagestyle}{%
    \fancyhf{}\fancyfoot[C]{\footnotesize\thepage}}
  \pagestyle{standardpagestyle}}

\definecolor{cLuna}{HTML}{0072B2}
\definecolor{cSol}{HTML}{D55E00}
\definecolor{cGem}{HTML}{009E73}
\definecolor{cDS}{HTML}{CC79A7}
\definecolor{cRef}{HTML}{7F7F7F}
\definecolor{cOver}{HTML}{E69F00}
\definecolor{cUnder}{HTML}{56B4E9}
\definecolor{cInk}{HTML}{333333}
\definecolor{pBlueL}{HTML}{E4F9FF}\definecolor{pBlue}{HTML}{AECFF9}\definecolor{pBlueD}{HTML}{5B8BC9}
\definecolor{pCream}{HTML}{FEF4D8}\definecolor{pYellow}{HTML}{FFE494}\definecolor{pYellowD}{HTML}{C9A33A}
\definecolor{pPink}{HTML}{FED4D7}\definecolor{pRose}{HTML}{FDB9BF}\definecolor{pPinkD}{HTML}{D9848E}
\definecolor{pPeach}{HTML}{F9DCC8}\definecolor{pOrange}{HTML}{FFB596}\definecolor{pOrangeD}{HTML}{D9774A}
\definecolor{pGreen}{HTML}{E1EFD9}\definecolor{pMauve}{HTML}{F6D5E9}
\pgfplotsset{
  compat=1.18,
  every axis/.append style={font=\scriptsize, tick label style={font=\tiny}, label style={font=\scriptsize},
    legend style={font=\tiny, draw=none, fill=none}, legend cell align=left, axis line style={black!55}, tick style={black!55},
    title style={font=\scriptsize, yshift=-1pt}},
  luna/.style={color=cLuna, mark=*, mark size=1.4pt, line width=0.8pt},
  sol/.style={color=cSol, mark=square*, mark size=1.25pt, line width=0.8pt},
  gem/.style={color=cGem, mark=triangle*, mark size=1.7pt, line width=0.8pt},
  ds/.style={color=cDS, mark=diamond*, mark size=1.7pt, line width=0.8pt},
  refline/.style={color=cRef, dashed, line width=0.5pt},
  bNone/.style={fill=black!12, draw=black!45},
  bCopy/.style={fill=cOver, draw=cOver!70!black},
  bCopyLight/.style={fill=cOver!35, draw=cOver!70!black},
  bOpen/.style={fill=white, draw=cOver!70!black},
  bGrey/.style={fill=black!45, draw=black!65},
  boxlegend/.style={area legend, legend image code/.code={\draw[##1] (0mm,-0.6mm) rectangle (2.4mm,1mm);}},
  ticklegend/.style={legend image code/.code={\draw[black, line width=0.8pt] (0mm,0.2mm) -- (2.4mm,0.2mm);}},
  linelegend/.style={legend image code/.code={\draw[mark repeat=2, mark phase=2, ##1] plot coordinates {(0mm,0mm) (1.6mm,0mm) (3.2mm,0mm)};}},
}
\tikzset{
  msg/.style={draw=black!45, fill=white, rounded corners=0.6pt, inner xsep=1.4pt, inner ysep=0.9pt,
    font=\fontsize{4.6}{5}\selectfont\ttfamily, anchor=west, minimum height=3.1mm},
  msgcopy/.style={msg, fill=cOver!22, draw=cOver!80!black},
  msgref/.style={msg, dash pattern=on 1.1pt off 0.8pt},
  agent/.style={circle, draw=black!60, fill=white, inner sep=0pt, minimum size=4.2mm, font=\fontsize{4.6}{5}\selectfont},
  observer/.style={agent, fill=cInk, text=white, draw=cInk},
  sensor/.style={regular polygon, regular polygon sides=3, draw=cInk, fill=cInk, inner sep=0pt, minimum size=2.1mm},
  flow/.style={-{Stealth[length=1.3mm,width=1mm]}, black!55, line width=0.45pt},
  doc/.style={draw=black!50, fill=white, minimum width=6.2mm, minimum height=8mm, inner sep=0.5pt,
    font=\fontsize{5}{5.5}\selectfont, align=center},
  paneltitle/.style={font=\scriptsize\bfseries, anchor=west, inner sep=0pt},
  note/.style={font=\fontsize{5}{6}\selectfont, text=black!70, align=left, inner sep=0pt},
}
\newcommand{\cond}[1]{\textsc{#1}}
\DeclareMathOperator{\logit}{logit}

\begin{abstract}
Multi-agent systems built on large language models (LLMs) restate observations as a matter of course: relays forward them, shared boards repeat them and discussion rounds echo them. An aggregator that pools such messages should count sources, not statements. We convert a reported probability into units of independent readings, which assigns every restatement a copy weight, 0 for an aggregator that counts sources and 1 for one that counts every statement, and yields the implied decision under any cost structure. Three testbeds hold the evidence fixed and vary how it is restated: message logs with an exact Bayesian oracle, web documents with appended copies, and logs written by LLM agent teams under four communication protocols. Across four models from three providers, a forwarded copy counts for 0.06 to 0.42 of a new reading, mostly because some replies count every statement. On 5\% to 40\% of logs that state one reading three times, the reported belief implies an early commitment that the oracle never makes. The models that count copies least and most on controlled logs do so on web copies and agent-written logs as well. A one-paragraph declaration of what a copy contributes brings the copy weight on controlled logs to 0.08 or less. A rule that has agents refer to readings instead of restating them cuts belief-implied early commitment from 11.2\% to 1.1\% and preserves genuine corroboration.
\end{abstract}


\let\ArxivMainItemize\itemize
\let\ArxivMainEndItemize\enditemize
\let\ArxivMainEnumerate\enumerate
\let\ArxivMainEndEnumerate\endenumerate
\let\ArxivMainDescription\description
\let\ArxivMainEndDescription\enddescription
\let\ArxivMainTrivlist\trivlist
\let\ArxivMainEndTrivlist\endtrivlist
\let\ArxivMainUrlBreaks\UrlBreaks
\let\ArxivMainUrlSpecials\UrlSpecials
\newmuskip\ArxivMainUrlMuskip
\ArxivMainUrlMuskip=\Urlmuskip
\usepackage{tabularx,longtable,fvextra,needspace,float,enumitem,xurl,adjustbox,bookmark,placeins}
\let\ArxivAppendixItemize\itemize
\let\ArxivAppendixEndItemize\enditemize
\let\ArxivAppendixEnumerate\enumerate
\let\ArxivAppendixEndEnumerate\endenumerate
\let\ArxivAppendixDescription\description
\let\ArxivAppendixEndDescription\enddescription
\let\ArxivAppendixTrivlist\trivlist
\let\ArxivAppendixEndTrivlist\endtrivlist
\let\ArxivAppendixUrlBreaks\UrlBreaks
\let\ArxivAppendixUrlSpecials\UrlSpecials
\newmuskip\ArxivAppendixUrlMuskip
\ArxivAppendixUrlMuskip=\Urlmuskip
\let\itemize\ArxivMainItemize
\let\enditemize\ArxivMainEndItemize
\let\enumerate\ArxivMainEnumerate
\let\endenumerate\ArxivMainEndEnumerate
\let\description\ArxivMainDescription
\let\enddescription\ArxivMainEndDescription
\let\trivlist\ArxivMainTrivlist
\let\endtrivlist\ArxivMainEndTrivlist
\let\UrlBreaks\ArxivMainUrlBreaks
\let\UrlSpecials\ArxivMainUrlSpecials
\Urlmuskip=\ArxivMainUrlMuskip

\newcommand{\arm}[1]{\emph{#1}}
\newcommand{\ph}[1]{\textcolor{blue!55!black}{\{#1\}}}
\DefineVerbatimEnvironment{Pattern}{Verbatim}{frame=none,commandchars=none,xleftmargin=1.2em}
\makeatletter
\AtBeginEnvironment{Verbatim}{\@beginparpenalty=\@M}
\AtBeginEnvironment{Pattern}{\@beginparpenalty=\@M}
\makeatother
\newcommand{\suppplot}[1]{%
  \begin{adjustbox}{max width=\linewidth,max totalheight=0.78\textheight}%
    \includegraphics{figures/#1.pdf}%
  \end{adjustbox}%
}

\makeatletter
\newcommand{\ArxivAppendixFormatting}{%
  \let\itemize\ArxivAppendixItemize
  \let\enditemize\ArxivAppendixEndItemize
  \let\enumerate\ArxivAppendixEnumerate
  \let\endenumerate\ArxivAppendixEndEnumerate
  \let\description\ArxivAppendixDescription
  \let\enddescription\ArxivAppendixEndDescription
  \let\trivlist\ArxivAppendixTrivlist
  \let\endtrivlist\ArxivAppendixEndTrivlist
  \let\UrlBreaks\ArxivAppendixUrlBreaks
  \let\UrlSpecials\ArxivAppendixUrlSpecials
  \Urlmuskip=\ArxivAppendixUrlMuskip
  \setlist{itemsep=1pt,topsep=3pt,parsep=0pt}%
  \renewcommand{\arraystretch}{1.1}%
  \setlength{\LTcapwidth}{\textwidth}%
  \setlength{\emergencystretch}{2em}%
  \raggedbottom
  \fvset{breaklines=true,fontsize=\small,baselinestretch=1,samepage=true,frame=single,framerule=0.4pt,framesep=2mm,
    rulecolor=\color{black!40},breaksymbolleft={\color{black!45}\tiny\ensuremath{\hookrightarrow}},
    commandchars=\\\{\}}%

  \setcounter{table}{0}%
  \setcounter{figure}{0}%
  \renewcommand{\theHtable}{app.\arabic{table}}%
  \renewcommand{\theHfigure}{app.\arabic{figure}}%
}
\makeatother
\hypersetup{pdfauthor={Jianxin Gao, Runze Li, Tianyi Yu, Liangwei Ren, Bohan Chen, Zining Wang}}

\begin{document}
\makeatletter
\let\ArxivAppendixLNMakecol\@LN@makecol
\let\@LN@makecol\@LN@orig@makecol
\makeatother

\theoremstyle{acmdefinition}
\newtheorem{assumption}{Assumption}
\theoremstyle{acmplain}
\newtheorem{observation}{Observation}
\maketitle

\begin{figure}[t]
\centering
\includegraphics[width=\columnwidth]{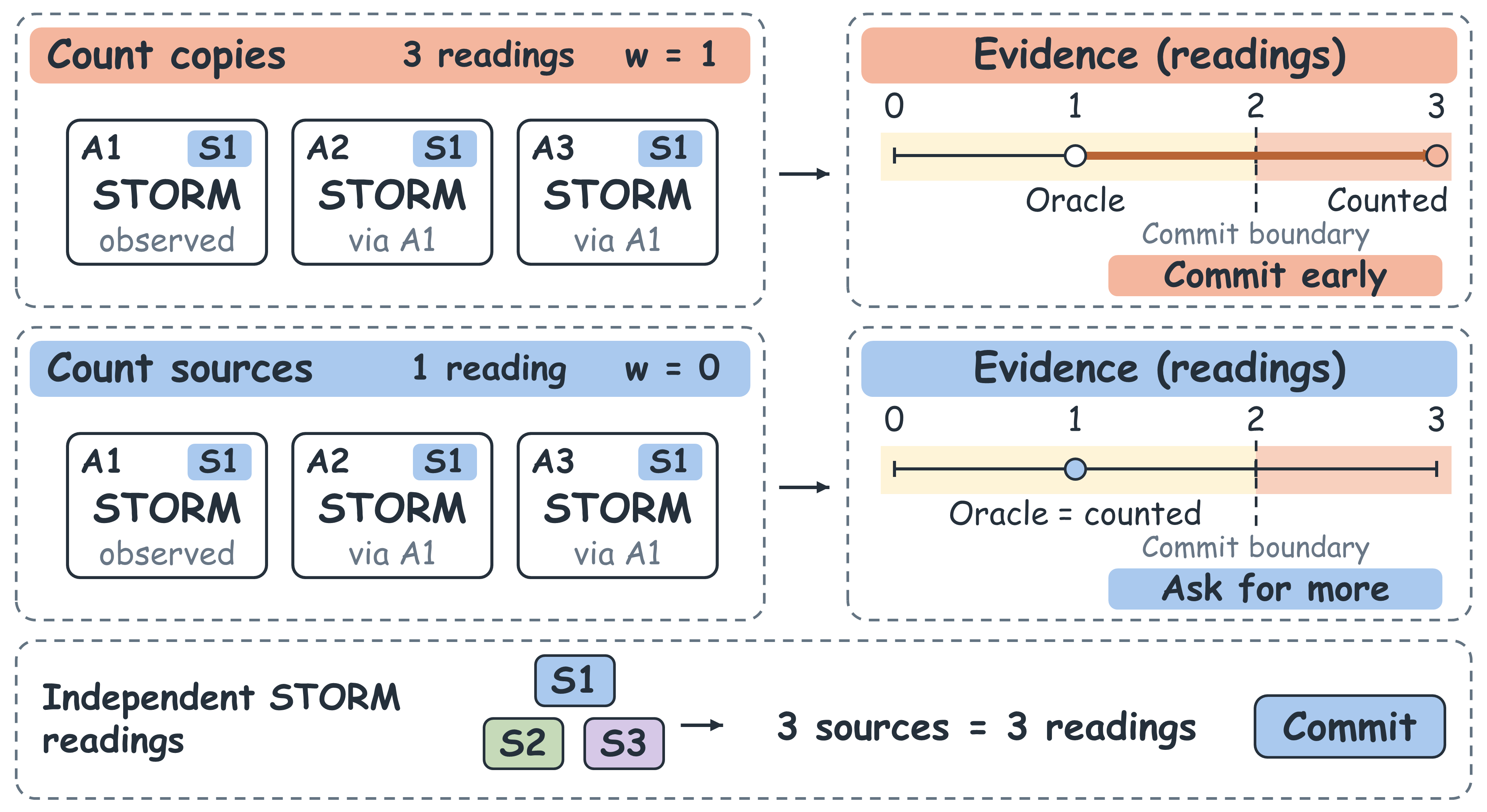}
\caption{One reading relayed twice. Agent A1 observes sensor S1, A2 and A3 pass the report on, and $w$ is the copy weight. Counting copies puts the evidence past the commit boundary, counting sources keeps it at one reading, and three independent sensors S1 to S3 justify committing.}
\label{fig:problem}
\Description{Two rows show the same log of three STORM messages: agent A1 observes sensor S1, and agents A2 and A3 pass A1's report on. In the upper row the aggregator counts copies (w = 1), places the evidence at three readings, beyond the commit boundary at two, and commits early. In the lower row it counts sources (w = 0), places the evidence at one reading, where the oracle also is, and asks for more. A bottom strip shows three independent sensors S1, S2 and S3, which give three readings and justify committing.}
\end{figure}

\section{Introduction}

A rover team must decide whether a dust storm will reach its landing site before it commits to a manoeuvre. Agent A1 reads sensor S1 and reports \textsc{storm}, and agents A2 and A3 pass the report on. The team's log now holds three \textsc{storm} messages but a single measurement (Figure~\ref{fig:problem}). Multi-agent systems built on LLMs produce such logs routinely. Relay chains forward observations, blackboards and shared memories rewrite them \cite{nii1986blackboard,park2023generative}, debate and round-table protocols circulate them before a decision \cite{du2024debate,chen2024reconcile}, hierarchical teams summarise them upward \cite{hong2024metagpt,li2023camel}, and retrieval returns near-duplicates of a single document \cite{cho2026rare}. Whatever component pools these messages, whether a judge, a planner, a router or the final answerer, has to decide how much the restatements count.

Probability theory gives an unambiguous answer. A message that only repeats an earlier observation adds no evidence, so the posterior depends on the set of distinct sources and not on how often each is restated \cite{genest1986combining,winkler1981combining,dietrich2013independent}. Distributed estimation calls the violation of this rule \emph{data incest} \cite{krishnamurthy2014incest}, and people commit it too \cite{enke2019correlation}. Classical aggregators avoid it because the dependence between sources is known, estimated, or enforced by the protocol. An LLM aggregator receives text instead. Provenance is written in plain view, in source fields, relay names and attribution lines, but nothing in the pipeline acts on it unless the model does. Repeated content is known to sway LLM answers \cite{xie2024chameleon,kurfali2025needles,schuster2026whose}. In a multi-agent system, three questions decide how much such sway matters: how much a restatement counts relative to an independent observation, whether that weight changes decisions, and which part of the system, the model or the protocol, controls it.

\begin{figure*}[t]
\centering
\includegraphics[width=\textwidth]{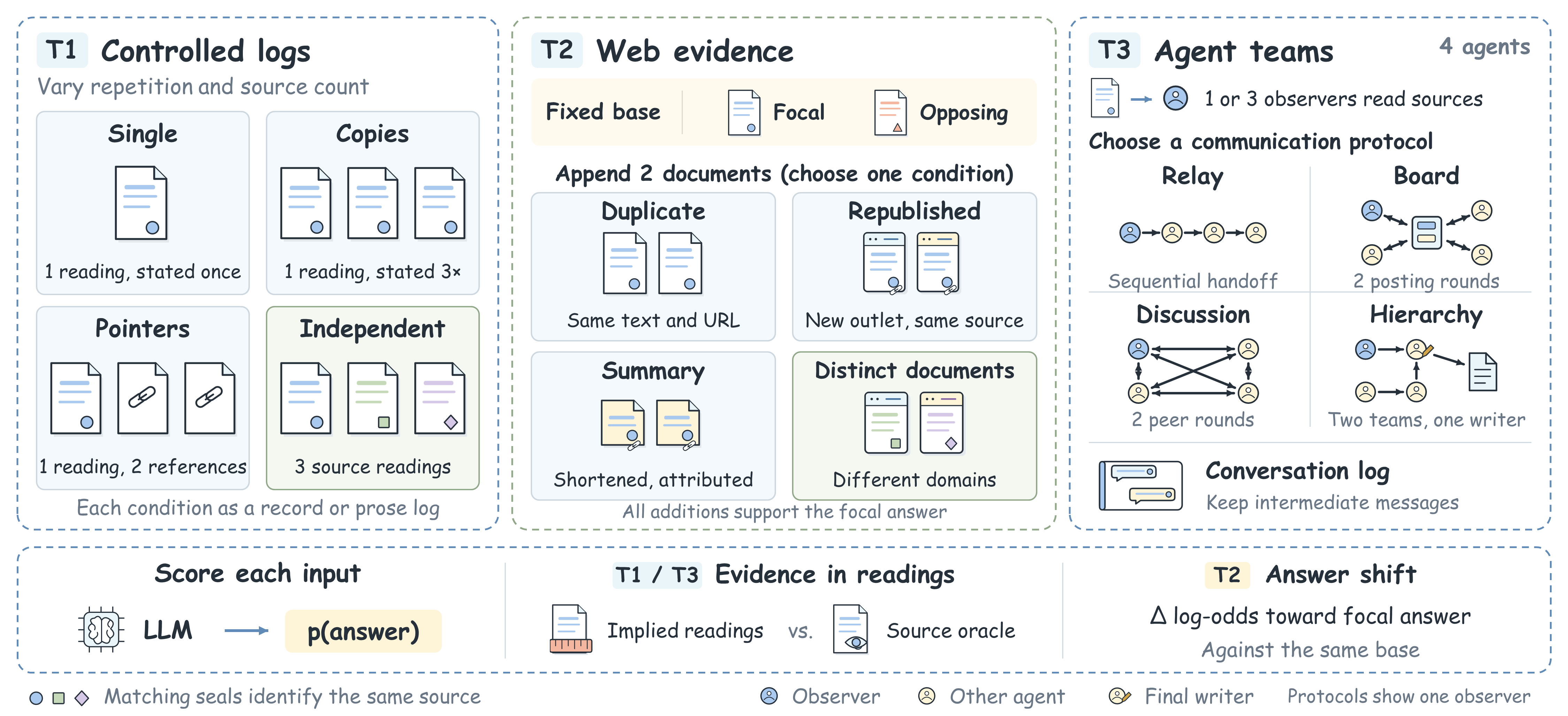}
\caption{The three testbeds T1 to T3. Each restates one source in several ways and compares it with independent sources. Every reply is a probability, which we convert into evidence in readings and compare with the source-counting oracle (T1, T3), or into a shift in log-odds toward the focal answer (T2).}
\label{fig:testbeds}
\Description{Four panels. T1, controlled logs: one reading stated once (Single), stated three times (Copies), stated once with two references (Pointers), or three readings from three sources (Independent), each rendered as records or prose. T2, web evidence: a fixed base of a focal and an opposing document, to which two documents are appended, either exact duplicates, republished copies, or shortened attributed summaries of the focal document, or two distinct documents from other domains. T3, agent teams: four agents with one or three observers communicate under a relay chain, a board with two posting rounds, a discussion with two peer rounds, or a hierarchy of two teams with one final writer, and the conversation log is kept. A bottom strip shows that each input is scored by an LLM that reports a probability, which is converted to evidence in readings and compared with the source-counting oracle for T1 and T3, or to a shift in log-odds toward the focal answer against the same base for T2.}
\end{figure*}

We answer them with a measure in units of independent readings. A reported probability is converted into the number of independent readings whose posterior it equals, so one reading stated once counts as one and three independent readings count as three. A restatement then receives a \emph{copy weight} $w$. An aggregator that counts sources gives copies $w=0$, and one that takes every statement for a new reading gives them $w=1$. Because the measure is read from the belief, before any threshold is applied, one set of reports gives the implied decision for every cost structure. The measure also separates two quantities. The copy weight belongs to the model, whereas the \emph{restatement count}, the number of times the log states a reading, is fixed by the protocol before any model is called. Under one assumption, which our controlled logs support, the protocol affects the aggregator's decisions only through this count, so a protocol that states each reading once leaves the copy weight nothing to act on (Section~\ref{sec:method}). We measure both quantities in three testbeds that hold the evidence fixed and vary only how it is restated (Figure~\ref{fig:testbeds}): controlled message logs with an exact Bayesian oracle (T1), web documents with appended duplicates, republished copies or summaries (T2), and logs written by LLM agent teams under four communication protocols (T3). The first gives exact answers, the second copies of real web documents, and the third the restatements that agents produce without being asked. Four models from three providers answer every condition on the same items. Our contributions are as follows.
\begin{itemize}
\item \textbf{A measure and a decomposition.} The copy weight is read against an exact posterior without fixing a decision threshold, and the decomposition predicts how the model and the protocol combine (Section~\ref{sec:method}).
\item \textbf{A diagnosis in three testbeds.} Every model counts a forwarded copy as part of a reading. The weight follows how often a value is written, whatever the number of records that carry it, and it arises mostly from replies that count every statement. These replies are frequent enough to push beliefs past a commit boundary on logs that hold a single reading. Agent teams restate close to the bound that their protocol allows, and the models that count copies least and most stay the same across the three testbeds (Sections~\ref{sec:setup} and~\ref{sec:diagnosis}).
\item \textbf{Two places to merge copies.} At the aggregator, a declaration of what a copy contributes, or an extraction step that lists the distinct readings, removes most of the weight. In the protocol, a reference rule removes the restatements themselves and brings agent-written logs close to one reading under all four models. Merging at either place keeps the evidence of three genuine observers near three readings (Section~\ref{sec:merging}).
\end{itemize}
Section~\ref{sec:related} relates these results to prior work.

\section{Measuring How Much a Copy Counts}\label{sec:method}

\subsection{Aggregation with restated evidence}
A hidden state $Y\in\{+1,-1\}$ has prior $1/2$. A sensor $s$ takes one reading $x_s\in\{+1,-1\}$ with $P(x_s=Y)=q_s$, and distinct sensors are conditionally independent given $Y$. Agents exchange messages, and an aggregator receives the resulting log $\mathcal M$. A message \emph{restates} a reading if the reading's value can be recovered from the message, either because the value appears in it or because the message asserts agreement with an earlier one. A message that only points to an earlier message does not restate it. The restatement count $\mu(o)$ of reading $o$ is the number of times $\mathcal M$ states it, and a message that writes the value more than once, such as a record with a relay trace, states it more than once. We call every statement of a reading after its first a \emph{copy}. With $\mathcal O$ the set of distinct readings in $\mathcal M$ and $L_o=\log\frac{q_o}{1-q_o}$ the log-likelihood ratio of reading $o$, the posterior is
\begin{equation}\label{eq:oracle}
p^\ast=P(Y{=}{+}1\mid\mathcal M)=\sigma\Bigl(\textstyle\sum_{o\in\mathcal O}x_oL_o\Bigr),
\end{equation}
which does not depend on $\mu$. We call this the \emph{source-counting norm} and an aggregator that reports $p^\ast$ the \emph{oracle}.

\subsection{Evidence in readings and the copy weight}
When all readings share reliability $q$, with $L=\log\frac{q}{1-q}$, and the primary reading has direction $a$, we express a reported probability $\hat p$ in readings:
\begin{equation}\label{eq:ehat}
\hat e=\frac{a\,\logit\hat p}{L}, \qquad e^\ast=\frac{a\,\logit p^\ast}{L}=a\textstyle\sum_{o\in\mathcal O}x_o .
\end{equation}
One reading gives $e^\ast=1$ however often it is restated, and $m$ agreeing readings give $e^\ast=m$. For a log that states one reading $m$ times, the copy weight is
\begin{equation}\label{eq:w}
w_m=\frac{\bar e-1}{m-1},
\end{equation}
where $\bar e$ is the mean of $\hat e$ over items. Counting sources gives $w_m=0$, and counting statements gives $w_m=1$. For example, if the sensor in Figure~\ref{fig:problem} has $q=0.70$ ($L=0.85$) and a model reports $\hat p=0.78$ for \textsc{storm}, then $\hat e=\logit(0.78)/0.85=1.49$ readings, and each of the two copies has received a weight of $0.25$. The \emph{discrimination} $D_m=(\bar e_{\mathrm{ind},m}-\bar e_{\mathrm{copy},m})/(m-1)$ compares $m$ independent readings with one reading stated $m$ times in the same format. The oracle has $D_m=1$, and a model that cannot tell the two apart has $D_m=0$.

Accuracy alone would miss copy counting. On a log whose readings agree, counting sources and counting statements choose the same state and differ only in how sure they are, which is what $\hat e$ records. The construction also carries over to documents without an exact posterior. There we divide the shift in log-odds that a copy causes by the shift that a distinct document causes (Section~\ref{sec:setup}).

\subsection{From beliefs to decisions}\label{sec:boundary}
An aggregator either commits to a state, earning $+1$ if right and $-1$ if wrong, or pays $c$ for one more reading of reliability $r$. At posterior $p$, committing is worth $|2p-1|$. For $p\in[1-r,r]$, querying is worth $2r-1-c$, since the aggregator then follows the new reading. The oracle therefore commits when $|2p-1|>2r-1-c$ and queries otherwise, and in readings the commit boundary is
\begin{equation}\label{eq:boundary}
e_b=\logit(r-c/2)/L .
\end{equation}
On a log whose readings all agree, an aggregator that acts on its reported belief commits to the supported state before the evidence justifies it exactly when $e^\ast<e_b<\hat e$. The share of such reports defines the belief-implied rate of \emph{early commitment}, or early commitment for short. A different cost structure only moves $e_b$, so the same reports give the rate for every cost structure. In our experiments $r$ is set so that $e_b=2$ up to rounding: one reading, however often it is stated, calls for a query, and three agreeing independent readings call for a commitment.

\subsection{Separating the model from the protocol}\label{sec:decomp}
How an aggregator responds to copies depends on the model, and how many copies a log contains depends on the protocol. Let
\begin{equation}\label{eq:excess}
S=\textstyle\sum_{o\in\mathcal O}\bigl(\mu(o)-1\bigr)
\end{equation}
be the number of copies in a log and $Z=\hat e-e^\ast$ the aggregator's excess evidence.

\begin{assumption}[Copy-count sufficiency]\label{ass:count}
Fix a model, a prompt and a form of copy, and consider logs that carry the same agreeing readings. The distribution of $Z$ depends on the log only through $S$. We write $F_S$ for its distribution function.
\end{assumption}

\begin{observation}[Protocol inflation]\label{obs:protocol}
Under Assumption~\ref{ass:count}, on a log where the oracle queries ($e^\ast<e_b$), the aggregator commits early with probability $1-F_S(e_b-e^\ast)$, which depends on the protocol only through $S$. A protocol that states every reading once has $S=0$, and every model then commits early with the probability $1-F_0(e_b-e^\ast)$ that it has on a log without copies, whatever its copy weight.
\end{observation}

The copy weight summarises how $F_S$ moves with $S$: $w_m$ is the mean of $Z$ at $S=m-1$ divided by $m-1$, the average weight of the first $m-1$ copies. A constant discount, $Z=wS+\varepsilon$ with noise $\varepsilon$ that does not depend on the log, gives $w_m=w$ at every $m$. A mixture in which each report counts either sources or every statement, which we call full counting, gives $w_m$ equal to the share of reports that count fully, and the models we test are closer to this second case (Section~\ref{sec:modes}).

Two testable predictions follow. First, records that point to a reading or pad the log without stating a value should leave $\hat e$ at the single-report level, and the weight should follow the number of statements, whatever the number of records (Section~\ref{sec:values}). Second, agent teams that state each reading once should bring every model to its single-report evidence, and a protocol that still restates should keep the evidence inflated (Section~\ref{sec:protocol}). The protocol bounds $\mu$ before any model is called. For a single observer whose reading every later message restates, a relay chain with $d$ relays gives $\mu\le d+1$, a board on which the observer posts once and $n$ agents then post in each of $k$ rounds gives $\mu\le 1+nk$, an all-to-all discussion among $n$ agents over $k$ rounds gives $\mu\le nk$, and a summary hierarchy with $h$ levels gives $\mu\le h+1$. In distributed estimation, the network topology determines how often a measurement returns to a node, and data-incest correction removes the repeats once the topology is known \cite{krishnamurthy2014incest}. Here the protocol plays the role of the topology.

\section{Testbeds and Models}\label{sec:setup}

\paragraph{T1: controlled logs.} T1 is a sensing task with an exact oracle. The system text states the world model: a uniform prior, one reading per sensor, stated reliabilities and independent errors. The user text is a message log followed by the question. The 1{,}000 items cross five domains, five reliabilities $q\in\{0.60,0.65,\dots,0.80\}$, two polarities and 20 instances. Records are one-line JSON objects with a sender, a \texttt{forwarded\_from} label, the sensor, its reliability, a reading identifier, a time and the value, such as \texttt{\{"sender": "Marten", "forwarded\_from": "r1", "sensor\_id": "OD-34", \dots, "value": "STORM"\}} for a forwarded copy. Each item is rendered in 27 forms, and Table~\ref{tab:t1} reports the main ones. \cond{Single} holds the original report alone. \cond{Copies} adds two forwarded copies, \cond{Pointers} two records that only point to the original (value \texttt{"see r1"}), and \cond{Padding} two status records with the token length of two copies. \cond{Independent} gives three sensors one original report each. Dose series state one reading $m\in\{1,2,3,5,8\}$ times or give $m$ independent readings, and two conflict probes set \cond{Copies} against one opposing reading (\cond{Copied tie}, $e^\ast=0$) or two (\cond{Copied minority}, $e^\ast=-1$). On a 500-item subset we add one paragraph to the system text (Section~\ref{sec:aggregator}), vary reasoning effort, and show the costs ($c=0.05$, with $r$ set per $q$ so that $e_b=2$) to ask the model to commit or query.

\begin{table*}[t]
\caption{Reported evidence on controlled logs (T1), as mean $\hat e$ in readings. Headers give the source-counting oracle $e^\ast$. Upper part: the 1{,}000 items in different forms. \textsc{Packed} is one record whose relay trace states the value three times, \textsc{Claims} adds two records that claim to match the original without stating its value, and \textsc{No labels} and \textsc{Prose} change how the records of \textsc{Copies} are written. Lower part: \textsc{Copies} on the 500-item subset, with one paragraph added to the system text or at another reasoning effort (Sol and Gemini cannot switch reasoning off).}
\label{tab:t1}
\small
\newcommand{\hd}[1]{\begin{tabular}[b]{@{}c@{}}#1\end{tabular}}
\renewcommand{\arraystretch}{1.1}
\setlength{\tabcolsep}{4pt}
\begin{tabular*}{\textwidth}{@{}l<{\hspace{8pt}}@{\extracolsep{\fill}}*{11}{c}@{}}
\toprule
& \multicolumn{3}{c}{One reading, stated once ($e^\ast{=}1$)} & \multicolumn{5}{c}{One reading, stated three times ($e^\ast{=}1$)} & \multicolumn{3}{c}{Other readings present} \\
\cmidrule(lr){2-4}\cmidrule(lr){5-9}\cmidrule(l){10-12}
& Single & Pointers & Padding & Copies & Packed & Claims & No labels & Prose & \hd{Independent\\($e^\ast{=}3$)} & \hd{Copied tie\\($e^\ast{=}0$)} & \hd{Copied minority\\($e^\ast{=}{-}1$)} \\
\midrule
GPT-6 Luna & 1.00 & 1.00 & 1.00 & 1.55 & 1.66 & 1.07 & 1.69 & 1.31 & 2.83 & 0.50 & $-$0.52 \\
GPT-6.1 Sol & 1.00 & 1.00 & 1.00 & 1.12 & 1.10 & 1.01 & 1.20 & 1.05 & 2.96 & 0.11 & $-$0.90 \\
Gemini 3.8 Flash & 1.00 & 1.00 & 1.00 & 1.42 & 1.36 & 1.04 & 1.53 & 1.24 & 2.94 & 0.35 & $-$0.64 \\
DeepSeek V4.1 Flash & 0.99 & 1.01 & 1.00 & 1.84 & 1.75 & 1.54 & 1.97 & 1.55 & 2.91 & 0.86 & $-$0.13 \\
\midrule
\end{tabular*}

\vspace{1pt}
\begin{tabular*}{\textwidth}{@{}l<{\hspace{8pt}}@{\extracolsep{\fill}}*{9}{c}@{}}
& \multicolumn{9}{c}{\textsc{Copies} ($e^\ast{=}1$) with one added paragraph, or at another reasoning effort} \\
\cmidrule(l){2-10}
& & & \multicolumn{2}{c}{Declaration} & \multicolumn{3}{c}{Placebo} & \multicolumn{2}{c}{Reasoning} \\
\cmidrule(lr){4-5}\cmidrule(lr){6-8}\cmidrule(l){9-10}
& No note & \hd{Counting\\instruction} & records & relays & unrelated & delivery & display & off & high \\
\midrule
GPT-6 Luna & 1.61 & 1.15 & 1.01 & 1.04 & 1.56 & 1.74 & 1.89 & 1.81 & 1.16 \\
GPT-6.1 Sol & 1.15 & 1.03 & 1.00 & 1.01 & 1.12 & 1.15 & 1.14 & -- & 1.03 \\
Gemini 3.8 Flash & 1.46 & 1.10 & 1.01 & 1.02 & 1.41 & 1.64 & 1.85 & -- & 1.12 \\
DeepSeek V4.1 Flash & 1.81 & 1.36 & 1.15 & 1.17 & 1.82 & 1.94 & 2.08 & 2.10 & 1.39 \\
\bottomrule
\end{tabular*}
\end{table*}

\paragraph{T2: web evidence.} From ConflictingQA \cite{wan2024evidence} we build 500 items, each with a focal document $F$, a document $O$ for the other answer, and two further documents $D_1,D_2$ for the focal answer, all from distinct domains. The base shows $F$ and $O$, and each condition appends two copies of $F$ or the two distinct documents $D_1$ and $D_2$. The copies are exact duplicates, versions republished by other outlets with attribution, or LLM summaries attributed to $F$'s outlet and checked for stance. Every condition is shown with and without a declaration that copies are not independent sources. Real documents have no exact posterior, so we measure the shift toward the focal answer, $\Delta=a(\logit\hat p_c-\logit\hat p_{\mathrm{base}})$. Under the source-counting norm a copy should not move the answer at all. The ratio $\rho=\Delta_{\mathrm{copy}}/\Delta_{\mathrm{distinct}}$ expresses a copy in units of a distinct document from another domain.

\paragraph{T3: agent teams.} For 500 world configurations, teams of four LLM agents run four protocols: a relay chain, a two-round board, a two-round all-to-all discussion, and a two-team hierarchy with a final summary. With one observer, agent A1 holds the only sensor ($e^\ast=1$). With three observers, A1 to A3 each hold one ($e^\ast=3$). Agents know only their own role and are neither asked nor forbidden to restate. Every protocol also runs under the reference rule of Section~\ref{sec:protocol}, which gives 4{,}000 logs without the rule and 4{,}000 with it. The aggregator sees the roster and the full log, so a reference to an earlier message can always be resolved. It answers with no added note, with a declaration, or by first listing the distinct readings and answering from that list (Section~\ref{sec:aggregator}), and answering from the true list of readings gives a reference value. An annotation pass by GPT-6.1 Sol at high reasoning effort counts the messages that state each reading, which gives $\mu$ because agents almost never state a value twice in one message. On 240 messages labelled by two of the authors, the annotation matches the human label on 98.8\%. The workers are GPT-6 Luna without reasoning, and a second worker model writes a further 800 logs.

\paragraph{Models and analysis.} We query GPT-6 Luna, GPT-6.1 Sol, Gemini 3.8 Flash and DeepSeek V4.1 Flash (Luna, Sol, Gemini and DeepSeek below) through their official APIs at low reasoning effort, the lowest level all four support, without setting sampling parameters. Each reply ends with a reported probability. Every model answers every condition on the same items, so each contrast compares two versions of one item that differ in a single factor, as in behavioural testing, contrast sets and paired evaluations \cite{ribeiro2020checklist,gardner2020contrast,gao2026mirrorcraft}. Pairing also removes item difficulty from the variance of each contrast \cite{miller2024errorbars}. Tests are Wilcoxon signed-rank tests on $\hat e$, exact McNemar tests on actions and Page's trend test for dose \cite{page1963ordered}, with Holm correction \cite{holm1979} within families fixed before data collection. Every pre-specified contrast behind the claims below is significant at the 5\% level after correction, and every control that adds no restated value, such as \cond{Pointers} and \cond{Padding}, is equivalent to the single report within 0.10 readings. The supplementary material gives the prompts, the item generation, all 27 forms, and intervals and tests for every number.

\paragraph{Validity of the unit.} On a single report the four models give $\bar e=0.99$ to $1.00$, the one-reading posterior, and on three independent readings $2.83$ to $2.96$ (Table~\ref{tab:t1}). Over-counting is therefore measured on a scale whose endpoints the models themselves reproduce. Two further replies to each of 500 logs move the condition means by at most 0.07 readings.

\section{Copy Counting in Three Testbeds}\label{sec:diagnosis}

\subsection{Every model counts a copy as part of a reading}\label{sec:t1}
On controlled logs, two forwarded copies raise the reported evidence from one reading to 1.12 (Sol), 1.42 (Gemini), 1.55 (Luna) and 1.84 (DeepSeek), although their provenance labels mark both records as copies (Table~\ref{tab:t1}). The copy weights are $w_3=0.06$, $0.21$, $0.27$ and $0.42$, so no model counts sources and none counts every statement. Three independent readings in the same format give 2.83 to 2.96, and the discrimination $D_3$ between copies and sources ranges from 0.53 (DeepSeek) to 0.92 (Sol). For every model the weight is positive at every reliability and in every domain, with a smallest value of 0.04 (Sol at $q=0.80$). More statements raise the evidence further, but beyond three each copy adds less (Figure~\ref{fig:decisions}a). At $m=8$ the models report 1.30 to 2.53 readings, an average weight of $w_8=0.04$ to $0.22$ per copy.

The conflict probes fix the oracle without any appeal to calibration. When one reading stated three times meets one opposing reading (\cond{Copied tie}, $e^\ast=0$), the models lean toward the copied value by 0.11 to 0.86 readings, whereas a plain tie between two sensors gives 0.00. When it meets two opposing readings (\cond{Copied minority}, $e^\ast=-1$), 4\% (Sol) to 39\% (DeepSeek) of the reports side with the copied reading. Three statements of one sensor then outvote two sensors.

\subsection{The weight follows restated values}\label{sec:values}
The controls confirm the first prediction of Section~\ref{sec:decomp}. Records that point to the original (\cond{Pointers}) or pad the log to the length of two copies (\cond{Padding}) leave $\hat e$ at one reading, and so does padding matched to eight copies. Neither the number of records nor the length of the log adds evidence, and references carry no weight. A single record whose relay trace repeats the value (\cond{Packed}) gives 1.10 to 1.75 readings, close to the 1.12 to 1.84 of three separate records, so the number of senders matters little. Records that claim to match the original without stating its value (\cond{Claims}) carry little weight for three models (1.01 to 1.07) and half a reading for DeepSeek (1.54). The form of the log modulates the weight without removing it. Dropping the provenance labels raises $\hat e$ on \cond{Copies} to 1.20 to 1.97, and writing the same records as prose lowers it to 1.05 to 1.55. Format and labels matter only for copies, since three independent readings give 2.82 to 2.96 readings in every format, with or without labels. The evidence rises with the number of times a value is written, whoever writes it.

\subsection{Copy counting is mostly all or nothing}\label{sec:modes}
The mean weight hides how it arises. On \cond{Copies}, from 47\% (DeepSeek) to 91\% (Sol) of reports give the one-reading posterior to within 0.15 readings, 4\% to 35\% give the three-reading posterior of full counting, and the rest fall in between. Fully counting reports carry 71\% to 84\% of the excess evidence. The copy weight thus measures mainly how often a model counts fully, the mixture case of Section~\ref{sec:decomp}, and a mean below the commit boundary can hide reports far above it (Section~\ref{sec:decisions}). The log does not fix which way a reply goes. Asked the same log three times, a model counts the copies in some replies but not in others for 16\% to 42\% of logs. With eight statements the reports spread out. Where eight readings can be registered ($q\le0.70$, since reported probabilities are clipped at 0.999), at most 6\% of reports count all eight, and 12\% to 69\% fall between one and eight.

\begin{figure}[t]
\centering
\begin{tikzpicture}
\begin{groupplot}[group style={group size=2 by 2, horizontal sep=12mm, vertical sep=13mm},
  width=0.355\columnwidth, height=28mm, scale only axis,
  ymajorgrids, grid style={black!8},
  xlabel style={yshift=3pt}, ylabel style={yshift=-4pt}]
\nextgroupplot[xmin=0.6, xmax=8.4, xtick={1,2,3,5,8}, xlabel={statements of one reading, $m$},
  ymin=0.9, ymax=2.7, ytick={1,1.5,2,2.5}, ylabel={reported evidence $\hat e$},
  title={(a) belief},
  legend columns=4, legend style={at={(1.18,1.18)}, anchor=south, column sep=3pt}, linelegend]
  \addplot[refline, forget plot] coordinates {(0.6,1) (8.4,1)};
\addplot[luna] coordinates {(1,0.9995) (2,1.2105) (3,1.5475) (5,1.8170) (8,2.0252)};
\addplot[sol] coordinates {(1,1.0000) (2,1.0565) (3,1.1151) (5,1.2017) (8,1.3049)};
\addplot[gem] coordinates {(1,0.9998) (2,1.1815) (3,1.4238) (5,1.7500) (8,1.9272)};
\addplot[ds] coordinates {(1,0.9943) (2,1.3495) (3,1.8426) (5,2.2568) (8,2.5260)};
  \legend{Luna, Sol, Gemini, DeepSeek}
  \node[font=\fontsize{4.8}{5}\selectfont, text=black!55, anchor=south east] at (axis cs:8.35,1) {oracle};
\nextgroupplot[xmin=0.6, xmax=8.4, xtick={1,2,3,5,8}, xlabel={statements of one reading, $m$},
  ymin=0, ymax=50, ytick={0,10,...,50}, ylabel={commits early (\%)}, title={(b) decision at $e_b=2$}]
\addplot[luna] coordinates {(1,0.0000) (2,4.4000) (3,24.6000) (5,24.1200) (8,33.0000)};
\addplot[sol] coordinates {(1,0.0000) (2,0.9000) (3,4.7000) (5,6.1000) (8,8.8000)};
\addplot[gem] coordinates {(1,0.0000) (2,3.2000) (3,19.1000) (5,20.9000) (8,27.0000)};
\addplot[ds] coordinates {(1,0.0000) (2,7.2000) (3,39.8000) (5,35.9000) (8,46.1000)};
\nextgroupplot[xmin=1, xmax=3, xtick={1,1.5,2,2.5,3}, xlabel={commit boundary $e_b$},
  ymin=0, ymax=80, ytick={0,20,...,80}, ylabel={commits early (\%)}, title={(c) every boundary, $m=3$}]
  \addplot[refline, forget plot] coordinates {(2,0) (2,80)};
\addplot[luna, mark repeat=6, mark phase=4] coordinates {(1.05,39.60) (1.10,39.60) (1.15,39.30) (1.20,38.70) (1.25,37.70) (1.30,35.60) (1.35,35.00) (1.40,33.10) (1.45,32.30) (1.50,32.20) (1.55,31.10) (1.60,28.60) (1.65,27.50) (1.70,27.10) (1.75,26.60) (1.80,25.60) (1.85,25.40) (1.90,25.10) (1.95,25.10) (2.00,24.60) (2.05,24.30) (2.10,24.20) (2.15,24.00) (2.20,24.00) (2.25,24.00) (2.30,24.00) (2.35,24.00) (2.40,24.00) (2.45,24.00) (2.50,24.00) (2.55,24.00) (2.60,24.00) (2.65,24.00) (2.70,23.60) (2.75,23.30) (2.80,23.30) (2.85,20.00) (2.90,16.40) (2.95,13.30) (3.00,6.90)};
\addplot[sol, mark repeat=6, mark phase=4] coordinates {(1.05,9.30) (1.10,9.30) (1.15,9.30) (1.20,9.30) (1.25,9.00) (1.30,8.50) (1.35,8.20) (1.40,7.60) (1.45,7.20) (1.50,7.20) (1.55,6.90) (1.60,6.50) (1.65,6.00) (1.70,5.60) (1.75,5.60) (1.80,5.20) (1.85,5.20) (1.90,4.90) (1.95,4.90) (2.00,4.70) (2.05,4.70) (2.10,4.60) (2.15,4.50) (2.20,4.50) (2.25,4.40) (2.30,4.40) (2.35,4.40) (2.40,4.40) (2.45,4.40) (2.50,4.40) (2.55,4.40) (2.60,4.40) (2.65,4.40) (2.70,4.40) (2.75,4.40) (2.80,4.40) (2.85,4.20) (2.90,3.70) (2.95,3.40) (3.00,1.80)};
\addplot[gem, mark repeat=6, mark phase=4] coordinates {(1.05,28.10) (1.10,28.10) (1.15,28.00) (1.20,28.00) (1.25,27.90) (1.30,26.70) (1.35,26.00) (1.40,24.90) (1.45,24.40) (1.50,24.30) (1.55,23.20) (1.60,22.40) (1.65,21.20) (1.70,20.60) (1.75,20.30) (1.80,20.00) (1.85,20.00) (1.90,19.90) (1.95,19.90) (2.00,19.10) (2.05,19.10) (2.10,19.00) (2.15,19.00) (2.20,19.00) (2.25,19.00) (2.30,19.00) (2.35,19.00) (2.40,19.00) (2.45,19.00) (2.50,19.00) (2.55,19.00) (2.60,19.00) (2.65,19.00) (2.70,18.90) (2.75,18.90) (2.80,18.90) (2.85,17.90) (2.90,15.80) (2.95,14.30) (3.00,5.80)};
\addplot[ds, mark repeat=6, mark phase=4] coordinates {(1.05,53.30) (1.10,53.20) (1.15,53.10) (1.20,52.90) (1.25,52.50) (1.30,50.10) (1.35,49.60) (1.40,48.30) (1.45,47.70) (1.50,47.40) (1.55,45.20) (1.60,43.40) (1.65,42.50) (1.70,42.00) (1.75,41.70) (1.80,40.50) (1.85,40.40) (1.90,40.10) (1.95,40.00) (2.00,39.80) (2.05,39.20) (2.10,39.00) (2.15,38.80) (2.20,38.80) (2.25,38.80) (2.30,38.80) (2.35,38.80) (2.40,38.80) (2.45,38.80) (2.50,38.80) (2.55,38.80) (2.60,38.80) (2.65,38.80) (2.70,38.60) (2.75,38.30) (2.80,38.30) (2.85,35.70) (2.90,32.20) (2.95,29.30) (3.00,12.20)};
\nextgroupplot[xmin=1, xmax=3, xtick={1,1.5,2,2.5,3}, xlabel={commit boundary $e_b$},
  ymin=0, ymax=80, ytick={0,20,...,80}, ylabel={commits early (\%)}, title={(d) every boundary, $m=8$}]
  \addplot[refline, forget plot] coordinates {(2,0) (2,80)};
\addplot[luna, mark repeat=6, mark phase=4] coordinates {(1.05,59.00) (1.10,59.00) (1.15,59.00) (1.20,59.00) (1.25,59.00) (1.30,57.60) (1.35,57.60) (1.40,56.40) (1.45,55.20) (1.50,55.00) (1.55,51.80) (1.60,48.60) (1.65,46.30) (1.70,45.20) (1.75,44.00) (1.80,41.90) (1.85,41.90) (1.90,40.40) (1.95,40.20) (2.00,33.00) (2.05,30.40) (2.10,29.70) (2.15,28.10) (2.20,28.10) (2.25,27.50) (2.30,27.20) (2.35,26.90) (2.40,26.50) (2.45,26.20) (2.50,25.60) (2.55,25.10) (2.60,24.60) (2.65,24.60) (2.70,24.20) (2.75,24.00) (2.80,24.00) (2.85,21.60) (2.90,20.00) (2.95,18.60) (3.00,15.60)};
\addplot[sol, mark repeat=6, mark phase=4] coordinates {(1.05,14.20) (1.10,14.20) (1.15,14.20) (1.20,14.20) (1.25,14.20) (1.30,14.10) (1.35,13.90) (1.40,13.60) (1.45,13.40) (1.50,13.40) (1.55,12.80) (1.60,12.40) (1.65,11.60) (1.70,11.10) (1.75,10.70) (1.80,10.30) (1.85,10.20) (1.90,10.10) (1.95,10.00) (2.00,8.80) (2.05,8.20) (2.10,7.90) (2.15,7.50) (2.20,7.50) (2.25,7.30) (2.30,7.20) (2.35,7.20) (2.40,7.10) (2.45,7.00) (2.50,7.00) (2.55,6.90) (2.60,6.90) (2.65,6.90) (2.70,6.80) (2.75,6.80) (2.80,6.80) (2.85,6.50) (2.90,6.20) (2.95,5.90) (3.00,5.00)};
\addplot[gem, mark repeat=6, mark phase=4] coordinates {(1.05,44.60) (1.10,44.60) (1.15,44.60) (1.20,44.60) (1.25,44.60) (1.30,43.50) (1.35,43.20) (1.40,42.00) (1.45,40.70) (1.50,40.60) (1.55,38.60) (1.60,35.70) (1.65,34.10) (1.70,33.80) (1.75,33.20) (1.80,32.10) (1.85,31.60) (1.90,30.80) (1.95,30.40) (2.00,27.00) (2.05,26.20) (2.10,25.60) (2.15,24.50) (2.20,24.50) (2.25,24.10) (2.30,24.00) (2.35,23.90) (2.40,23.80) (2.45,23.60) (2.50,23.20) (2.55,22.60) (2.60,22.40) (2.65,22.40) (2.70,22.00) (2.75,21.40) (2.80,21.40) (2.85,20.10) (2.90,19.30) (2.95,18.60) (3.00,15.00)};
\addplot[ds, mark repeat=6, mark phase=4] coordinates {(1.05,75.80) (1.10,75.80) (1.15,75.80) (1.20,75.80) (1.25,75.80) (1.30,74.70) (1.35,74.20) (1.40,72.10) (1.45,70.20) (1.50,70.00) (1.55,66.80) (1.60,62.50) (1.65,60.30) (1.70,59.20) (1.75,58.30) (1.80,55.70) (1.85,55.10) (1.90,53.60) (1.95,53.50) (2.00,46.10) (2.05,42.80) (2.10,42.10) (2.15,40.00) (2.20,40.00) (2.25,38.70) (2.30,38.30) (2.35,37.70) (2.40,37.20) (2.45,37.10) (2.50,36.40) (2.55,35.20) (2.60,34.60) (2.65,34.60) (2.70,34.20) (2.75,33.90) (2.80,33.90) (2.85,31.60) (2.90,30.30) (2.95,29.00) (3.00,23.60)};
\end{groupplot}
\end{tikzpicture}
\caption{Copies on controlled logs (T1). (a) Reported evidence and (b) the share of logs that would commit early when one reading is stated $m$ times. (c, d) The same share at every commit boundary $e_b$, with the dashed line at $e_b=2$. The oracle stays at one reading and queries on every log.}
\label{fig:decisions}
\Description{Four line charts, one line per model. Top: against the number of statements of one reading (1, 2, 3, 5, 8), reported evidence starts at one reading for every model and rises, steeply up to three statements and more slowly after, to between 1.3 and 2.5 readings at eight statements. The early-commitment rate at a boundary of two readings starts at zero and rises to between 9 and 46 percent. Bottom: against the commit boundary from one to three readings, the early-commitment rate for three statements stays almost flat between two and three readings and then drops near three readings. For eight statements it declines steadily.}
\end{figure}
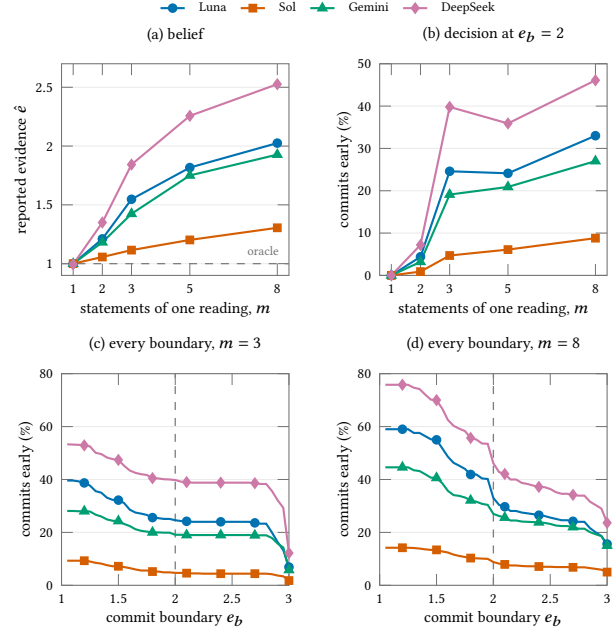

\subsection{Copies tip decisions}\label{sec:decisions}
Over-counting matters when it moves a decision. With the commit boundary at two readings, the oracle never commits on a log that holds one reading, no matter how many messages state it. The models do (Figure~\ref{fig:decisions}b). One reading stated three times crosses the boundary in 4.7\% (Sol), 19.1\% (Gemini), 24.6\% (Luna) and 39.8\% (DeepSeek) of logs, and eight statements in 8.8\% to 46.1\%, while the single report, \cond{Pointers} and \cond{Padding} cross it in at most 0.3\%. The mean evidence on \cond{Copies} stays below the boundary for every model (1.12 to 1.84), so an aggregator that gave every copy the same partial weight would not cross it at all. Most crossings come from full counting, and resampling removes few of them. On 500 logs answered three times, the median of the three replies crosses the boundary in 3.4\% to 36.4\% of logs, against 6.4\% to 37.8\% for a single reply, and all three replies cross it in 1.0\% to 18.0\%. Moving the boundary gives the rate under any other cost structure (Figure~\ref{fig:decisions}c, d). With three statements the rate barely changes between two and three readings and drops only near three, the value that full counting reaches. At $e_b=2.5$, 4\% to 39\% of logs still commit early, within a point of the rate at $e_b=2$. With eight statements the rate declines steadily as the boundary rises.

The same holds when the models choose the action themselves. Shown the costs and asked to commit or query, they commit on \cond{Copies} 6 to 35 percentage points more often than on the single report of the same item, a paired difference that removes each model's baseline tendency to commit. For every model this increase matches, to within 6 points, the share of the same logs whose reported belief crosses the boundary, so the beliefs we measure predict how far copies move the chosen action. On three independent readings, where committing is correct, the models commit in 89\% to 99\% of cases.

\begin{table}[t]
\caption{Web documents (T2) and agent-written logs (T3), without and with merging. T2: shift toward the focal document $F$ caused by two distinct documents, in log-odds, and the weight $\rho$ of each type of copy in units of that shift, without and with the declaration. T3: mean $\hat e$ on agent-written logs, where the three-observer rows use logs that contain all three readings. The reference rule changes the agents' prompt, and the other rows change the aggregator.}
\label{tab:t2}
\small
\renewcommand{\arraystretch}{1.05}
\setlength{\tabcolsep}{4pt}
\begin{tabular*}{\columnwidth}{@{}l@{\extracolsep{\fill}}cccc@{}}
\toprule
& Luna & Sol & Gemini & DeepSeek \\
\midrule
\multicolumn{5}{@{}l}{\textbf{T2}: web documents} \\
\multicolumn{5}{@{}l}{\textit{No note}} \\
\quad shift, two distinct documents & 0.71 & 0.59 & 0.82 & 0.93 \\
\quad $\rho$, duplicate & 0.30 & 0.14 & 0.16 & 0.32 \\
\quad $\rho$, republished & 0.45 & 0.26 & 0.31 & 0.49 \\
\quad $\rho$, summary & 0.64 & 0.40 & 0.50 & 0.71 \\
\addlinespace[2pt]
\multicolumn{5}{@{}l}{\textit{Declaration}} \\
\quad shift, two distinct documents & 0.66 & 0.55 & 0.74 & 0.81 \\
\quad $\rho$, duplicate & 0.10 & 0.07 & 0.10 & 0.18 \\
\quad $\rho$, republished & 0.22 & 0.09 & 0.13 & 0.30 \\
\quad $\rho$, summary & 0.37 & 0.19 & 0.25 & 0.47 \\
\midrule
\multicolumn{5}{@{}l}{\textbf{T3}: agent teams, mean $\hat e$} \\
\multicolumn{5}{@{}l}{\textit{One observer ($e^\ast=1$)}} \\
\quad no note & 1.34 & 1.09 & 1.30 & 1.70 \\
\quad declaration & 1.04 & 1.02 & 1.03 & 1.19 \\
\quad extract first & 0.97 & 0.99 & 0.98 & 0.95 \\
\quad reference rule & 1.01 & 0.99 & 1.01 & 1.06 \\
\addlinespace[2pt]
\multicolumn{5}{@{}l}{\textit{Three observers ($e^\ast=3$)}} \\
\quad no note & 3.06 & 3.02 & 3.11 & 3.42 \\
\quad declaration & 2.88 & 2.95 & 2.96 & 3.09 \\
\quad extract first & 2.85 & 2.92 & 2.91 & 2.87 \\
\quad reference rule & 2.88 & 2.96 & 2.97 & 2.99 \\
\quad true list & 2.90 & 2.96 & 2.95 & 2.94 \\
\bottomrule
\end{tabular*}
\end{table}

\subsection{Web copies count too}\label{sec:t2}
With real documents the answer also rests on what a model already knows, so each condition is compared with the same two base documents. Two distinct supporting documents move the answer toward the focal side by 0.59 to 0.93 in log-odds (Table~\ref{tab:t2}). Copies of $F$ move it as well, although none of them adds a source. In units of a distinct document, an exact duplicate counts for $\rho=0.14$ to $0.32$, a republished copy for 0.26 to 0.49, and a summary for 0.40 to 0.71. The further a copy departs from its original, the more it counts, even though republished copies and summaries both name $F$'s outlet.

\begin{figure}[t]
\centering
\begin{tikzpicture}[
  who/.style={anchor=north east, inner sep=0pt, font=\fontsize{5.4}{6.4}\selectfont\sffamily\bfseries, text=black!70},
  m/.style={anchor=north west, draw, rounded corners=0.8pt, text width=0.795\columnwidth, inner xsep=2pt,
    inner ysep=1.1pt, align=left, font=\fontsize{5.4}{6.4}\selectfont\sffamily},
  og/.style={m, fill=pCream, draw=pYellowD},
  st/.style={m, fill=pPeach, draw=pOrangeD},
  rf/.style={m, fill=pBlueL, draw=pBlueD},
  ttl/.style={anchor=south west, inner sep=0pt, font=\scriptsize}]
\def\gap{0.9pt}
\node[ttl] at (0,0.4mm) {(a) no rule: five messages state the reading};
\node[og] (a1) at (0,0) {OD-93 reported CLEAR at 11:01 (rd-9476; reliability 0.75).};
\node[st] (a2) at ([yshift=-\gap]a1.south west) {OD-93 reading rd-9476 at 11:01 was CLEAR; sensor reliability is 0.75.};
\node[st] (a3) at ([yshift=-\gap]a2.south west) {OD-93 reported CLEAR at 11:01 (rd-9476; reliability 0.75).};
\node[st] (a4) at ([yshift=-\gap]a3.south west) {Raven's single CLEAR reading (OD-93, reliability 0.75) is one reading only; I can't verify it independently.};
\node[st] (a5) at ([yshift=-\gap]a4.south west) {Kestrel: Raven reports that sensor OD-93 (reliability 0.75) recorded CLEAR at 11:01 (reading rd-9476).};
\node[ttl] at ([yshift=-3.2mm]a5.south west) {(b) reference rule: one message states it};
\node[og] (b1) at ([yshift=-3.6mm]a5.south west) {Sensor OD-93 (reliability 0.75) recorded CLEAR at 11:01 (reading rd-9476).};
\node[rf] (b2) at ([yshift=-\gap]b1.south west) {See Raven's report on OD-93.};
\node[rf] (b3) at ([yshift=-\gap]b2.south west) {I have no sensor readings of my own. See Raven's report on sensor OD-93.};
\node[rf] (b4) at ([yshift=-\gap]b3.south west) {See Raven's report on sensor OD-93.};
\node[rf] (b5) at ([yshift=-\gap]b4.south west) {See Raven's report on OD-93.};
\foreach \p in {a,b} {
  \node[who] at ([xshift=-1.6pt,yshift=-1.3pt]\p1.north west) {Raven};
  \node[who] at ([xshift=-1.6pt,yshift=-1.3pt]\p2.north west) {Raven};
  \node[who] at ([xshift=-1.6pt,yshift=-1.3pt]\p3.north west) {Puffin};
  \node[who] at ([xshift=-1.6pt,yshift=-1.3pt]\p4.north west) {Lynx};
  \node[who] at ([xshift=-1.6pt,yshift=-1.3pt]\p5.north west) {Kestrel};
}
\end{tikzpicture}
\caption{One board log (T3) written without (a) and with (b) the reference rule. Raven holds the only sensor and reports first. Orange messages restate its reading, and blue messages refer to it.}
\label{fig:transcript}
\Description{Two short message logs from the same board configuration. In the first, Raven reports a CLEAR reading, then Raven, Puffin, Lynx and Kestrel each restate the reading, Lynx while noting that it is one reading only. In the second, written under the reference rule, Raven reports the reading once and the other messages say only to see Raven's report.}
\end{figure}
\begin{figure}[t]
\centering
\begin{tikzpicture}
\begin{groupplot}[group style={group size=2 by 1, horizontal sep=12mm},
  width=0.355\columnwidth, height=28mm, scale only axis,
  xmin=1.7, xmax=6.3, xtick={2,...,6}, xticklabels={2,3,4,5,6+},
  ymajorgrids, grid style={black!8},
  xlabel={statements of A1's reading}, xlabel style={yshift=3pt}, ylabel style={yshift=-4pt}]
\nextgroupplot[ymin=0.9, ymax=2.35, ytick={1,1.5,2}, ylabel={reported evidence $\hat e$},
  title={(a) belief},
  legend columns=4, legend style={at={(1.18,1.18)}, anchor=south, column sep=3pt}, linelegend]
  \addplot[refline, forget plot] coordinates {(1.7,1) (6.3,1)};
\addplot[luna] coordinates {(2,1.0316) (3,1.1650) (4,1.3471) (5,1.4566) (6,1.6146)};
\addplot[sol] coordinates {(2,1.0010) (3,1.0352) (4,1.0761) (5,1.1095) (6,1.2195)};
\addplot[gem] coordinates {(2,1.0385) (3,1.1471) (4,1.2703) (5,1.4598) (6,1.5423)};
\addplot[ds] coordinates {(2,1.0856) (3,1.3698) (4,1.7398) (5,1.9178) (6,2.2437)};
  \legend{Luna, Sol, Gemini, DeepSeek}
  \node[font=\fontsize{4.8}{5}\selectfont, text=black!55, anchor=south east] at (axis cs:6.25,1) {oracle};
\nextgroupplot[ymin=0, ymax=36, ytick={0,10,20,30}, ylabel={commits early (\%)}, title={(b) decision}]
\addplot[luna] coordinates {(2,1.2300) (3,5.5400) (4,11.5900) (5,14.8300) (6,18.8500)};
\addplot[sol] coordinates {(2,0.6200) (3,1.4800) (4,2.6700) (5,3.1100) (6,5.7500)};
\addplot[gem] coordinates {(2,0.6200) (3,5.5400) (4,8.9100) (5,16.2700) (6,15.3400)};
\addplot[ds] coordinates {(2,0.6200) (3,12.9400) (4,24.6400) (5,27.0300) (6,33.8700)};
\end{groupplot}
\end{tikzpicture}
\caption{Agent-written logs (T3), one observer, no note, pooled over protocols. (a) Reported evidence and (b) the share of logs that would commit early, against the number of messages that state the observer's reading.}
\label{fig:t3}
\Description{Two line charts against the number of messages that state the observer's reading, from 2 to 6 or more, one line per model. Reported evidence rises from about one reading for every model, most steeply for DeepSeek, and the early-commitment rate rises from near zero to between 6 and 34 percent.}
\end{figure}
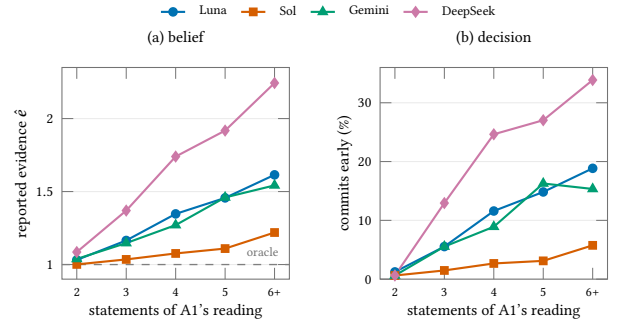
\begin{table}[t]
\caption{Agent-team protocols (T3, one observer) without (free) and with the reference rule: bound on $\mu$ from Section~\ref{sec:decomp}, median restatement count, share of restating messages, and early commitment averaged over the four aggregators.}
\label{tab:protocols}
\small
\setlength{\tabcolsep}{3pt}
\begin{tabular*}{\columnwidth}{@{}l@{\extracolsep{\fill}}ccccccc@{}}
\toprule
& & \multicolumn{2}{c}{Median $\mu$} & \multicolumn{2}{c}{Restating (\%)} & \multicolumn{2}{c}{Commits early (\%)} \\
\cmidrule(lr){3-4}\cmidrule(lr){5-6}\cmidrule(l){7-8}
& Bound & free & rule & free & rule & free & rule \\
\midrule
Relay chain & 4 & 4 & 1 & 99 & 6 & 12.1 & 0.4 \\
Board & 5 & 5 & 1 & 83 & 3 & 13.6 & 0.4 \\
Discussion & 8 & 7 & 1 & 65 & 1 & 13.2 & 0.3 \\
Hierarchy & 3 & 3 & 2 & 49 & 33 & 6.1 & 3.3 \\
\bottomrule
\end{tabular*}
\end{table}

\subsection{Agent teams restate by default}\label{sec:t3}
When agents write the log themselves, restating is their default. Without a rule, 71\% of the messages written by agents that never observed a reading restate it, from 49\% in the hierarchy to 99\% in the relay chain. These agents almost never claim that the reading was verified (0.04\% of messages). In Figure~\ref{fig:transcript}a, Lynx notes that Raven's reading is one reading only and cannot be verified, and repeats its value in the same sentence. The protocol bounds the count, and the agents come close to the bound. The median number of messages that state the observer's reading is 4 in the relay chain, 5 on the board, 7 in the discussion and 3 in the hierarchy, against bounds of 4, 5, 8 and 3 (Table~\ref{tab:protocols}). A second worker model restates as often, and more often in the discussion.

The aggregator turns these restatements into evidence (Figure~\ref{fig:t3}a). On one-observer logs, the reported evidence rises with the restatement count by 0.04 (Sol), 0.11 (Gemini), 0.12 (Luna) and 0.22 (DeepSeek) readings per restatement, estimated within protocol. Pooled over protocols, 11.2\% of one-observer logs would commit early, and for DeepSeek the rate reaches 34\% at a restatement count of six or more (Figure~\ref{fig:t3}b). On logs from the second worker model the slopes range from 0.09 to 0.24, again lowest for Sol and highest for DeepSeek. Relays also lose readings. In 26\% of three-observer relay chains, and in no other protocol, one reading never reaches the log, so we score three-observer logs only when all three readings arrive. On these complete logs restatement still blurs genuine corroboration. The discrimination between three observers and one, which the oracle puts at one reading per additional observer, falls to 0.71 to 0.95 in the relay chain, the protocol that restates most.

\paragraph{One ranking across testbeds.} The three testbeds put the weight of a copy on three scales: $w_3$ on controlled logs, the slope per restatement on agent-written logs, and $\rho$ for each of the three kinds of web copy. What counts as a copy differs, from an exact JSON forward to a news summary, and so do the values. The two ends of the ranking stay fixed. On all five measures Sol counts copies least and DeepSeek most, with Gemini and Luna between them. The values also track the form of the copy. Agents restate in prose, and on controlled logs prose copies with provenance labels carry a weight of 0.03 to 0.27, within 0.06 of each model's slope on agent-written logs.

\section{Merging Copies Before the Decision}\label{sec:merging}

Section~\ref{sec:diagnosis} places the inflation where the aggregator meets restated text. Copies can therefore be merged on either side of that meeting: by the aggregator that reads the log, or by the protocol that writes it.

\subsection{Merging at the aggregator}\label{sec:aggregator}
A \emph{declaration} is a paragraph in the aggregator's system text that states what a copy contributes, and it cuts most of the copy weight. On controlled logs, a declaration that forwarded records add no evidence brings $\hat e$ on \cond{Copies} to 1.00 to 1.01 for three models and to 1.15 for DeepSeek ($w_3\le0.08$, Table~\ref{tab:t1}, lower part), and two further wordings, one about relays and one about copies, have the same effect ($w_3\le0.09$). Unlike a general warning against repetition \cite{schuster2026whose}, the declaration names the records it applies to. A \cond{Packed} record has an empty \texttt{forwarded\_from} field, and on such records the evidence under the same declaration stays at 1.02 to 1.33 readings on the 500-item subset, against 1.09 to 1.78 without a note. The response to independent readings moves by 0.06 readings or less, and the share of reports that side with the copied minority falls to 3\% or less. A counting instruction, which asks the model to count each distinct reading once without saying why, is weaker on its own ($w_3=0.01$ to $0.18$) and brings $w_3$ to 0.03 or less when added to the declaration. Documentation alone does not merge copies. A field guide explaining that \texttt{forwarded\_from} names the record a copy came from leaves the copy weight at 0.04 to 0.37, against 0.08 to 0.40 without it.

Placebos show that the declaration works through its content and not through its length or position. A paragraph of the same length about timestamps and the answer format changes the copy weight by at most 0.03. The placebo that points most directly at the forwarded records, a note on how they are displayed, raises $\hat e$ on \cond{Copies} by 0.27 to 0.39 readings for Luna, Gemini and DeepSeek and leaves Sol unchanged. Drawing attention to copies without a reason to merge them makes them count more. More reasoning also lowers the weight. High effort brings it to 0.01 (Sol) to 0.20 (DeepSeek), and switching reasoning off raises it to 0.40 for Luna and 0.55 for DeepSeek.

The aggregator can also merge copies explicitly. On agent-written logs, extracting the distinct readings first and answering from that list brings one-observer logs to 0.95 to 0.99 readings, with no early commitment in any of the 2{,}000 one-observer logs per model (Table~\ref{tab:t2}). The declaration, which there states that agents without sensors can only repeat readings, brings them to 1.02 to 1.19. On web documents a copy has to be recognised from its content, and the declaration reaches less far. It brings an exact duplicate to $\rho=0.07$ to $0.18$ and a republished copy to 0.09 to 0.30, while summaries, the copies furthest from their original, keep a weight of 0.19 to 0.47. Where copies cannot be recognised in the text, they have to be merged upstream, in the protocol.

\subsection{Merging in the protocol}\label{sec:protocol}
By Observation~\ref{obs:protocol}, once a protocol states each reading only once, the copy weight no longer matters, provided that references carry no weight, as on controlled logs (Section~\ref{sec:values}). The \emph{reference rule} tests this second prediction. It adds a short paragraph to each agent's prompt that asks the agent to state each reading in full once and to refer to readings already in the log without restating them. Logs written under the rule are scored with no note, so the rule changes only the protocol. Under the rule, agents without sensors point to a reading instead of repeating it (Figure~\ref{fig:transcript}b), and relays pass on pointers such as ``see Falcon's report on PM-21'', which the aggregator resolves against the full log. The share of restating messages falls from 71\% to 10\%, and the median restatement count falls to one in the relay chain, on the board and in the discussion (Table~\ref{tab:protocols}). On one-observer logs from these three protocols, all four aggregators then report 0.95 to 1.03 readings, although their copy weights differ sevenfold. Their mean early commitment, 12\% to 14\% without the rule, falls to 0.5\% or less. DeepSeek, which counts copies most, commits early on 23\% to 25\% of these logs without the rule and on 1.4\% or less with it.

The hierarchy confirms the prediction from the other side. Its summaries still restate the reading under the rule (median count 2, 33\% of messages), and the evidence stays inflated, by up to 0.22 readings for DeepSeek, in line with a restatement count above one. Pooled over all four protocols, early commitment falls from 11.2\% to 1.1\%. Under the rule no relay loses a reading, and the discrimination between three observers and one is 0.92 to 0.99 in every protocol.

\subsection{Genuine corroboration survives merging}\label{sec:keep}
A fix that made every log look like one reading would be useless. With three observers, where three readings are genuine, merging keeps the reported evidence close to what the same model reports when given the true list of readings (2.90 to 2.96, Table~\ref{tab:t2}). The gap is within 0.06 readings under the reference rule and 0.07 when extracting first, whose list matches the true readings in 93\% to 97\% of logs, and within 0.02 under the declaration for three models. DeepSeek over-counts genuine corroboration as well, with 3.42 readings without merging and 3.09 under the declaration. On web documents the declaration lowers the shift that distinct documents cause by 0.04 to 0.12 log-odds. When the models choose actions, the declaration brings the extra commitment that copies cause to one point or less for three models and to 11 points for DeepSeek, while commitment on three independent readings stays at 89\% to 99\%.

\paragraph{Choosing where to merge.} The three fixes need different things. A declaration needs copies that the aggregator can recognise in the text, which holds for forwarded records and duplicated documents and holds less for summaries. Extraction costs a second model call per decision, and the answer inherits any reading that the list misses. The reference rule needs control over the agents' prompts and a log in which every reference resolves, and a protocol that asks agents to summarise keeps some restatement under it.

\section{Related Work}\label{sec:related}

\paragraph{Information flow in LLM multi-agent systems.}
Debate and round-table protocols can improve reasoning \cite{du2024debate,chen2024reconcile}, although gains at matched budget are contested \cite{smit2024mad}, majority voting accounts for much of the gain attributed to debate \cite{choi2025debatevote}, and multi-agent systems fail in recurring ways \cite{cemri2025why}. Much of this work tunes how agents exchange information, by grouping agents to reduce the cost of debate \cite{liu2026groupdebate}, choosing between voting and consensus \cite{kaesberg2025voting}, or shaping the communication topology so that errors propagate less \cite{shen2025topologies,huang2025resilience}. Agents conform to numerically dominant groups \cite{choi2025conformity,weng2025conformity}, debates among models with similar responses converge on the majority view, shared misconceptions included \cite{estornell2024debate}, agents form reputations from second-hand gossip \cite{ren2026repunet}, and agreement after an exchange is a poor signal of correctness \cite{dai2026consensus}. Correlated errors shrink the effective size of judge panels \cite{verga2024juries,kohli2026judges}, and evidence spread across agents often fails to be pooled \cite{li2025hidden,yan2026truth}. In shared memories, derivatives of one source form false majorities unless an arbiter estimates how many independent sources there are \cite{lin2026cama}. Other work tests assumptions that multi-agent designs take for granted, such as the interchangeability of agents that fill the same role \cite{gao2026interchangeability}. We test an assumption about the component that reads the messages, namely that it counts each source once, and we trace the statements it reads back to the protocol that writes them.

\paragraph{Repetition, conflict and sources in LLM context.}
Model answers follow the majority of retrieved evidence and shift with the number of supporting documents \cite{xie2024chameleon,wang2025ramdocs}, and retrieved content that departs further from the truth is adopted less often \cite{wu2024clasheval}. Sourcing cues are underused \cite{wan2024evidence}, paraphrases of one argument flip answers more often than distinct supporting documents \cite{naphade2026groupqa}, and among contradictory facts hidden in a long context, models favour the repeated one \cite{kurfali2025needles}. Under knowledge conflicts, repeating a less credible source reverses preferences for credible ones whether the repetition is attributed to one source or to several. A warning against repetition weakens this effect and fine-tuning reduces it further \cite{schuster2026whose}, much as an instruction to ignore irrelevant context reduces distraction \cite{shi2023distracted}. A recent benchmark poses exact Bayesian cases that contrast one report echoed several times with independent reports \cite{gulliblebench2026}. Graph agents that reach one origin along redundant paths shift their stated confidence in a direction that depends on the model, and all of them revisit that origin more often \cite{wang2026graphecho}. In people, repetition alone raises judged truth \cite{hasher1977frequency,dechene2010truth}. Relative to these studies, we express the effect of repetition in units of observations against an exact posterior, follow it to decisions with explicit costs, and trace it to the agent protocols that create the restatements.

\paragraph{Aggregation under dependence.}
Opinion pooling, jury theorems and belief merging make the treatment of dependent sources explicit \cite{genest1986combining,winkler1981combining,ladha1992condorcet,dietrich2013independent,konieczny2002merging,everaere2015belief}, and social learning models show how repeated exposure distorts beliefs \cite{degroot1974consensus,golub2010naive,jain2025interacting}. Persuasion bias, the failure to discount information that reaches an agent repeatedly, gives well-connected sources outsized influence \cite{demarzo2003persuasion}, and naive herding treats the actions of others as independent signals although they repeat earlier information \cite{eyster2010naive}. Data-incest correction removes double counting when the network is known \cite{krishnamurthy2014incest}, and covariance intersection stays consistent when it is not \cite{julier1997nondivergent}. Pooling reports of one evidential root as if they were independent grows overconfident as the reports multiply \cite{bara2026sybil}, and leaving the combination to arithmetic with a discount for dependent sources avoids this by design \cite{wu2026split}. Provenance semirings separate alternative derivations of a fact from its joint use \cite{green2007provenance}, deduplication removes repeated content \cite{broder1997resemblance,lee2022dedup,abbas2023semdedup}, and provenance-aware memories for agents are emerging \cite{wang2026mapgraph,liao2026provenance}. All of these handle dependence with an explicit model or rule. We measure whether an LLM honours dependence that it is only told about in text.

\paragraph{Probabilistic reasoning in LLMs.}
LLMs update approximately like Bayesian agents on in-context evidence \cite{gupta2025coinflips}, and their verbalised probabilities carry calibration error \cite{tian2023calibration}. We measure deviations from the Bayesian posterior in units of independent observations, and the conflict probes of Section~\ref{sec:t1} read over-counting from the direction of the answer.

\section{Conclusion}

We measured how much LLM aggregators count restated evidence, in units of independent readings. On controlled logs, web documents and agent-written logs, a copy carries a model-specific fraction of the weight of a source, enough to move decisions that the evidence does not support. Most of this weight comes from occasional full counting, which crosses commit boundaries even where the average belief stays below them. The copy weight acts on a quantity that the protocol sets, and the two can be measured separately: the weight from reported probabilities on paired logs, the restatement count from transcripts. Merging copies before the decision removes most of the inflation, whether the aggregator merges them or the protocol never creates them. For protocol design the results point to a simple practice, in which agents refer to readings already in the log instead of restating them.

Our controlled logs make provenance honest and explicit. Labels are correct, copies are exact, and the aggregator sees the full log, so every reference can be resolved. Forged identifiers, relays that alter what they forward and sources with shared failure modes call for an oracle that models these dependencies, a natural next step for this measure. The measure itself needs only reported probabilities on logs that differ in how the evidence is restated, so it applies to any aggregator and any protocol that can be run on such logs.

\clearpage
\balance

\clearpage
\nobalance
\makeatletter
\let\@LN@makecol\ArxivAppendixLNMakecol
\makeatother
\appendix
\begingroup
\ArxivAppendixFormatting
\section{Overview}\label{app:sec:glance}

Table~\ref{app:tab:episodes} lists every kind of episode in the study. An episode is the smallest unit that produces one recorded value. The number of episodes per model is the number of items times the number of conditions. All conditions of a testbed use the same items, so every contrast in the paper is paired within item.

\begin{table*}[!tp]
\centering\small
\caption{Episode types. The four forms are \cond{Single}, \cond{Copies}, \cond{Independent} and \cond{Copied minority} (Table~\ref{app:tab:renderings}). The arms are listed in Table~\ref{app:tab:arms}.}
\label{app:tab:episodes}
\begin{tabularx}{\linewidth}{@{}>{\raggedright\arraybackslash}p{2.45cm}>{\raggedright\arraybackslash}X>{\raggedright\arraybackslash}p{2.2cm}l@{}}
\toprule
Block & One episode & Output & Episodes \\
\midrule
T1 forms & one item, one of the 27 forms, \arm{No note}, one model & probability & $1000\times27$ per model \\
T1 notes & one item, one of 7 arms, one of the 4 forms & probability & $500\times28$ per model \\
T1 actions & one item, \arm{No note} or \arm{Declaration: records}, one of the 4 forms, costs shown & action & $500\times8$ per model \\
T1 policy & one stated posterior, costs shown, no evidence & action & $350\times2$ per model \\
T1 reasoning & one item, one of the 4 forms, one of 6 ablation settings & probability & $500\times4$ per setting \\
T1 replicates & one item, one of the 4 forms, sample 1 or 2 & probability & $500\times8$ per model \\
T1 additional notes & one item, \arm{Declaration: copies} or \arm{Declaration and instruction} on one of the 4 forms, or \arm{Field guide}, \arm{Declaration: records} or \arm{Placebo: display} on \cond{Claims} or \cond{Packed} & probability & $500\times14$ per model \\
\midrule
T2 & one item, one of 10 document sets & probability & $500\times10$ per model \\
\midrule
T3 & one configuration, protocol and structure: worker turns, annotation, aggregator calls & transcript, $\mu$, probabilities & 4{,}000 \\
T3 reference rule & the same with the reference rule in the worker prompt, aggregator without note & transcript, $\mu$, probability & 4{,}000 \\
T3 true list & one configuration and structure, true readings listed & probability & $1000$ per model \\
T3 second worker & as T3 without a rule, on the configurations with $j\le2$, written by \texttt{deepseek-off} instead of \texttt{luna-none}; aggregator without note & transcript, $\mu$, probability & 800 \\
\bottomrule
\end{tabularx}
\end{table*}

\subsection{Models and settings}

The study uses four models, called by short names throughout: Luna (GPT-6 Luna, model identifier \texttt{gpt-6-luna}), Sol (GPT-6.1 Sol, \texttt{gpt-6.1-sol}), Gemini (Gemini 3.8 Flash, \texttt{gemini-3.8-flash}) and DeepSeek (DeepSeek V4.1 Flash, \texttt{deepseek-flash}). Table~\ref{app:tab:settings} lists the settings in which they are called.

\begin{table*}[!tp]
\centering\small
\caption{Model settings. The four low-effort settings run every block, and the others only the blocks in the last column.}
\label{app:tab:settings}
\begin{tabularx}{\linewidth}{@{}lllrX@{}}
\toprule
Setting & Model & Reasoning control & Max output & Used in \\
\midrule
\texttt{luna-low} & Luna & effort \texttt{low} & 16k & all blocks \\
\texttt{sol-low} & Sol & effort \texttt{low} & 16k & all blocks; T2 summaries \\
\texttt{gemini-low} & Gemini & thinking level \texttt{low} & 16k & all blocks; T2 stance checks \\
\texttt{deepseek-low} & DeepSeek & thinking on, effort \texttt{low} & 16k & all blocks \\
\midrule
\texttt{luna-none} & Luna & effort \texttt{none} & 4k & T1 reasoning; T3 worker \\
\texttt{deepseek-off} & DeepSeek & thinking off & 4k & T1 reasoning; T3 second worker \\
\texttt{luna-high} & Luna & effort \texttt{high} & 64k & T1 reasoning \\
\texttt{sol-high} & Sol & effort \texttt{high} & 64k & T1 reasoning; T3 annotation \\
\texttt{gemini-high} & Gemini & thinking level \texttt{high} & 64k & T1 reasoning \\
\texttt{deepseek-high} & DeepSeek & thinking on, effort \texttt{high} & 64k & T1 reasoning \\
\bottomrule
\end{tabularx}
\end{table*}

Every call is stateless and consists of one system text and one user text. No temperature, top-$p$, seed, structured-output mode or tool is set. The output limits in Table~\ref{app:tab:settings} are generous so that reasoning is not truncated.

\paragraph{API calls.} Luna and Sol use the OpenAI Responses API with \texttt{instructions} set to the system text, \texttt{input} set to the user text, and \texttt{reasoning.effort}. Gemini uses \texttt{generateContent} with \texttt{system\_instruction} set to the system text, one user part, and \texttt{thinking\_config.thinking\_level}. DeepSeek uses \texttt{chat/completions} with one system and one user message, \texttt{thinking} set to \texttt{\{"type": "enabled"\}} or \texttt{\{"type": "disabled"\}}, and \texttt{reasoning\_effort}. Reasoning summaries, where an API returns them, are never parsed. All calls used the providers' official APIs.

\subsection{Replies and parsing}\label{app:sec:parsing}
Every prompt asks for a final line of a fixed form. The answer is read from the last line that matches, after deleting \texttt{*} and \texttt{\textasciigrave}. Reasoning text is never parsed.
\begin{itemize}
\item Probability: the pattern below, case-insensitive. A value with \texttt{\%} is divided by 100. A value outside $[0,1]$ is a failure.
\begin{Pattern}
PROBABILITY\s*[:=]\s*(\d*\.?\d+)\s*(%)?\s*\.?\s*$
\end{Pattern}
\item Action: after upper-casing and replacing \texttt{\_} and \texttt{-} by spaces, the pattern below, recorded as \texttt{Q}, \texttt{P} or \texttt{N}.
\begin{Pattern}
ACTION\s*[:=]\s*(QUERY|COMMIT\s+(<POS>|<NEG>))
\end{Pattern}
\end{itemize}
A reply that does not parse, or that is cut off by the output limit, is requested again up to two more times. A reply that still fails after the third attempt counts as missing.
\section{T1: Controlled logs}\label{app:sec:t1}

\subsection{World, oracle and items}\label{app:sec:t1items}
A hidden state has two values with prior $1/2$. Each sensor takes one reading, which matches the state with probability $q$, and distinct sensors err independently. With $L=\log\frac{q}{1-q}$ and $a$ the direction of the primary reading, the oracle evidence of a log is $e^\ast=a\sum_{o\in\mathcal O}x_o$ over the distinct readings $\mathcal O$, identified by \texttt{reading\_id}.

There are 1{,}000 items: 5 domains (Table~\ref{app:tab:domains}) $\times$ $q\in\{0.60,\allowbreak 0.65,\allowbreak 0.70,\allowbreak 0.75,\allowbreak 0.80\}$ $\times$ 2 polarities $\times$ 20 items, indexed $j=1,\dots,20$. Items with $j\le10$ form the 500-item subset used by all T1 blocks except T1 forms.

Each item fixes the following, and all forms of an item reuse them:
\begin{itemize}
\item eight sensors, each with an identifier \texttt{<prefix>-NN}, a reading identifier \texttt{rd-NNNN} and a time \texttt{HH:MM};
\item seven relay names drawn from Kestrel, Heron, Otter, Lynx, Falcon, Marten, Wren, Badger, Ibis, Puffin, Raven and Stoat,
\item values for the heartbeat fields.
\end{itemize}
The values are drawn from a pseudo-random generator seeded with the item identifier.

\begin{table*}[!tp]
\centering\small
\caption{Domains. \texttt{POS} is the state named in the statement.}
\label{app:tab:domains}
\begin{tabularx}{\linewidth}{@{}lXllc@{}}
\toprule
ID & Team; statement & \texttt{POS} & \texttt{NEG} & Prefix \\
\midrule
D1 & the surface operations team of a planetary rover mission; a dust storm will reach the landing site within the next six hours & \texttt{STORM} & \texttt{CLEAR} & \texttt{OD} \\
D2 & a reservoir ecology monitoring team; invasive zebra mussels are established in the reservoir & \texttt{PRESENT} & \texttt{ABSENT} & \texttt{ED} \\
D3 & the process safety team of a chemical plant; the coolant loop of reactor 2 is overheating & \texttt{OVERHEAT} & \texttt{NORMAL} & \texttt{TP} \\
D4 & a city air-quality monitoring team; wildfire smoke will push fine-particle levels above the health limit in the east district tonight & \texttt{EXCEED} & \texttt{BELOW} & \texttt{PM} \\
D5 & a coastal water-quality monitoring team; a harmful algal bloom is present in the north bay & \texttt{BLOOM} & \texttt{NO\_BLOOM} & \texttt{FL} \\
\bottomrule
\end{tabularx}
\end{table*}

\subsection{Records and forms}\label{app:sec:t1forms}
The 27 forms are listed in Table~\ref{app:tab:renderings}. Full prompt and record templates are provided separately in Section~\ref{app:sec:promptcollection}, with a complete episode in Section~\ref{app:sec:t1episode}.

{\small\setlength{\tabcolsep}{4pt}
\begin{table*}[!tp]
\centering
\caption{The 27 forms of block T1 forms (JSON with labels unless stated). A subscript gives the dose $m$ of a series. Without one, $m=3$, so \cond{Copies} is \cond{Copies}$_3$. Statements: statements of the primary reading's value, where a claim of equality counts as one. $e^\ast$: oracle evidence.}\label{app:tab:renderings}
\small\setlength{\tabcolsep}{3pt}
\begin{adjustbox}{max width=\textwidth}
\begin{tabular}{@{}>{\raggedright\arraybackslash}p{0.3\linewidth}>{\raggedright\arraybackslash}p{0.43\linewidth}ccc@{}}
\toprule
Form & Construction & Records & Statements & $e^\ast$ \\
\midrule
\cond{Single} & original of sensor 1 & 1 & 1 & 1 \\
\cond{Single (prose)} & the same in prose & 1 & 1 & 1 \\
\cond{Packed}$_m$, $m\in\{3,8\}$ & packed record with 2 or 7 hops & 1 & 3, 8 & 1 \\
\cond{Trace} & packed reference record with 2 hops & 1 & 1 & 1 \\
\cond{Pointers} & original, then references from relays 1 and 2 & 3 & 1 & 1 \\
\cond{Claims} & original, then equality claims from relays 1 and 2 & 3 & 3 & 1 \\
\cond{Copies}$_m$, $m\in\{2,3,5,8\}$ & original, then forwards from relays $1,\dots,m-1$ & $m$ & $m$ & 1 \\
\cond{Copies (one relay)} & as \cond{Copies}, both forwards from relay 1 & 3 & 3 & 1 \\
\cond{Padding}$_m$, $m\in\{3,8\}$ & original, then 2 or 7 heartbeats matched to \cond{Copies} or \cond{Copies}$_8$ & 3, 8 & 1 & 1 \\
\cond{Independent}$_m$, $m\in\{2,3,5,8\}$ & originals of sensors $1,\dots,m$, ordered by time, then by sensor identifier & $m$ & 1 & $m$ \\
\cond{Copies (no labels)} & \cond{Copies} without labels & 3 & 3 & 1 \\
\cond{Copies (prose)} & \cond{Copies} in prose, with labels & 3 & 3 & 1 \\
\cond{Copies (prose, no labels)} & \cond{Copies} in prose, without labels & 3 & 3 & 1 \\
\cond{Independent (no labels)} & \cond{Independent} without labels & 3 & 1 & 3 \\
\cond{Independent (prose)} & \cond{Independent} in prose, with labels & 3 & 1 & 3 \\
\cond{Independent (prose, no labels)} & \cond{Independent} in prose, without labels & 3 & 1 & 3 \\
\cond{Tie} & sensor 1 with the primary value, sensor 2 with the opposite & 2 & 1 & 0 \\
\cond{Copied tie} & block A: sensor 1 and two forwards; block B: sensor 2, opposite value & 4 & 3 & 0 \\
\cond{Copied minority} & block A as above; block B: sensors 2 and 3, opposite value & 5 & 3 & $-1$ \\
\bottomrule
\end{tabular}
\end{adjustbox}
\end{table*}

}

In the conflict probes, block A comes first when $j$ is odd, and block B when $j$ is even. Within block B, records are ordered by time and then by sensor identifier. Forwards point to the \texttt{report\_id} that the original of sensor 1 receives in display order.
\section{T2: Web evidence}\label{app:sec:t2}

\subsection{Items}\label{app:sec:t2items}
We use ConflictingQA~\cite{wan2024evidence}: controversial yes-or-no questions with web paragraphs whose stance two models agreed on. The file \texttt{data.pkl} comes from \url{https://github.com/AlexWan0/rag-convincingness} (MIT licence) and is read with \texttt{pandas.read\_pickle}. The fields used are \texttt{search\_query} (the question), \texttt{url}, \texttt{title}, \texttt{text\_window} and \texttt{stance}.
\begin{enumerate}
\item Rows are kept if their stance, stripped and lower-cased, is \texttt{yes} or \texttt{no} and their \texttt{text\_window} has at least 50 words. Whitespace in texts and titles is collapsed, and one row is kept per URL and then per text within a question. The registrable domain of a URL is \texttt{domain + "." + suffix} from \texttt{tldextract}, and the display domain is the host without \texttt{www.}.
\item Within each question and stance, rows are sorted by URL and deterministically shuffled, same side first. Up to two disjoint triples $(F,D_1,D_2)$ are built from the same side by taking each next unused row whose registrable domain differs from those already in the triple, and $O$ is the first opposite-side row whose domain differs from all three.
\item All triples are ordered by a deterministic hash of their question, stance and triple index, and the first 500 are kept as items $k=0,\dots,499$, named \texttt{T2-0000} onwards. Bootstrap intervals resample questions, since items from one question share documents.
\item $F$ precedes $O$ when $k$ is even, and $O$ precedes $F$ otherwise.
\end{enumerate}

\subsection{Copies and conditions}\label{app:sec:t2copies}
Each copy is appended after $F$ and $O$.
\begin{itemize}
\item Duplicate: $F$ repeated verbatim with its own URL and title.
\item Republished copy $j\in\{0,1\}$: $F$'s title and text under \path{https://<outlet>/syndicated/<h>}, with the line \texttt{Republished from <display domain of F>.} The outlet is entry $(k+j)\bmod4$ of \path{news-digest.example.com}, \path{worldwire.example.net}, \path{dailybrief.example.org}, \path{headline-hub.example.com}, and \texttt{<h>} is the first 8 hex digits of the SHA-256 of $F$'s URL.
\item Summary $j\in\{0,1\}$: a paraphrase of $F$ generated by Sol (\texttt{sol-low}) in two independent calls and checked for stance by Gemini (\texttt{gemini-low}) (Section~\ref{app:sec:t2prompts}). It is titled \texttt{Summary: <F title>}, carries the line \texttt{Summary of an article originally published by <display domain of F>.}, and sits at \path{https://<outlet>/summary/<h>-<j+1>}, where the outlet is entry $(k+j)\bmod4$ of \path{summary-desk.example.com}, \path{briefing-room.example.net}, \path{recap-daily.example.org}, \path{digest-review.example.com}. Every model sees the same two summaries.
\end{itemize}
The ten conditions are \cond{Base}, the pair $[F,O]$, and the base with two duplicates (\cond{Duplicate}), two republished copies (\cond{Republished}), two summaries (\cond{Summary}) or $D_1$ and $D_2$ (\cond{Distinct}), each without and with the declaration.

Full system and user texts, including the paraphrase and stance-check prompts, are provided in Section~\ref{app:sec:t2prompts}.
\section{T3: Agent teams}\label{app:sec:t3}

\subsection{Configurations, roles and protocols}\label{app:sec:t3config}
There are 500 configurations: 5 domains $\times$ 5 reliabilities $\times$ 2 polarities $\times$ 10 ($j=1,\dots,10$), with the same domains as T1. Each fixes four agent names $A_1,\dots,A_4$ and three sensors. All readings take the configuration's polarity. The two evidence structures are:
\begin{itemize}
\item \emph{one observer}: $A_1$ is the only observer (sensor 1), and $e^\ast=1$.
\item \emph{three observers}: $A_1$, $A_2$ and $A_3$ observe sensors 1, 2 and 3, $A_4$ has no sensor, and $e^\ast=3$.
\end{itemize}
Every configuration runs all four protocols under both structures and two communication rules: no rule (4{,}000 episodes) and the reference rule of Section~\ref{app:sec:t3worker} (4{,}000 episodes). The aggregator is scored under the arms \emph{no note}, \emph{declaration}, \emph{extract first} and \emph{true list} (Section~\ref{app:sec:t3agg}).

\begin{table*}[!tp]
\centering\small
\caption{Protocol schedules. Each turn is one call to the worker model (Luna, setting \texttt{luna-none}). A turn shows the speaker only what is listed in parentheses.}
\label{app:tab:schedules}
\begin{tabularx}{\linewidth}{@{}lXc@{}}
\toprule
Protocol & Turns & \begin{tabular}[b]{@{}c@{}}Messages\\(one/three observers)\end{tabular} \\
\midrule
Relay & $A_1\to A_2$; $A_2$ (sees $A_1$'s message) $\to A_3$; $A_3$ (sees $A_2$'s) $\to A_4$; $A_4$ (sees $A_3$'s) $\to$ aggregator & 4/4 \\
Board & round 1: each observer posts, seeing the board; round 2: $A_1$ to $A_4$ each post, seeing the board & 5/7 \\
Discussion & two rounds of $A_1$ to $A_4$, each seeing the whole discussion & 8/8 \\
Hierarchy & $A_1\to A_2$ and $A_3\to A_4$ (reports); $A_2$ and $A_4$ summarise their member's report for the aggregator; $A_2$ writes a final summary seeing all messages & 5/5 \\
\bottomrule
\end{tabularx}
\end{table*}

Full worker and aggregator prompts are provided separately in Sections~\ref{app:sec:t3worker} and~\ref{app:sec:t3agg}, and a complete episode in Section~\ref{app:sec:t3episode}.

\subsection{Annotation}\label{app:sec:t3annot}
Sol (setting \texttt{sol-high}) annotates each transcript once using the prompt in Section~\ref{app:sec:t3annotprompt}.

A line matching
\begin{Pattern}
^\s*m(\d+)\s*:\s*(.*?)\s*$
\end{Pattern}
gives the readings stated by message $k$, as items \texttt{<sensor\_id>}\allowbreak\texttt{=<value>} separated by semicolons or the word \texttt{none}, with an optional trailing \texttt{| CLAIM}. An annotation is valid only if every message has a line, and an invalid one is requested again, up to three attempts. From a valid annotation, three counts are derived:
\begin{itemize}
\item $\mu_k$ is the number of messages that state sensor $k$'s reading, which is the restatement count of the main text because fewer than 1\% of these messages state the value twice.
\item \texttt{n\_claim} is the number of messages flagged \texttt{CLAIM}.
\item \texttt{n\_restating} is the number of messages that state any reading and were not sent by that reading's observer.
\end{itemize}
A three-observer log is \emph{complete} if $\mu_1,\mu_2,\mu_3\ge1$, that is, if every observer's reading is stated at least once. Only complete logs carry three readings, so three-observer results use complete logs. The share of incomplete logs is reported by protocol (Section~\ref{app:sec:res-t3}).

To check the annotation, 240 messages were drawn from the transcripts: 25 from each protocol and structure without the rule, and 10 from each protocol under the reference rule with one observer. Two authors labelled each message independently on two questions, whether it states sensor 1's reading and whether it is a \texttt{CLAIM}, and agreed on all 240 messages on both questions. The annotation matches their label on 98.8\% of messages for the first question (54\% of the messages state the reading) and on all messages for the second.
\section{Measures and tests}\label{app:sec:analysis}

Probabilities are clipped to $[0.001,0.999]$. The largest measurable evidence is therefore $\logit(0.999)/L$: 17.0, 11.2 and 8.2 readings for $q=0.60$, 0.65 and 0.70, but 6.3 and 5.0 for $q=0.75$ and 0.80, where eight readings cannot be registered. Results that depend on eight readings are therefore also reported for $q\le0.70$ (Table~\ref{app:tab:res-sat}). Below, $\bar e(X)$ is the mean reported evidence on form $X$, and a subscript gives the dose of a series as in Table~\ref{app:tab:renderings}.
\begin{itemize}
\item \emph{Reported evidence} (T1, T3): $\hat e=a\,\logit\hat p/L$.
\item \emph{Copy weight}: $w_m=\bigl(\bar e(\text{\cond{Copies}}_m)-1\bigr)/(m-1)$, and likewise for \cond{Packed}$_m$ and for the variants of \cond{Copies}.
\item \emph{Discrimination}: $D_m=\allowbreak\frac{\bar e(\text{\cond{Independent}}_m)-\bar e(\text{\cond{Copies}}_m)}{m-1}$. In T3, $D=(\bar e_{\text{three}}-\bar e_{\text{one}})/2$, the difference between three observers and one observer, paired by configuration, over complete three-observer logs.
\item \emph{Belief-implied early commitment} at boundary $e_b$: $P(\hat e>e_b)$ on one-reading forms. A report exactly at the boundary counts as a query, as for the oracle. Reports below $-e_b$, which would commit to the state that the readings oppose, make up less than 0.3\% of any condition and are not counted. It is reported at $e_b=2$ (the action-channel boundary) and $e_b=1.15$, and as a curve over $e_b\in[1,3]$. \emph{Unnecessary query}: $P(\hat e\le e_b)$ on \cond{Independent}.
\item \emph{Conflict probes}: mean $\hat e$ (oracle 0 for \cond{Tie} and \cond{Copied tie}, $-1$ for \cond{Copied minority}), and for \cond{Copied minority} the share with $\hat e>0$, which sides with the copied reading.
\item \emph{Actions}: accuracy against the oracle action. The implied action applies the oracle decision rule, with the item's $r$ and $c$, to the same model's probability on the same item, form and arm. A wrong emitted action is a belief-stage error if the implied action is also wrong, and an action-stage error otherwise.
\item \emph{T3 slope}: least-squares slope of $\hat e$ on $\mu_1-1$ over one-observer episodes under \emph{no note}, with protocol fixed effects, per model.
\item \emph{Reference rule}: per protocol and model, $\mu_1$, $\hat e$ under \emph{no note} and $P(\hat e>2)$ on one-observer episodes, and $D$, each under the reference rule against no rule, paired by configuration.
\item \emph{T2}: $\Delta_c=a\,(\logit\hat p_c-\logit\hat p_{\text{base}})$, with $a=+1$ when the focal answer is Yes, and $\rho_c=\bar\Delta_c/\bar\Delta_{\text{distinct}}$.
\item \emph{Synthesis}: the weight of a copy in units of an independent source, measured three ways: $w_3$ in T1, the T3 slope, and $\rho$ per copy type in T2.
\end{itemize}
The copy weight and the discrimination are also reported by reliability and by domain, and the T3 measures by protocol.

\begin{table*}[!tp]
\centering\small
\caption{Pre-specified test families. Holm correction within each family and model. F1 to F5 and T3-F1, T3-F3, T3-F5, T2-F1, T2-F3 are confirmatory. Tests: Wilcoxon signed-rank on $\hat e$ or $\Delta$ (paired, or one-sample against the oracle), exact McNemar on binary outcomes, Page's test for dose. Intervals: 10{,}000 bootstrap resamples over items (T1), configurations (T3) or questions (T2).}
\label{app:tab:tests}
\begin{tabularx}{\linewidth}{@{}>{\raggedright\arraybackslash}p{2.3cm}>{\raggedright\arraybackslash}X@{}}
\toprule
Family & Contrasts \\
\midrule
F1 decomposition & \cond{Copies}$-$\cond{Single}; \cond{Packed}$-$\cond{Trace}; \cond{Pointers}$-$\cond{Single}; \cond{Padding}$-$\cond{Single}; \cond{Claims}$-$\cond{Pointers}; \cond{Copies}$-$\cond{Padding}; \cond{Copies}$-$\cond{Copies (one relay)}; \cond{Independent}$-$\cond{Copies} \\
F2 representation & on \cond{Copies}, labels within JSON and within prose, and format with and without labels; the same on \cond{Independent}; \cond{Single}$-$\cond{Single (prose)} \\
F3 dose & Page's test over \cond{Single}, \cond{Copies}$_2$, \dots, \cond{Copies}$_8$ and over \cond{Single}, \cond{Independent}$_2$, \dots, \cond{Independent}$_8$; \cond{Packed}$_8-$\cond{Packed}; \cond{Copies}$_8-$\cond{Padding}$_8$; \cond{Padding}$_8-$\cond{Single} \\
F4 conflict & \cond{Copied tie}$-$\cond{Tie}; \cond{Copied tie} against 0; \cond{Copied minority} against $-1$ \\
F5 notes & each arm against \arm{No note} on \cond{Copies}, \cond{Independent}, \cond{Copied minority}, \cond{Single}; \arm{Placebo: display}$-$\arm{Placebo: unrelated} and \arm{Placebo: delivery}$-$\arm{Placebo: unrelated} on \cond{Copies}; \arm{Declaration: records}$-$\arm{Declaration: relays} on \cond{Copies} \\
F6 actions & emitted early commitment on \cond{Copies} against \cond{Single}; \arm{Declaration: records} against \arm{No note} on \cond{Copies}, \cond{Independent}, \cond{Copied minority} \\
F7 reasoning & each ablation setting against low effort on \cond{Copies}, \cond{Independent}, \cond{Copied minority} \\
T3-F1 to F5 & $\hat e$ against 1 with one observer, per protocol; $D$ against 1 per protocol; \emph{extract first} and \emph{declaration} against \emph{no note} on both structures; slope against 0; reference rule against no rule per protocol ($\mu_1$ and $\hat e$ with one observer, $D$) \\
T2-F1 to F3 & $\Delta$ against 0 for the three copy types; each copy type against distinct documents; declaration effect for all four \\
\bottomrule
\end{tabularx}
\end{table*}

Null controls (\cond{Pointers}, \cond{Trace}, \cond{Padding}, \cond{Single (prose)} and \cond{Padding}$_8$, each against \cond{Single}) are called equivalent only if the 90\% paired interval lies within $\pm0.10$ readings. Paired contrasts use items with both values present.
\FloatBarrier
\clearpage
\section{Full results}\label{app:sec:results}

This section follows the order of the main paper. Intervals are 95\% percentile bootstrap intervals with 10{,}000 resamples over items, configurations or questions. Where no event occurs, as for early commitment under \emph{extract first}, no interval is shown, because the bootstrap cannot bound a rate that is zero in the sample. With $n$ independent units and no event, the one-sided 95\% upper bound is about $3/n$. Section~\ref{app:sec:res-tests} gives the pre-specified tests.

\subsection{T1 Controlled logs}\label{app:sec:res-t1}

\subsubsection{Copy weight, discrimination and conflict probes (Sections 4.1 and 4.2 of the main paper)}
Table~\ref{app:tab:res-forms} gives the mean reported evidence $\hat e$, in readings, for all 27 forms on the 1{,}000 items, next to the oracle value $e^\ast$. Table~\ref{app:tab:res-weights} gives the copy weight $w_m=(\bar e-1)/(m-1)$ of \cond{Copies} and of the other forms that state one reading several times, and the discrimination $D_m$ between $m$ independent readings and one reading stated $m$ times. Figure~\ref{app:fig:res-dose} plots the dose series, Table~\ref{app:tab:res-conflict} gives the conflict probes, and Table~\ref{app:tab:res-breakdown} breaks $w_3$ and $D_3$ down by reliability and by domain.

{\footnotesize\setlength{\tabcolsep}{4pt}
\begin{table*}[!tp]
\centering
\caption{Mean reported evidence $\hat e$ for every form of the controlled logs.}\label{app:tab:res-forms}
\small\setlength{\tabcolsep}{3pt}
\begin{adjustbox}{max width=\textwidth}
\begin{tabular}{@{}>{\raggedright\arraybackslash}p{0.30\linewidth}ccccc@{}}
\toprule
Form & $e^\ast$ & Luna & Sol & Gemini & DeepSeek \\
\midrule
\multicolumn{6}{@{}l}{\emph{One reading, stated once}} \\
\quad \cond{Single} & 1 & 1.00\,{\scriptsize[1.00, 1.00]} & 1.00\,{\scriptsize[1.00, 1.00]} & 1.00\,{\scriptsize[1.00, 1.00]} & 0.99\,{\scriptsize[0.99, 1.00]} \\
\quad \cond{Single} (prose) & 1 & 1.00\,{\scriptsize[0.99, 1.00]} & 1.00\,{\scriptsize[1.00, 1.00]} & 1.00\,{\scriptsize[1.00, 1.00]} & 1.00\,{\scriptsize[1.00, 1.00]} \\
\quad \cond{Pointers} & 1 & 1.00\,{\scriptsize[1.00, 1.01]} & 1.00\,{\scriptsize[1.00, 1.00]} & 1.00\,{\scriptsize[1.00, 1.00]} & 1.01\,{\scriptsize[1.00, 1.02]} \\
\quad \cond{Trace} & 1 & 1.00\,{\scriptsize[1.00, 1.01]} & 1.00\,{\scriptsize[1.00, 1.01]} & 1.01\,{\scriptsize[1.00, 1.01]} & 1.00\,{\scriptsize[1.00, 1.01]} \\
\quad \cond{Padding} & 1 & 1.00\,{\scriptsize[0.99, 1.00]} & 1.00\,{\scriptsize[1.00, 1.00]} & 1.00\,{\scriptsize[0.99, 1.00]} & 1.00\,{\scriptsize[0.99, 1.00]} \\
\quad \cond{Padding}, $m=8$ & 1 & 1.00\,{\scriptsize[1.00, 1.01]} & 1.00\,{\scriptsize[1.00, 1.00]} & 1.00\,{\scriptsize[1.00, 1.00]} & 1.00\,{\scriptsize[1.00, 1.01]} \\
\addlinespace[3pt]
\multicolumn{6}{@{}l}{\emph{One reading, stated several times}} \\
\quad \cond{Copies}, 2 statements & 1 & 1.21\,{\scriptsize[1.19, 1.24]} & 1.06\,{\scriptsize[1.04, 1.07]} & 1.18\,{\scriptsize[1.16, 1.20]} & 1.35\,{\scriptsize[1.32, 1.38]} \\
\quad \cond{Copies} & 1 & 1.55\,{\scriptsize[1.50, 1.60]} & 1.12\,{\scriptsize[1.09, 1.14]} & 1.42\,{\scriptsize[1.38, 1.47]} & 1.84\,{\scriptsize[1.79, 1.90]} \\
\quad \cond{Copies}, 5 statements & 1 & 1.82\,{\scriptsize[1.74, 1.90]} & 1.20\,{\scriptsize[1.16, 1.25]} & 1.75\,{\scriptsize[1.67, 1.84]} & 2.26\,{\scriptsize[2.16, 2.35]} \\
\quad \cond{Copies}, 8 statements & 1 & 2.03\,{\scriptsize[1.93, 2.12]} & 1.30\,{\scriptsize[1.24, 1.37]} & 1.93\,{\scriptsize[1.83, 2.03]} & 2.53\,{\scriptsize[2.42, 2.64]} \\
\quad \cond{Copies} (one relay) & 1 & 1.62\,{\scriptsize[1.57, 1.67]} & 1.11\,{\scriptsize[1.09, 1.14]} & 1.37\,{\scriptsize[1.33, 1.42]} & 1.78\,{\scriptsize[1.73, 1.84]} \\
\quad \cond{Copies} (no labels) & 1 & 1.69\,{\scriptsize[1.64, 1.75]} & 1.20\,{\scriptsize[1.16, 1.23]} & 1.53\,{\scriptsize[1.48, 1.58]} & 1.97\,{\scriptsize[1.91, 2.03]} \\
\quad \cond{Copies} (prose) & 1 & 1.31\,{\scriptsize[1.27, 1.35]} & 1.05\,{\scriptsize[1.04, 1.07]} & 1.24\,{\scriptsize[1.20, 1.28]} & 1.55\,{\scriptsize[1.49, 1.60]} \\
\quad \cond{Copies} (prose, no labels) & 1 & 1.42\,{\scriptsize[1.37, 1.46]} & 1.14\,{\scriptsize[1.11, 1.17]} & 1.32\,{\scriptsize[1.28, 1.37]} & 1.70\,{\scriptsize[1.64, 1.76]} \\
\quad \cond{Packed} & 1 & 1.66\,{\scriptsize[1.61, 1.71]} & 1.10\,{\scriptsize[1.07, 1.12]} & 1.36\,{\scriptsize[1.31, 1.40]} & 1.75\,{\scriptsize[1.70, 1.81]} \\
\quad \cond{Packed}, $m=8$ & 1 & 2.33\,{\scriptsize[2.22, 2.43]} & 1.24\,{\scriptsize[1.18, 1.30]} & 1.83\,{\scriptsize[1.73, 1.93]} & 2.46\,{\scriptsize[2.35, 2.58]} \\
\quad \cond{Claims} & 1 & 1.07\,{\scriptsize[1.05, 1.10]} & 1.01\,{\scriptsize[1.00, 1.02]} & 1.04\,{\scriptsize[1.02, 1.05]} & 1.54\,{\scriptsize[1.49, 1.59]} \\
\addlinespace[3pt]
\multicolumn{6}{@{}l}{\emph{Independent readings}} \\
\quad \cond{Independent}, $m=2$ & 2 & 1.96\,{\scriptsize[1.95, 1.97]} & 1.99\,{\scriptsize[1.99, 1.99]} & 1.99\,{\scriptsize[1.98, 1.99]} & 1.96\,{\scriptsize[1.94, 1.97]} \\
\quad \cond{Independent} & 3 & 2.83\,{\scriptsize[2.81, 2.85]} & 2.96\,{\scriptsize[2.95, 2.96]} & 2.94\,{\scriptsize[2.93, 2.95]} & 2.91\,{\scriptsize[2.89, 2.93]} \\
\quad \cond{Independent}, $m=5$ & 5 & 4.47\,{\scriptsize[4.42, 4.52]} & 4.94\,{\scriptsize[4.91, 4.97]} & 4.83\,{\scriptsize[4.79, 4.86]} & 4.79\,{\scriptsize[4.75, 4.83]} \\
\quad \cond{Independent}, $m=8$ & 8 & 5.73\,{\scriptsize[5.62, 5.83]} & 6.74\,{\scriptsize[6.64, 6.83]} & 6.58\,{\scriptsize[6.48, 6.68]} & 6.52\,{\scriptsize[6.43, 6.62]} \\
\quad \cond{Independent} (no labels) & 3 & 2.84\,{\scriptsize[2.82, 2.86]} & 2.95\,{\scriptsize[2.94, 2.96]} & 2.94\,{\scriptsize[2.92, 2.95]} & 2.90\,{\scriptsize[2.88, 2.91]} \\
\quad \cond{Independent} (prose) & 3 & 2.84\,{\scriptsize[2.82, 2.85]} & 2.94\,{\scriptsize[2.93, 2.95]} & 2.93\,{\scriptsize[2.92, 2.94]} & 2.89\,{\scriptsize[2.87, 2.91]} \\
\quad \cond{Independent} (prose, no labels) & 3 & 2.82\,{\scriptsize[2.80, 2.84]} & 2.94\,{\scriptsize[2.93, 2.95]} & 2.93\,{\scriptsize[2.91, 2.94]} & 2.89\,{\scriptsize[2.87, 2.91]} \\
\addlinespace[3pt]
\multicolumn{6}{@{}l}{\emph{Conflicting readings}} \\
\quad \cond{Tie} & 0 & 0.00\,{\scriptsize[0.00, 0.00]} & 0.00\,{\scriptsize[0.00, 0.00]} & 0.00\,{\scriptsize[0.00, 0.00]} & 0.00\,{\scriptsize[0.00, 0.00]} \\
\quad \cond{Copied tie} & 0 & 0.50\,{\scriptsize[0.45, 0.55]} & 0.11\,{\scriptsize[0.09, 0.14]} & 0.35\,{\scriptsize[0.31, 0.40]} & 0.86\,{\scriptsize[0.80, 0.91]} \\
\quad \cond{Copied minority} & $-$1 & $-$0.52\,{\scriptsize[$-$0.56, $-$0.46]} & $-$0.90\,{\scriptsize[$-$0.92, $-$0.87]} & $-$0.64\,{\scriptsize[$-$0.68, $-$0.60]} & $-$0.13\,{\scriptsize[$-$0.18, $-$0.07]} \\
\bottomrule
\end{tabular}
\end{adjustbox}
\end{table*}
\par}
\begin{table*}[!tp]
\centering
\caption{Copy weight and discrimination on the controlled logs.}\label{app:tab:res-weights}
\small\setlength{\tabcolsep}{4pt}
\begin{adjustbox}{max width=\linewidth}
\begin{tabular}{@{}lcccc@{}}
\toprule
& Luna & Sol & Gemini & DeepSeek \\
\midrule
\multicolumn{5}{@{}l}{\emph{Copy weight $w_m$ of \cond{Copies}}} \\
\quad $m=2$ & 0.21\,{\scriptsize[0.19, 0.24]} & 0.06\,{\scriptsize[0.04, 0.07]} & 0.18\,{\scriptsize[0.16, 0.20]} & 0.35\,{\scriptsize[0.32, 0.38]} \\
\quad $m=3$ & 0.27\,{\scriptsize[0.25, 0.30]} & 0.06\,{\scriptsize[0.05, 0.07]} & 0.21\,{\scriptsize[0.19, 0.24]} & 0.42\,{\scriptsize[0.39, 0.45]} \\
\quad $m=5$ & 0.20\,{\scriptsize[0.18, 0.22]} & 0.05\,{\scriptsize[0.04, 0.06]} & 0.19\,{\scriptsize[0.17, 0.21]} & 0.31\,{\scriptsize[0.29, 0.34]} \\
\quad $m=8$ & 0.15\,{\scriptsize[0.13, 0.16]} & 0.04\,{\scriptsize[0.03, 0.05]} & 0.13\,{\scriptsize[0.12, 0.15]} & 0.22\,{\scriptsize[0.20, 0.23]} \\
\addlinespace[3pt]
\multicolumn{5}{@{}l}{\emph{Discrimination $D_m$ between \cond{Independent} and \cond{Copies}}} \\
\quad $m=2$ & 0.75\,{\scriptsize[0.72, 0.77]} & 0.93\,{\scriptsize[0.92, 0.95]} & 0.81\,{\scriptsize[0.78, 0.83]} & 0.61\,{\scriptsize[0.57, 0.64]} \\
\quad $m=3$ & 0.64\,{\scriptsize[0.62, 0.67]} & 0.92\,{\scriptsize[0.91, 0.93]} & 0.76\,{\scriptsize[0.73, 0.78]} & 0.53\,{\scriptsize[0.50, 0.56]} \\
\quad $m=5$ & 0.66\,{\scriptsize[0.64, 0.69]} & 0.94\,{\scriptsize[0.92, 0.95]} & 0.77\,{\scriptsize[0.75, 0.79]} & 0.63\,{\scriptsize[0.61, 0.66]} \\
\quad $m=8$ & 0.53\,{\scriptsize[0.51, 0.55]} & 0.78\,{\scriptsize[0.76, 0.79]} & 0.66\,{\scriptsize[0.64, 0.68]} & 0.57\,{\scriptsize[0.55, 0.59]} \\
\addlinespace[3pt]
\multicolumn{5}{@{}l}{\emph{Copy weight of the other forms with one reading stated several times}} \\
\quad \cond{Packed}, $w_3$ & 0.33\,{\scriptsize[0.30, 0.36]} & 0.05\,{\scriptsize[0.04, 0.06]} & 0.18\,{\scriptsize[0.16, 0.20]} & 0.38\,{\scriptsize[0.35, 0.40]} \\
\quad \cond{Packed}, $w_8$ & 0.19\,{\scriptsize[0.17, 0.21]} & 0.03\,{\scriptsize[0.03, 0.04]} & 0.12\,{\scriptsize[0.10, 0.13]} & 0.21\,{\scriptsize[0.19, 0.23]} \\
\quad \cond{Claims}, $w_3$ & 0.04\,{\scriptsize[0.03, 0.05]} & 0.01\,{\scriptsize[0.00, 0.01]} & 0.02\,{\scriptsize[0.01, 0.03]} & 0.27\,{\scriptsize[0.24, 0.30]} \\
\quad \cond{Copies} (one relay), $w_3$ & 0.31\,{\scriptsize[0.28, 0.34]} & 0.06\,{\scriptsize[0.04, 0.07]} & 0.19\,{\scriptsize[0.16, 0.21]} & 0.39\,{\scriptsize[0.36, 0.42]} \\
\quad \cond{Copies} (no labels), $w_3$ & 0.35\,{\scriptsize[0.32, 0.37]} & 0.10\,{\scriptsize[0.08, 0.12]} & 0.26\,{\scriptsize[0.24, 0.29]} & 0.48\,{\scriptsize[0.46, 0.51]} \\
\quad \cond{Copies} (prose), $w_3$ & 0.15\,{\scriptsize[0.13, 0.18]} & 0.03\,{\scriptsize[0.02, 0.04]} & 0.12\,{\scriptsize[0.10, 0.14]} & 0.27\,{\scriptsize[0.25, 0.30]} \\
\quad \cond{Copies} (prose, no labels), $w_3$ & 0.21\,{\scriptsize[0.18, 0.23]} & 0.07\,{\scriptsize[0.06, 0.09]} & 0.16\,{\scriptsize[0.14, 0.18]} & 0.35\,{\scriptsize[0.32, 0.38]} \\
\bottomrule
\end{tabular}
\end{adjustbox}
\end{table*}
\begin{figure*}[!tp]
\centering
\suppplot{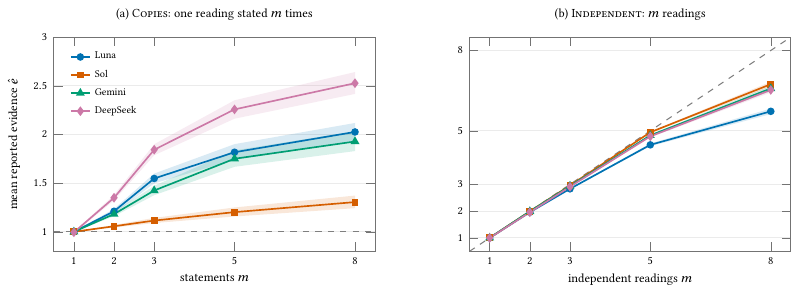}
\caption{Mean reported evidence against the number of statements of one reading (a) and of independent readings (b), with 95\% intervals. Dashed lines mark the oracle.}\label{app:fig:res-dose}
\Description{Two line charts with one line per model and a shaded interval band. In panel a, reported evidence rises above the oracle value of one as the reading is stated more often. In panel b, it rises with the number of independent readings, below the oracle line for larger numbers.}
\end{figure*}
\begin{table*}[!tp]
\centering
\caption{Conflict probes on the controlled logs.}\label{app:tab:res-conflict}
\small\setlength{\tabcolsep}{4pt}
\begin{adjustbox}{max width=\linewidth}
\begin{tabular}{@{}lcccc@{}}
\toprule
& Luna & Sol & Gemini & DeepSeek \\
\midrule
\cond{Tie}: mean $\hat e$ ($e^\ast=0$) & 0.00\,{\scriptsize[0.00, 0.00]} & 0.00\,{\scriptsize[0.00, 0.00]} & 0.00\,{\scriptsize[0.00, 0.00]} & 0.00\,{\scriptsize[0.00, 0.00]} \\
\cond{Copied tie}: mean $\hat e$ ($e^\ast=0$) & 0.50\,{\scriptsize[0.45, 0.55]} & 0.11\,{\scriptsize[0.09, 0.14]} & 0.35\,{\scriptsize[0.31, 0.40]} & 0.86\,{\scriptsize[0.80, 0.91]} \\
\cond{Copied minority}: mean $\hat e$ ($e^\ast=$-$1$) & $-$0.52\,{\scriptsize[$-$0.56, $-$0.46]} & $-$0.90\,{\scriptsize[$-$0.92, $-$0.87]} & $-$0.64\,{\scriptsize[$-$0.68, $-$0.60]} & $-$0.13\,{\scriptsize[$-$0.18, $-$0.07]} \\
\cond{Copied minority}: answers with $\hat e>0$ (\%) & 20.3\,{\scriptsize[17.8, 22.8]} & 4.1\,{\scriptsize[2.9, 5.4]} & 15.4\,{\scriptsize[13.2, 17.6]} & 38.9\,{\scriptsize[35.9, 41.9]} \\
\bottomrule
\end{tabular}
\end{adjustbox}
\end{table*}
{\footnotesize\setlength{\tabcolsep}{4pt}
\begin{table*}[!tp]
\centering
\caption{Copy weight and discrimination by reliability and by domain.}\label{app:tab:res-breakdown}
\small\setlength{\tabcolsep}{3pt}
\begin{adjustbox}{max width=\textwidth}
\begin{tabular}{@{}lcccc@{}}
\toprule
& Luna & Sol & Gemini & DeepSeek \\
\midrule
\multicolumn{5}{@{}l}{\textbf{Copy weight $w_3$}} \\
\multicolumn{5}{@{}l}{\emph{Reliability $q$}} \\
\quad $0.60$ & 0.29\,{\scriptsize[0.23, 0.35]} & 0.05\,{\scriptsize[0.02, 0.07]} & 0.25\,{\scriptsize[0.20, 0.31]} & 0.46\,{\scriptsize[0.40, 0.52]} \\
\quad $0.65$ & 0.28\,{\scriptsize[0.22, 0.34]} & 0.05\,{\scriptsize[0.02, 0.07]} & 0.19\,{\scriptsize[0.14, 0.24]} & 0.41\,{\scriptsize[0.35, 0.47]} \\
\quad $0.70$ & 0.29\,{\scriptsize[0.23, 0.35]} & 0.09\,{\scriptsize[0.05, 0.13]} & 0.22\,{\scriptsize[0.17, 0.28]} & 0.48\,{\scriptsize[0.41, 0.55]} \\
\quad $0.75$ & 0.29\,{\scriptsize[0.23, 0.34]} & 0.07\,{\scriptsize[0.04, 0.10]} & 0.23\,{\scriptsize[0.18, 0.29]} & 0.39\,{\scriptsize[0.33, 0.45]} \\
\quad $0.80$ & 0.22\,{\scriptsize[0.17, 0.28]} & 0.04\,{\scriptsize[0.02, 0.06]} & 0.16\,{\scriptsize[0.12, 0.21]} & 0.37\,{\scriptsize[0.30, 0.43]} \\
\addlinespace[3pt]
\multicolumn{5}{@{}l}{\emph{Domain}} \\
\quad Planetary rover & 0.34\,{\scriptsize[0.28, 0.41]} & 0.06\,{\scriptsize[0.03, 0.09]} & 0.21\,{\scriptsize[0.16, 0.26]} & 0.39\,{\scriptsize[0.33, 0.46]} \\
\quad Reservoir ecology & 0.25\,{\scriptsize[0.20, 0.31]} & 0.06\,{\scriptsize[0.04, 0.09]} & 0.17\,{\scriptsize[0.13, 0.22]} & 0.45\,{\scriptsize[0.39, 0.52]} \\
\quad Chemical plant & 0.29\,{\scriptsize[0.23, 0.34]} & 0.05\,{\scriptsize[0.02, 0.08]} & 0.19\,{\scriptsize[0.14, 0.24]} & 0.48\,{\scriptsize[0.41, 0.54]} \\
\quad City air quality & 0.26\,{\scriptsize[0.20, 0.31]} & 0.05\,{\scriptsize[0.03, 0.08]} & 0.22\,{\scriptsize[0.17, 0.28]} & 0.44\,{\scriptsize[0.37, 0.50]} \\
\quad Coastal water & 0.23\,{\scriptsize[0.18, 0.29]} & 0.07\,{\scriptsize[0.04, 0.10]} & 0.26\,{\scriptsize[0.21, 0.32]} & 0.34\,{\scriptsize[0.29, 0.41]} \\
\addlinespace[3pt]
\multicolumn{5}{@{}l}{\textbf{Discrimination $D_3$}} \\
\multicolumn{5}{@{}l}{\emph{Reliability $q$}} \\
\quad $0.60$ & 0.63\,{\scriptsize[0.57, 0.69]} & 0.95\,{\scriptsize[0.92, 0.97]} & 0.73\,{\scriptsize[0.67, 0.78]} & 0.51\,{\scriptsize[0.44, 0.57]} \\
\quad $0.65$ & 0.64\,{\scriptsize[0.58, 0.69]} & 0.93\,{\scriptsize[0.90, 0.95]} & 0.78\,{\scriptsize[0.73, 0.83]} & 0.54\,{\scriptsize[0.47, 0.61]} \\
\quad $0.70$ & 0.66\,{\scriptsize[0.60, 0.73]} & 0.92\,{\scriptsize[0.88, 0.96]} & 0.78\,{\scriptsize[0.72, 0.84]} & 0.51\,{\scriptsize[0.44, 0.58]} \\
\quad $0.75$ & 0.62\,{\scriptsize[0.55, 0.67]} & 0.89\,{\scriptsize[0.86, 0.93]} & 0.72\,{\scriptsize[0.67, 0.78]} & 0.55\,{\scriptsize[0.49, 0.61]} \\
\quad $0.80$ & 0.67\,{\scriptsize[0.61, 0.72]} & 0.91\,{\scriptsize[0.88, 0.93]} & 0.77\,{\scriptsize[0.72, 0.82]} & 0.56\,{\scriptsize[0.50, 0.63]} \\
\addlinespace[3pt]
\multicolumn{5}{@{}l}{\emph{Domain}} \\
\quad Planetary rover & 0.57\,{\scriptsize[0.51, 0.63]} & 0.92\,{\scriptsize[0.89, 0.95]} & 0.75\,{\scriptsize[0.69, 0.81]} & 0.55\,{\scriptsize[0.49, 0.62]} \\
\quad Reservoir ecology & 0.67\,{\scriptsize[0.61, 0.72]} & 0.91\,{\scriptsize[0.88, 0.94]} & 0.80\,{\scriptsize[0.74, 0.84]} & 0.50\,{\scriptsize[0.43, 0.56]} \\
\quad Chemical plant & 0.62\,{\scriptsize[0.56, 0.68]} & 0.93\,{\scriptsize[0.90, 0.96]} & 0.77\,{\scriptsize[0.72, 0.82]} & 0.48\,{\scriptsize[0.41, 0.55]} \\
\quad City air quality & 0.67\,{\scriptsize[0.61, 0.72]} & 0.93\,{\scriptsize[0.89, 0.95]} & 0.75\,{\scriptsize[0.69, 0.80]} & 0.52\,{\scriptsize[0.45, 0.59]} \\
\quad Coastal water & 0.68\,{\scriptsize[0.62, 0.74]} & 0.91\,{\scriptsize[0.88, 0.94]} & 0.72\,{\scriptsize[0.66, 0.77]} & 0.62\,{\scriptsize[0.56, 0.68]} \\
\bottomrule
\end{tabular}
\end{adjustbox}
\end{table*}
\par}
\subsubsection{Full counting and repeated replies (Section 4.3 of the main paper)}
Table~\ref{app:tab:res-modes} shows how the reports on the forms that state one reading several times are distributed between the one-reading posterior, the posterior that counts every statement, and the values in between, and how much of the excess evidence $\bar e-1$ the reports that count every statement carry. Table~\ref{app:tab:res-sat} gives the share of replies at the clipping bound, where the reported evidence reaches $\logit(0.999)/L$, and the results at $m=8$ on the 600 items with $q\le0.70$, where eight readings can be registered. Clipping lowers $\bar e$ on \cond{Independent}$_8$, and with it $D_8$ in Table~\ref{app:tab:res-weights}. Table~\ref{app:tab:res-repl} compares the main sample with two replicate samples on the 500-item subset, and Table~\ref{app:tab:res-switch} follows each \cond{Copies} log across the three samples.

{\footnotesize\setlength{\tabcolsep}{4pt}
\begin{table*}[!tp]
\centering
\caption{Distribution of the reports on forms that state one reading several times: share of reports within 0.15 readings of the one-reading posterior, between, and within 0.15 readings of $m$ readings (every statement counted), and the share of the excess evidence carried by the last group (all in \%).}\label{app:tab:res-modes}
\small\setlength{\tabcolsep}{3pt}
\begin{adjustbox}{max width=\textwidth}
\begin{tabular}{@{}lcccc@{}}
\toprule
Form & one reading & between & $m$ readings & share of excess \\
\midrule
\multicolumn{5}{@{}l}{\textbf{GPT-6 Luna}} \\
\quad \cond{Copies}, 2 statements & 73.2\,{\scriptsize[70.4, 76.0]} & 10.3\,{\scriptsize[8.5, 12.3]} & 16.0\,{\scriptsize[13.8, 18.3]} & 75.7\,{\scriptsize[70.4, 81.1]} \\
\quad \cond{Copies} & 60.7\,{\scriptsize[57.7, 63.7]} & 19.3\,{\scriptsize[16.9, 21.8]} & 19.6\,{\scriptsize[17.2, 22.0]} & 70.8\,{\scriptsize[66.0, 75.3]} \\
\quad \cond{Copies}, 5 statements & 52.3\,{\scriptsize[49.2, 55.5]} & 38.1\,{\scriptsize[35.2, 41.2]} & 7.1\,{\scriptsize[5.5, 8.8]} & 34.9\,{\scriptsize[28.6, 41.0]} \\
\quad \cond{Copies}, 8 statements & 41.0\,{\scriptsize[37.9, 44.1]} & 57.4\,{\scriptsize[54.3, 60.4]} & 1.3\,{\scriptsize[0.7, 2.0]} & 8.9\,{\scriptsize[4.7, 13.5]} \\
\quad \cond{Packed} & 53.6\,{\scriptsize[50.5, 56.7]} & 22.5\,{\scriptsize[19.9, 25.2]} & 23.1\,{\scriptsize[20.5, 25.7]} & 69.1\,{\scriptsize[64.6, 73.3]} \\
\quad \cond{Claims} & 94.1\,{\scriptsize[92.6, 95.5]} & 3.3\,{\scriptsize[2.2, 4.5]} & 2.6\,{\scriptsize[1.7, 3.6]} & 68.6\,{\scriptsize[55.4, 79.6]} \\
\quad \cond{Copies} (prose) & 77.5\,{\scriptsize[75.0, 80.1]} & 10.2\,{\scriptsize[8.3, 12.1]} & 12.0\,{\scriptsize[10.0, 14.0]} & 76.5\,{\scriptsize[70.2, 82.4]} \\
\addlinespace[3pt]
\multicolumn{5}{@{}l}{\textbf{GPT-6.1 Sol}} \\
\quad \cond{Copies}, 2 statements & 92.9\,{\scriptsize[91.2, 94.4]} & 2.5\,{\scriptsize[1.6, 3.5]} & 4.6\,{\scriptsize[3.4, 6.0]} & 80.9\,{\scriptsize[72.1, 88.1]} \\
\quad \cond{Copies} & 90.7\,{\scriptsize[88.8, 92.5]} & 5.1\,{\scriptsize[3.8, 6.5]} & 4.2\,{\scriptsize[3.0, 5.5]} & 72.7\,{\scriptsize[62.9, 80.6]} \\
\quad \cond{Copies}, 5 statements & 88.8\,{\scriptsize[86.9, 90.8]} & 8.3\,{\scriptsize[6.6, 10.0]} & 2.4\,{\scriptsize[1.5, 3.4]} & 47.8\,{\scriptsize[34.9, 59.3]} \\
\quad \cond{Copies}, 8 statements & 85.8\,{\scriptsize[83.6, 87.9]} & 13.0\,{\scriptsize[11.0, 15.1]} & 1.0\,{\scriptsize[0.4, 1.6]} & 23.0\,{\scriptsize[10.8, 34.5]} \\
\quad \cond{Packed} & 93.0\,{\scriptsize[91.4, 94.5]} & 3.2\,{\scriptsize[2.1, 4.3]} & 3.8\,{\scriptsize[2.7, 5.0]} & 78.2\,{\scriptsize[68.2, 86.4]} \\
\quad \cond{Claims} & 99.0\,{\scriptsize[98.3, 99.6]} & 0.6\,{\scriptsize[0.2, 1.1]} & 0.4\,{\scriptsize[0.1, 0.8]} & 75.3\,{\scriptsize[30.9, 93.8]} \\
\quad \cond{Copies} (prose) & 95.7\,{\scriptsize[94.4, 96.9]} & 2.4\,{\scriptsize[1.5, 3.4]} & 1.9\,{\scriptsize[1.1, 2.8]} & 71.8\,{\scriptsize[56.6, 83.1]} \\
\addlinespace[3pt]
\multicolumn{5}{@{}l}{\textbf{Gemini 3.8 Flash}} \\
\quad \cond{Copies}, 2 statements & 78.1\,{\scriptsize[75.5, 80.6]} & 7.0\,{\scriptsize[5.4, 8.6]} & 14.8\,{\scriptsize[12.7, 17.0]} & 81.2\,{\scriptsize[76.4, 85.8]} \\
\quad \cond{Copies} & 72.0\,{\scriptsize[69.2, 74.7]} & 10.1\,{\scriptsize[8.3, 12.0]} & 17.9\,{\scriptsize[15.6, 20.3]} & 83.8\,{\scriptsize[79.8, 87.3]} \\
\quad \cond{Copies}, 5 statements & 64.3\,{\scriptsize[61.3, 67.2]} & 23.3\,{\scriptsize[20.7, 26.0]} & 10.9\,{\scriptsize[9.0, 12.9]} & 58.1\,{\scriptsize[51.9, 63.8]} \\
\quad \cond{Copies}, 8 statements & 55.3\,{\scriptsize[52.3, 58.3]} & 41.4\,{\scriptsize[38.4, 44.5]} & 2.5\,{\scriptsize[1.6, 3.5]} & 18.9\,{\scriptsize[12.5, 25.1]} \\
\quad \cond{Packed} & 76.2\,{\scriptsize[73.6, 78.9]} & 9.2\,{\scriptsize[7.4, 11.0]} & 14.6\,{\scriptsize[12.5, 16.8]} & 81.5\,{\scriptsize[76.7, 85.8]} \\
\quad \cond{Claims} & 97.3\,{\scriptsize[96.3, 98.3]} & 1.4\,{\scriptsize[0.7, 2.2]} & 1.3\,{\scriptsize[0.7, 2.0]} & 72.5\,{\scriptsize[53.3, 86.2]} \\
\quad \cond{Copies} (prose) & 84.3\,{\scriptsize[82.0, 86.5]} & 5.5\,{\scriptsize[4.2, 6.9]} & 10.2\,{\scriptsize[8.3, 12.1]} & 83.4\,{\scriptsize[77.7, 88.2]} \\
\addlinespace[3pt]
\multicolumn{5}{@{}l}{\textbf{DeepSeek V4.1 Flash}} \\
\quad \cond{Copies}, 2 statements & 58.8\,{\scriptsize[55.7, 61.9]} & 11.1\,{\scriptsize[9.2, 13.1]} & 29.7\,{\scriptsize[26.8, 32.6]} & 84.9\,{\scriptsize[81.6, 88.0]} \\
\quad \cond{Copies} & 46.8\,{\scriptsize[43.7, 49.9]} & 17.4\,{\scriptsize[15.1, 19.8]} & 35.4\,{\scriptsize[32.5, 38.4]} & 83.5\,{\scriptsize[80.6, 86.3]} \\
\quad \cond{Copies}, 5 statements & 34.9\,{\scriptsize[31.9, 37.9]} & 46.1\,{\scriptsize[43.1, 49.2]} & 17.0\,{\scriptsize[14.7, 19.4]} & 54.2\,{\scriptsize[49.2, 58.8]} \\
\quad \cond{Copies}, 8 statements & 24.2\,{\scriptsize[21.5, 26.8]} & 71.7\,{\scriptsize[68.9, 74.5]} & 3.3\,{\scriptsize[2.2, 4.5]} & 15.1\,{\scriptsize[10.6, 19.8]} \\
\quad \cond{Packed} & 53.3\,{\scriptsize[50.3, 56.3]} & 15.7\,{\scriptsize[13.5, 18.0]} & 30.6\,{\scriptsize[27.8, 33.5]} & 80.9\,{\scriptsize[77.5, 84.1]} \\
\quad \cond{Claims} & 65.3\,{\scriptsize[62.3, 68.2]} & 11.7\,{\scriptsize[9.8, 13.8]} & 22.6\,{\scriptsize[20.0, 25.2]} & 83.6\,{\scriptsize[79.7, 87.2]} \\
\quad \cond{Copies} (prose) & 65.6\,{\scriptsize[62.6, 68.5]} & 11.2\,{\scriptsize[9.3, 13.2]} & 22.8\,{\scriptsize[20.1, 25.4]} & 82.8\,{\scriptsize[78.4, 86.9]} \\
\bottomrule
\end{tabular}
\end{adjustbox}
\end{table*}
\par}
\begin{table*}[!tp]
\centering
\caption{Clipping of reported probabilities at 0.999, and the dose of eight statements on the items where eight readings can be registered.}\label{app:tab:res-sat}
\small\setlength{\tabcolsep}{4pt}
\begin{adjustbox}{max width=\linewidth}
\begin{tabular}{@{}lcccc@{}}
\toprule
& Luna & Sol & Gemini & DeepSeek \\
\midrule
\multicolumn{5}{@{}l}{\emph{Replies at the clipping bound (\%), all items}} \\
\quad \cond{Copies}, 5 statements & 0.9 & 0.0 & 1.9 & 2.3 \\
\quad \cond{Copies}, 8 statements & 0.8 & 1.1 & 2.2 & 5.6 \\
\quad \cond{Packed}, $m=8$ & 1.8 & 0.7 & 3.1 & 5.2 \\
\quad \cond{Independent}, $m=5$ & 4.8 & 16.2 & 13.1 & 12.6 \\
\quad \cond{Independent}, $m=8$ & 10.3 & 45.6 & 37.5 & 37.4 \\
\addlinespace[3pt]
\multicolumn{5}{@{}l}{\emph{\cond{Copies}, 8 statements, items with $q\le0.70$}} \\
\quad one reading (\%) & 39.7\,{\scriptsize[35.7, 43.5]} & 85.7\,{\scriptsize[82.8, 88.4]} & 55.2\,{\scriptsize[51.2, 59.0]} & 24.0\,{\scriptsize[20.6, 27.4]} \\
\quad between (\%) & 57.7\,{\scriptsize[53.7, 61.6]} & 12.3\,{\scriptsize[9.8, 15.0]} & 39.3\,{\scriptsize[35.5, 43.3]} & 69.2\,{\scriptsize[65.5, 72.9]} \\
\quad eight readings (\%) & 2.2\,{\scriptsize[1.1, 3.4]} & 1.7\,{\scriptsize[0.7, 2.8]} & 4.2\,{\scriptsize[2.6, 5.8]} & 5.5\,{\scriptsize[3.7, 7.4]} \\
\quad copy weight $w_8$ & 0.17\,{\scriptsize[0.15, 0.19]} & 0.05\,{\scriptsize[0.03, 0.06]} & 0.15\,{\scriptsize[0.13, 0.17]} & 0.23\,{\scriptsize[0.21, 0.26]} \\
\quad discrimination $D_8$ & 0.67\,{\scriptsize[0.64, 0.69]} & 0.91\,{\scriptsize[0.90, 0.93]} & 0.79\,{\scriptsize[0.77, 0.82]} & 0.70\,{\scriptsize[0.67, 0.72]} \\
\bottomrule
\end{tabular}
\end{adjustbox}
\end{table*}
\begin{table*}[!tp]
\centering
\caption{Mean $\hat e$ in the main sample and two replicate samples, the spread of the three means, and the mean within-item SD.}\label{app:tab:res-repl}
\small\setlength{\tabcolsep}{4pt}
\begin{adjustbox}{max width=\linewidth}
\begin{tabular}{@{}lccccc@{}}
\toprule
Form & main sample & replicate 1 & replicate 2 & spread & within-item SD \\
\midrule
\multicolumn{6}{@{}l}{\textbf{GPT-6 Luna}} \\
\quad \cond{Single} & 1.00 & 1.00 & 1.00 & 0.004 & 0.007 \\
\quad \cond{Copies} & 1.61 & 1.60 & 1.61 & 0.013 & 0.419 \\
\quad \cond{Independent} & 2.84 & 2.81 & 2.84 & 0.026 & 0.159 \\
\quad \cond{Copied minority} & $-$0.50 & $-$0.52 & $-$0.55 & 0.045 & 0.384 \\
\addlinespace[3pt]
\multicolumn{6}{@{}l}{\textbf{GPT-6.1 Sol}} \\
\quad \cond{Single} & 1.00 & 1.00 & 1.00 & 0.001 & 0.001 \\
\quad \cond{Copies} & 1.15 & 1.14 & 1.12 & 0.035 & 0.145 \\
\quad \cond{Independent} & 2.96 & 2.96 & 2.95 & 0.007 & 0.029 \\
\quad \cond{Copied minority} & $-$0.89 & $-$0.89 & $-$0.91 & 0.026 & 0.121 \\
\addlinespace[3pt]
\multicolumn{6}{@{}l}{\textbf{Gemini 3.8 Flash}} \\
\quad \cond{Single} & 1.00 & 1.00 & 1.00 & 0.002 & 0.002 \\
\quad \cond{Copies} & 1.46 & 1.42 & 1.41 & 0.051 & 0.328 \\
\quad \cond{Independent} & 2.94 & 2.93 & 2.94 & 0.017 & 0.061 \\
\quad \cond{Copied minority} & $-$0.65 & $-$0.61 & $-$0.62 & 0.041 & 0.309 \\
\addlinespace[3pt]
\multicolumn{6}{@{}l}{\textbf{DeepSeek V4.1 Flash}} \\
\quad \cond{Single} & 0.99 & 1.00 & 1.00 & 0.007 & 0.007 \\
\quad \cond{Copies} & 1.81 & 1.83 & 1.79 & 0.044 & 0.477 \\
\quad \cond{Independent} & 2.92 & 2.91 & 2.90 & 0.022 & 0.113 \\
\quad \cond{Copied minority} & $-$0.12 & $-$0.16 & $-$0.18 & 0.065 & 0.481 \\
\bottomrule
\end{tabular}
\end{adjustbox}
\end{table*}
\begin{table*}[!tp]
\centering
\caption{The same \cond{Copies} log in the three samples (500-item subset). A reply over-counts if $\hat e>1+0.15$ and counts in full if $\hat e$ is within 0.15 readings of three readings (all in \%).}\label{app:tab:res-switch}
\footnotesize\setlength{\tabcolsep}{4pt}
\begin{adjustbox}{max width=\linewidth}
\begin{tabular}{@{}lcccccc@{}}
\toprule
 & & \multicolumn{3}{c}{logs over-counted} & \multicolumn{2}{c}{logs counted in full} \\
\cmidrule(lr){3-5}\cmidrule(l){6-7}
Model & over-counting replies & in no reply & in some, not all & in every reply & in at least one reply & in every reply \\
\midrule
GPT-6 Luna & 42.5\,{\scriptsize[38.9, 46.1]} & 39.8\,{\scriptsize[35.4, 44.2]} & 35.4\,{\scriptsize[31.2, 39.6]} & 24.8\,{\scriptsize[21.2, 28.6]} & 39.6\,{\scriptsize[35.2, 44.0]} & 6.4\,{\scriptsize[4.4, 8.6]} \\
GPT-6.1 Sol & 10.7\,{\scriptsize[8.7, 12.9]} & 80.4\,{\scriptsize[76.8, 83.8]} & 16.0\,{\scriptsize[12.8, 19.4]} & 3.6\,{\scriptsize[2.0, 5.4]} & 11.8\,{\scriptsize[9.2, 14.6]} & 1.0\,{\scriptsize[0.2, 2.0]} \\
Gemini 3.8 Flash & 28.1\,{\scriptsize[24.9, 31.2]} & 54.6\,{\scriptsize[50.2, 59.0]} & 32.6\,{\scriptsize[28.4, 36.8]} & 12.8\,{\scriptsize[10.0, 15.8]} & 32.2\,{\scriptsize[28.2, 36.2]} & 6.6\,{\scriptsize[4.6, 8.8]} \\
DeepSeek V4.1 Flash & 50.7\,{\scriptsize[47.3, 54.3]} & 28.2\,{\scriptsize[24.2, 32.2]} & 41.8\,{\scriptsize[37.4, 46.2]} & 30.0\,{\scriptsize[26.0, 34.2]} & 54.2\,{\scriptsize[49.8, 58.6]} & 15.0\,{\scriptsize[12.0, 18.2]} \\
\bottomrule
\end{tabular}
\end{adjustbox}
\end{table*}
\subsubsection{Decisions (Section 4.4 of the main paper)}
Table~\ref{app:tab:res-commit} gives the belief-implied early commitment $P(\hat e>e_b)$ on the forms with one reading, at the boundary of the action channel, $e_b=2$, and at $e_b=1.15$, together with the unnecessary query $P(\hat e\le 2)$ on \cond{Independent}. Figure~\ref{app:fig:res-curve} traces early commitment over boundaries from 1 to 3, and Table~\ref{app:tab:res-median} gives the commitment implied by the median of three replies to the same log. Table~\ref{app:tab:res-actions} gives the emitted actions, the commitment that copies add to them, its agreement with the belief-implied commitment, and the policy control, which states a posterior instead of showing a log.

{\footnotesize\setlength{\tabcolsep}{4pt}
\begin{table*}[!tp]
\centering
\caption{Belief-implied early commitment and unnecessary query on the controlled logs.}\label{app:tab:res-commit}
\small\setlength{\tabcolsep}{3pt}
\begin{adjustbox}{max width=\textwidth}
\begin{tabular}{@{}lcccc@{}}
\toprule
Form & Luna & Sol & Gemini & DeepSeek \\
\midrule
\multicolumn{5}{@{}l}{\emph{Boundary $e_b=2$: $P(\hat e>2)$, \%}} \\
\quad \cond{Single} & 0.0 & 0.0 & 0.0 & 0.0 \\
\quad \cond{Single} (prose) & 0.0 & 0.0 & 0.0 & 0.0 \\
\quad \cond{Pointers} & 0.1\,{\scriptsize[0.0, 0.3]} & 0.0 & 0.0 & 0.3\,{\scriptsize[0.0, 0.7]} \\
\quad \cond{Trace} & 0.0 & 0.1\,{\scriptsize[0.0, 0.3]} & 0.2\,{\scriptsize[0.0, 0.5]} & 0.1\,{\scriptsize[0.0, 0.3]} \\
\quad \cond{Padding} & 0.0 & 0.0 & 0.0 & 0.0 \\
\quad \cond{Padding}, $m=8$ & 0.0 & 0.0 & 0.0 & 0.0 \\
\quad \cond{Copies}, 2 statements & 4.4\,{\scriptsize[3.2, 5.7]} & 0.9\,{\scriptsize[0.4, 1.5]} & 3.2\,{\scriptsize[2.2, 4.3]} & 7.2\,{\scriptsize[5.6, 8.8]} \\
\quad \cond{Copies} & 24.6\,{\scriptsize[22.0, 27.2]} & 4.7\,{\scriptsize[3.4, 6.0]} & 19.1\,{\scriptsize[16.7, 21.5]} & 39.8\,{\scriptsize[36.8, 42.9]} \\
\quad \cond{Copies}, 5 statements & 24.1\,{\scriptsize[21.5, 26.9]} & 6.1\,{\scriptsize[4.7, 7.6]} & 20.9\,{\scriptsize[18.4, 23.4]} & 35.9\,{\scriptsize[32.9, 38.8]} \\
\quad \cond{Copies}, 8 statements & 33.0\,{\scriptsize[30.1, 36.0]} & 8.8\,{\scriptsize[7.1, 10.6]} & 27.0\,{\scriptsize[24.3, 29.8]} & 46.1\,{\scriptsize[43.0, 49.2]} \\
\quad \cond{Copies} (one relay) & 28.3\,{\scriptsize[25.6, 31.1]} & 4.6\,{\scriptsize[3.4, 5.9]} & 16.9\,{\scriptsize[14.6, 19.3]} & 37.1\,{\scriptsize[34.1, 40.1]} \\
\quad \cond{Copies} (no labels) & 31.5\,{\scriptsize[28.7, 34.4]} & 8.5\,{\scriptsize[6.8, 10.2]} & 24.0\,{\scriptsize[21.4, 26.7]} & 46.1\,{\scriptsize[42.9, 49.1]} \\
\quad \cond{Copies} (prose) & 14.2\,{\scriptsize[12.1, 16.4]} & 2.1\,{\scriptsize[1.3, 3.0]} & 11.2\,{\scriptsize[9.2, 13.2]} & 26.4\,{\scriptsize[23.6, 29.2]} \\
\quad \cond{Copies} (prose, no labels) & 18.8\,{\scriptsize[16.4, 21.2]} & 5.7\,{\scriptsize[4.3, 7.2]} & 14.4\,{\scriptsize[12.3, 16.6]} & 33.6\,{\scriptsize[30.7, 36.6]} \\
\quad \cond{Packed} & 29.1\,{\scriptsize[26.3, 32.0]} & 4.1\,{\scriptsize[2.9, 5.3]} & 16.3\,{\scriptsize[14.0, 18.6]} & 35.4\,{\scriptsize[32.5, 38.4]} \\
\quad \cond{Packed}, $m=8$ & 39.6\,{\scriptsize[36.6, 42.7]} & 7.1\,{\scriptsize[5.6, 8.7]} & 24.0\,{\scriptsize[21.4, 26.6]} & 42.1\,{\scriptsize[39.0, 45.1]} \\
\quad \cond{Claims} & 3.0\,{\scriptsize[2.0, 4.1]} & 0.4\,{\scriptsize[0.1, 0.8]} & 1.5\,{\scriptsize[0.8, 2.3]} & 24.8\,{\scriptsize[22.2, 27.5]} \\
\addlinespace[3pt]
\multicolumn{5}{@{}l}{\emph{Boundary $e_b=1.15$: $P(\hat e>1.15)$, \%}} \\
\quad \cond{Single} & 0.2\,{\scriptsize[0.0, 0.5]} & 0.0 & 0.1\,{\scriptsize[0.0, 0.3]} & 0.2\,{\scriptsize[0.0, 0.5]} \\
\quad \cond{Single} (prose) & 0.3\,{\scriptsize[0.0, 0.7]} & 0.1\,{\scriptsize[0.0, 0.3]} & 0.3\,{\scriptsize[0.0, 0.7]} & 0.3\,{\scriptsize[0.0, 0.7]} \\
\quad \cond{Pointers} & 0.9\,{\scriptsize[0.4, 1.5]} & 0.0 & 0.3\,{\scriptsize[0.0, 0.7]} & 1.0\,{\scriptsize[0.4, 1.7]} \\
\quad \cond{Trace} & 1.0\,{\scriptsize[0.4, 1.7]} & 0.3\,{\scriptsize[0.0, 0.7]} & 0.6\,{\scriptsize[0.2, 1.1]} & 0.4\,{\scriptsize[0.1, 0.8]} \\
\quad \cond{Padding} & 0.2\,{\scriptsize[0.0, 0.5]} & 0.0 & 0.1\,{\scriptsize[0.0, 0.3]} & 0.3\,{\scriptsize[0.0, 0.7]} \\
\quad \cond{Padding}, $m=8$ & 0.7\,{\scriptsize[0.2, 1.3]} & 0.1\,{\scriptsize[0.0, 0.3]} & 0.5\,{\scriptsize[0.1, 1.0]} & 1.0\,{\scriptsize[0.4, 1.7]} \\
\quad \cond{Copies}, 2 statements & 26.7\,{\scriptsize[23.9, 29.5]} & 7.1\,{\scriptsize[5.6, 8.8]} & 21.9\,{\scriptsize[19.4, 24.5]} & 41.1\,{\scriptsize[38.0, 44.1]} \\
\quad \cond{Copies} & 39.3\,{\scriptsize[36.3, 42.3]} & 9.3\,{\scriptsize[7.5, 11.2]} & 28.0\,{\scriptsize[25.3, 30.8]} & 53.1\,{\scriptsize[50.0, 56.2]} \\
\quad \cond{Copies}, 5 statements & 47.6\,{\scriptsize[44.5, 50.8]} & 11.2\,{\scriptsize[9.2, 13.1]} & 35.7\,{\scriptsize[32.8, 38.7]} & 65.0\,{\scriptsize[62.0, 68.0]} \\
\quad \cond{Copies}, 8 statements & 59.0\,{\scriptsize[55.9, 62.1]} & 14.2\,{\scriptsize[12.1, 16.4]} & 44.6\,{\scriptsize[41.6, 47.6]} & 75.8\,{\scriptsize[73.2, 78.5]} \\
\quad \cond{Copies} (one relay) & 43.1\,{\scriptsize[40.1, 46.2]} & 9.0\,{\scriptsize[7.3, 10.8]} & 25.3\,{\scriptsize[22.7, 28.1]} & 49.0\,{\scriptsize[45.9, 52.1]} \\
\quad \cond{Copies} (no labels) & 47.5\,{\scriptsize[44.4, 50.6]} & 15.1\,{\scriptsize[12.9, 17.3]} & 35.7\,{\scriptsize[32.8, 38.7]} & 60.6\,{\scriptsize[57.6, 63.6]} \\
\bottomrule
\end{tabular}
\end{adjustbox}
\end{table*}

\begin{table*}[!tp]
\centering
\ContinuedFloat
\caption{Belief-implied early commitment and unnecessary query on the controlled logs. (continued).}
\small\setlength{\tabcolsep}{3pt}
\begin{adjustbox}{max width=\textwidth}
\begin{tabular}{@{}lcccc@{}}
\toprule
Form & Luna & Sol & Gemini & DeepSeek \\
\midrule
\multicolumn{5}{@{}l}{\emph{Boundary $e_b=1.15$: $P(\hat e>1.15)$, \%}} \\
\quad \cond{Copies} (prose) & 22.3\,{\scriptsize[19.7, 24.8]} & 4.3\,{\scriptsize[3.1, 5.6]} & 15.7\,{\scriptsize[13.5, 18.0]} & 34.1\,{\scriptsize[31.2, 37.1]} \\
\quad \cond{Copies} (prose, no labels) & 29.5\,{\scriptsize[26.8, 32.3]} & 11.0\,{\scriptsize[9.1, 13.0]} & 22.1\,{\scriptsize[19.6, 24.7]} & 43.0\,{\scriptsize[40.0, 46.1]} \\
\quad \cond{Packed} & 46.4\,{\scriptsize[43.3, 49.5]} & 7.0\,{\scriptsize[5.5, 8.6]} & 23.8\,{\scriptsize[21.1, 26.4]} & 46.7\,{\scriptsize[43.7, 49.7]} \\
\quad \cond{Packed}, $m=8$ & 66.8\,{\scriptsize[63.8, 69.7]} & 10.8\,{\scriptsize[8.9, 12.8]} & 40.0\,{\scriptsize[37.0, 43.0]} & 71.1\,{\scriptsize[68.2, 73.9]} \\
\quad \cond{Claims} & 5.9\,{\scriptsize[4.5, 7.4]} & 1.0\,{\scriptsize[0.4, 1.7]} & 2.7\,{\scriptsize[1.7, 3.7]} & 34.5\,{\scriptsize[31.6, 37.4]} \\
\addlinespace[3pt]
\multicolumn{5}{@{}l}{\emph{Unnecessary query: $P(\hat e\le 2)$, \%}} \\
\quad \cond{Independent} & 3.9\,{\scriptsize[2.8, 5.1]} & 0.3\,{\scriptsize[0.0, 0.7]} & 0.9\,{\scriptsize[0.4, 1.5]} & 3.6\,{\scriptsize[2.5, 4.8]} \\
\bottomrule
\end{tabular}
\end{adjustbox}
\end{table*}
\par}
\begin{figure*}[!tp]
\centering
\suppplot{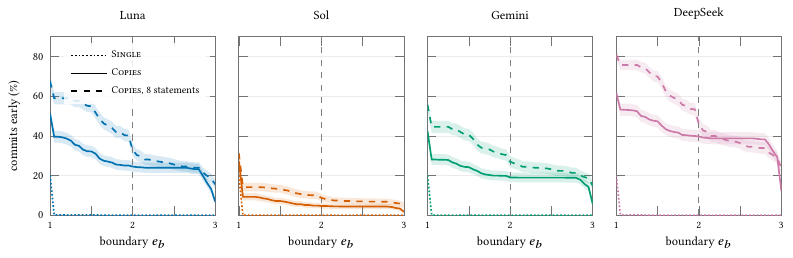}
\caption{Belief-implied early commitment $P(\hat e>e_b)$ against the boundary $e_b$, with 95\% intervals. The vertical line marks the action-channel boundary $e_b=2$.}\label{app:fig:res-curve}
\Description{Four line charts, one per model, each with three lines for one reading stated once, three times and eight times. The share of logs above the boundary falls as the boundary rises, and the lines for restated readings lie above the line for the single report.}
\end{figure*}
\begin{table*}[!tp]
\centering
\caption{Commitment implied by one reply and by the median of the three replies to the same log (500-item subset, $\hat e>2$, in \%).}\label{app:tab:res-median}
\footnotesize\setlength{\tabcolsep}{4pt}
\begin{adjustbox}{max width=\linewidth}
\begin{tabular}{@{}lccccc@{}}
\toprule
 & \multicolumn{3}{c}{\cond{Copies}: early commitment} & \multicolumn{2}{c}{\cond{Independent}: commitment} \\
\cmidrule(lr){2-4}\cmidrule(l){5-6}
Model & one reply & median of three & all three replies & one reply & median of three \\
\midrule
GPT-6 Luna & 27.4\,{\scriptsize[23.4, 31.4]} & 24.0\,{\scriptsize[20.2, 27.8]} & 10.4\,{\scriptsize[7.8, 13.2]} & 96.4\,{\scriptsize[94.6, 98.0]} & 98.8\,{\scriptsize[97.8, 99.6]} \\
GPT-6.1 Sol & 6.4\,{\scriptsize[4.4, 8.6]} & 3.4\,{\scriptsize[2.0, 5.0]} & 1.0\,{\scriptsize[0.2, 2.0]} & 100.0\,{\scriptsize[100.0, 100.0]} & 100.0\,{\scriptsize[100.0, 100.0]} \\
Gemini 3.8 Flash & 20.8\,{\scriptsize[17.4, 24.4]} & 16.6\,{\scriptsize[13.4, 20.0]} & 7.4\,{\scriptsize[5.2, 9.8]} & 99.8\,{\scriptsize[99.4, 100.0]} & 99.6\,{\scriptsize[99.0, 100.0]} \\
DeepSeek V4.1 Flash & 37.8\,{\scriptsize[33.6, 42.2]} & 36.4\,{\scriptsize[32.2, 40.6]} & 18.0\,{\scriptsize[14.8, 21.6]} & 96.8\,{\scriptsize[95.2, 98.2]} & 99.0\,{\scriptsize[98.0, 99.8]} \\
\bottomrule
\end{tabular}
\end{adjustbox}
\end{table*}
\begin{table*}[!tp]
\centering
\caption{Actions on the 500-item subset. Commitment is the share of emitted commit actions. The oracle queries on \cond{Single}, \cond{Copies} and \cond{Copied minority} and commits on \cond{Independent}. The commitment added by copies is \cond{Copies} minus \cond{Single}, paired by item, and the belief-implied commitment is the share of the same items whose belief on \cond{Copies} exceeds the boundary. A wrong emitted action arises at the action stage when the action implied by the same model's belief is correct. The policy control states the posterior as a number instead of showing a log (seven stated posteriors per reliability, Section~\ref{app:sec:t1prompts}).}\label{app:tab:res-actions}
\small\setlength{\tabcolsep}{3pt}
\begin{adjustbox}{max width=\linewidth}
\begin{tabular}{@{}lcccc@{}}
\toprule
& Luna & Sol & Gemini & DeepSeek \\
\midrule
\multicolumn{5}{@{}l}{\emph{Commitment (\%), no note}} \\
\quad \cond{Single} & 2.4\,{\scriptsize[1.2, 3.8]} & 1.0\,{\scriptsize[0.2, 2.0]} & 6.2\,{\scriptsize[4.2, 8.4]} & 35.8\,{\scriptsize[31.6, 40.0]} \\
\quad \cond{Copies} & 24.2\,{\scriptsize[20.4, 28.0]} & 6.6\,{\scriptsize[4.6, 8.8]} & 30.4\,{\scriptsize[26.4, 34.6]} & 70.4\,{\scriptsize[66.2, 74.4]} \\
\quad \cond{Independent} & 89.4\,{\scriptsize[86.6, 92.0]} & 98.8\,{\scriptsize[97.8, 99.6]} & 99.2\,{\scriptsize[98.4, 99.8]} & 98.4\,{\scriptsize[97.2, 99.4]} \\
\quad \cond{Copied minority} & 8.0\,{\scriptsize[5.6, 10.4]} & 3.2\,{\scriptsize[1.8, 4.8]} & 21.6\,{\scriptsize[18.0, 25.2]} & 53.4\,{\scriptsize[49.2, 57.8]} \\
\addlinespace[3pt]
\multicolumn{5}{@{}l}{\emph{Commitment (\%), \arm{Declaration: records}}} \\
\quad \cond{Single} & 2.2\,{\scriptsize[1.0, 3.6]} & 0.6\,{\scriptsize[0.0, 1.4]} & 7.4\,{\scriptsize[5.2, 9.8]} & 35.0\,{\scriptsize[31.0, 39.2]} \\
\quad \cond{Copies} & 2.4\,{\scriptsize[1.2, 3.8]} & 1.0\,{\scriptsize[0.2, 2.0]} & 6.8\,{\scriptsize[4.8, 9.0]} & 45.6\,{\scriptsize[41.4, 49.8]} \\
\quad \cond{Independent} & 88.6\,{\scriptsize[85.8, 91.2]} & 99.0\,{\scriptsize[98.0, 99.8]} & 98.4\,{\scriptsize[97.2, 99.4]} & 98.0\,{\scriptsize[96.6, 99.2]} \\
\quad \cond{Copied minority} & 9.2\,{\scriptsize[6.8, 11.8]} & 4.4\,{\scriptsize[2.8, 6.2]} & 22.4\,{\scriptsize[18.8, 26.0]} & 51.9\,{\scriptsize[47.6, 56.2]} \\
\addlinespace[3pt]
\multicolumn{5}{@{}l}{\emph{\cond{Copies} against belief, no note}} \\
\quad commitment added by copies (points) & 21.8\,{\scriptsize[17.8, 25.8]} & 5.6\,{\scriptsize[3.2, 8.0]} & 24.2\,{\scriptsize[19.8, 28.8]} & 34.6\,{\scriptsize[28.6, 40.4]} \\
\quad belief-implied commitment (\%) & 27.4\,{\scriptsize[23.6, 31.4]} & 6.4\,{\scriptsize[4.4, 8.6]} & 20.8\,{\scriptsize[17.4, 24.4]} & 37.8\,{\scriptsize[33.6, 42.0]} \\
\quad difference (points) & $-$5.6\,{\scriptsize[$-$10.4, $-$1.0]} & $-$0.8\,{\scriptsize[$-$3.4, 1.8]} & 3.4\,{\scriptsize[$-$1.8, 8.4]} & $-$3.2\,{\scriptsize[$-$9.8, 3.4]} \\
\addlinespace[3pt]
\multicolumn{5}{@{}l}{\emph{Where wrong actions arise (\%)}} \\
\quad action-stage share of wrong actions, no note & 70.8 & 80.0 & 79.3 & 80.4 \\
\quad action-stage share of wrong actions, \arm{Declaration: records} & 96.8 & 100.0 & 100.0 & 97.9 \\
\quad policy control, query at the one-reading posterior & 100.0 & 99.0 & 95.0 & 70.0 \\
\quad policy control, accuracy over the seven stated posteriors & 87.0 & 93.6 & 84.0 & 69.3 \\
\bottomrule
\end{tabular}
\end{adjustbox}
\end{table*}
\subsubsection{Notes and reasoning effort (Section 5.1 of the main paper)}
Tables~\ref{app:tab:res-notes} to~\ref{app:tab:res-notes-cp} report the notes on the 500-item subset: the paired change in $\hat e$ against no note, the copy weight $w_3$ and the share of \cond{Copied minority} answers that side with the copied reading ($\hat e>0$) under each note, and three notes on \cond{Claims} and \cond{Packed}. Table~\ref{app:tab:res-reasoning} gives the reasoning settings on the same subset. Sol and Gemini cannot switch reasoning off.

{\footnotesize\setlength{\tabcolsep}{4pt}
\begin{table*}[!tp]
\centering
\caption{Paired change in mean $\hat e$ against no note on the 500-item subset.}\label{app:tab:res-notes}
\small\setlength{\tabcolsep}{3pt}
\begin{adjustbox}{max width=\textwidth}
\begin{tabular}{@{}lcccc@{}}
\toprule
Note & \cond{Single} & \cond{Copies} & \cond{Independent} & \cond{Copied minority} \\
\midrule
\multicolumn{5}{@{}l}{\textbf{GPT-6 Luna}} \\
\quad No note (mean $\hat e$) & 1.00\,{\scriptsize[1.00, 1.01]} & 1.61\,{\scriptsize[1.54, 1.68]} & 2.84\,{\scriptsize[2.81, 2.86]} & $-$0.50\,{\scriptsize[$-$0.57, $-$0.43]} \\
\quad Field guide & 0.00\,{\scriptsize[0.00, 0.01]} & $-$0.17\,{\scriptsize[$-$0.26, $-$0.08]} & $-$0.01\,{\scriptsize[$-$0.05, 0.03]} & $-$0.10\,{\scriptsize[$-$0.18, $-$0.01]} \\
\quad Declaration: records & 0.00\,{\scriptsize[0.00, 0.00]} & $-$0.60\,{\scriptsize[$-$0.68, $-$0.53]} & 0.01\,{\scriptsize[$-$0.02, 0.04]} & $-$0.47\,{\scriptsize[$-$0.54, $-$0.40]} \\
\quad Declaration: relays & 0.00\,{\scriptsize[0.00, 0.00]} & $-$0.56\,{\scriptsize[$-$0.64, $-$0.49]} & 0.01\,{\scriptsize[$-$0.03, 0.05]} & $-$0.46\,{\scriptsize[$-$0.54, $-$0.40]} \\
\quad Declaration: copies & 0.00\,{\scriptsize[$-$0.01, 0.00]} & $-$0.55\,{\scriptsize[$-$0.62, $-$0.48]} & 0.02\,{\scriptsize[$-$0.03, 0.05]} & $-$0.45\,{\scriptsize[$-$0.53, $-$0.38]} \\
\quad Declaration and instruction & 0.00\,{\scriptsize[$-$0.01, 0.00]} & $-$0.60\,{\scriptsize[$-$0.68, $-$0.53]} & 0.03\,{\scriptsize[0.00, 0.07]} & $-$0.49\,{\scriptsize[$-$0.56, $-$0.42]} \\
\quad Counting instruction & 0.00\,{\scriptsize[$-$0.01, 0.00]} & $-$0.46\,{\scriptsize[$-$0.54, $-$0.37]} & 0.03\,{\scriptsize[$-$0.01, 0.07]} & $-$0.34\,{\scriptsize[$-$0.42, $-$0.27]} \\
\quad Placebo: display & $-$0.01\,{\scriptsize[$-$0.02, 0.00]} & 0.28\,{\scriptsize[0.18, 0.37]} & 0.01\,{\scriptsize[$-$0.03, 0.05]} & 0.38\,{\scriptsize[0.29, 0.48]} \\
\quad Placebo: delivery & 0.00\,{\scriptsize[0.00, 0.00]} & 0.13\,{\scriptsize[0.03, 0.23]} & $-$0.01\,{\scriptsize[$-$0.05, 0.03]} & 0.21\,{\scriptsize[0.12, 0.31]} \\
\quad Placebo: unrelated & $-$0.01\,{\scriptsize[$-$0.02, 0.00]} & $-$0.05\,{\scriptsize[$-$0.14, 0.04]} & $-$0.01\,{\scriptsize[$-$0.04, 0.03]} & $-$0.01\,{\scriptsize[$-$0.10, 0.09]} \\
\addlinespace[3pt]
\multicolumn{5}{@{}l}{\textbf{GPT-6.1 Sol}} \\
\quad No note (mean $\hat e$) & 1.00\,{\scriptsize[1.00, 1.00]} & 1.15\,{\scriptsize[1.11, 1.20]} & 2.96\,{\scriptsize[2.95, 2.97]} & $-$0.89\,{\scriptsize[$-$0.92, $-$0.85]} \\
\quad Field guide & 0.00 & $-$0.07\,{\scriptsize[$-$0.12, $-$0.02]} & 0.00\,{\scriptsize[$-$0.01, 0.01]} & $-$0.06\,{\scriptsize[$-$0.10, $-$0.01]} \\
\quad Declaration: records & 0.00 & $-$0.15\,{\scriptsize[$-$0.19, $-$0.11]} & $-$0.01\,{\scriptsize[$-$0.02, 0.00]} & $-$0.11\,{\scriptsize[$-$0.14, $-$0.07]} \\
\quad Declaration: relays & 0.00 & $-$0.15\,{\scriptsize[$-$0.19, $-$0.11]} & $-$0.01\,{\scriptsize[$-$0.02, 0.00]} & $-$0.11\,{\scriptsize[$-$0.15, $-$0.07]} \\
\quad Declaration: copies & 0.00 & $-$0.15\,{\scriptsize[$-$0.19, $-$0.11]} & $-$0.01\,{\scriptsize[$-$0.02, 0.00]} & $-$0.11\,{\scriptsize[$-$0.15, $-$0.07]} \\
\quad Declaration and instruction & 0.00 & $-$0.15\,{\scriptsize[$-$0.20, $-$0.11]} & 0.00\,{\scriptsize[$-$0.01, 0.01]} & $-$0.11\,{\scriptsize[$-$0.15, $-$0.08]} \\
\quad Counting instruction & 0.00 & $-$0.13\,{\scriptsize[$-$0.17, $-$0.08]} & $-$0.01\,{\scriptsize[$-$0.02, 0.00]} & $-$0.09\,{\scriptsize[$-$0.13, $-$0.05]} \\
\quad Placebo: display & 0.00 & $-$0.02\,{\scriptsize[$-$0.07, 0.04]} & $-$0.01\,{\scriptsize[$-$0.02, 0.00]} & 0.05\,{\scriptsize[$-$0.01, 0.10]} \\
\quad Placebo: delivery & 0.00 & 0.00\,{\scriptsize[$-$0.06, 0.05]} & $-$0.01\,{\scriptsize[$-$0.02, 0.00]} & 0.02\,{\scriptsize[$-$0.03, 0.07]} \\
\quad Placebo: unrelated & 0.00 & $-$0.03\,{\scriptsize[$-$0.09, 0.02]} & 0.00\,{\scriptsize[$-$0.01, 0.01]} & $-$0.02\,{\scriptsize[$-$0.07, 0.03]} \\
\addlinespace[3pt]
\multicolumn{5}{@{}l}{\textbf{Gemini 3.8 Flash}} \\
\quad No note (mean $\hat e$) & 1.00\,{\scriptsize[1.00, 1.00]} & 1.46\,{\scriptsize[1.39, 1.53]} & 2.94\,{\scriptsize[2.93, 2.95]} & $-$0.65\,{\scriptsize[$-$0.71, $-$0.59]} \\
\quad Field guide & 0.00\,{\scriptsize[0.00, 0.00]} & $-$0.12\,{\scriptsize[$-$0.21, $-$0.04]} & 0.00\,{\scriptsize[$-$0.02, 0.02]} & $-$0.06\,{\scriptsize[$-$0.14, 0.01]} \\
\quad Declaration: records & 0.00\,{\scriptsize[0.00, 0.00]} & $-$0.44\,{\scriptsize[$-$0.51, $-$0.38]} & 0.00\,{\scriptsize[$-$0.02, 0.02]} & $-$0.33\,{\scriptsize[$-$0.40, $-$0.27]} \\
\quad Declaration: relays & 0.00\,{\scriptsize[0.00, 0.00]} & $-$0.44\,{\scriptsize[$-$0.51, $-$0.37]} & $-$0.01\,{\scriptsize[$-$0.03, 0.01]} & $-$0.34\,{\scriptsize[$-$0.40, $-$0.27]} \\
\quad Declaration: copies & $-$0.01\,{\scriptsize[$-$0.02, 0.00]} & $-$0.45\,{\scriptsize[$-$0.52, $-$0.38]} & $-$0.01\,{\scriptsize[$-$0.02, 0.01]} & $-$0.35\,{\scriptsize[$-$0.41, $-$0.28]} \\
\quad Declaration and instruction & 0.00\,{\scriptsize[$-$0.01, 0.00]} & $-$0.45\,{\scriptsize[$-$0.52, $-$0.39]} & $-$0.01\,{\scriptsize[$-$0.03, 0.01]} & $-$0.34\,{\scriptsize[$-$0.41, $-$0.28]} \\
\quad Counting instruction & 0.00\,{\scriptsize[0.00, 0.00]} & $-$0.36\,{\scriptsize[$-$0.43, $-$0.29]} & 0.01\,{\scriptsize[$-$0.01, 0.02]} & $-$0.29\,{\scriptsize[$-$0.36, $-$0.23]} \\
\quad Placebo: display & 0.00\,{\scriptsize[0.00, 0.00]} & 0.39\,{\scriptsize[0.29, 0.48]} & 0.01\,{\scriptsize[$-$0.01, 0.02]} & 0.40\,{\scriptsize[0.30, 0.49]} \\
\quad Placebo: delivery & 0.00\,{\scriptsize[$-$0.01, 0.00]} & 0.18\,{\scriptsize[0.08, 0.27]} & 0.00\,{\scriptsize[$-$0.02, 0.02]} & 0.21\,{\scriptsize[0.12, 0.30]} \\
\bottomrule
\end{tabular}
\end{adjustbox}
\end{table*}

\begin{table*}[!tp]
\centering
\ContinuedFloat
\caption{Paired change in mean $\hat e$ against no note on the 500-item subset. (continued).}
\small\setlength{\tabcolsep}{3pt}
\begin{adjustbox}{max width=\textwidth}
\begin{tabular}{@{}lcccc@{}}
\toprule
Note & \cond{Single} & \cond{Copies} & \cond{Independent} & \cond{Copied minority} \\
\midrule
\multicolumn{5}{@{}l}{\textbf{Gemini 3.8 Flash}} \\
\quad Placebo: unrelated & 0.00\,{\scriptsize[0.00, 0.00]} & $-$0.05\,{\scriptsize[$-$0.14, 0.04]} & 0.00\,{\scriptsize[$-$0.02, 0.02]} & 0.03\,{\scriptsize[$-$0.05, 0.12]} \\
\addlinespace[3pt]
\multicolumn{5}{@{}l}{\textbf{DeepSeek V4.1 Flash}} \\
\quad No note (mean $\hat e$) & 0.99\,{\scriptsize[0.98, 1.00]} & 1.81\,{\scriptsize[1.72, 1.89]} & 2.92\,{\scriptsize[2.90, 2.94]} & $-$0.12\,{\scriptsize[$-$0.20, $-$0.03]} \\
\quad Field guide & 0.00\,{\scriptsize[$-$0.01, 0.02]} & $-$0.07\,{\scriptsize[$-$0.18, 0.04]} & $-$0.02\,{\scriptsize[$-$0.04, 0.01]} & $-$0.17\,{\scriptsize[$-$0.27, $-$0.07]} \\
\quad Declaration: records & 0.00\,{\scriptsize[$-$0.01, 0.02]} & $-$0.65\,{\scriptsize[$-$0.73, $-$0.57]} & $-$0.06\,{\scriptsize[$-$0.10, $-$0.03]} & $-$0.77\,{\scriptsize[$-$0.85, $-$0.69]} \\
\quad Declaration: relays & 0.01\,{\scriptsize[0.00, 0.02]} & $-$0.64\,{\scriptsize[$-$0.72, $-$0.55]} & $-$0.02\,{\scriptsize[$-$0.05, 0.01]} & $-$0.71\,{\scriptsize[$-$0.80, $-$0.63]} \\
\quad Declaration: copies & 0.00\,{\scriptsize[$-$0.01, 0.02]} & $-$0.63\,{\scriptsize[$-$0.71, $-$0.55]} & $-$0.02\,{\scriptsize[$-$0.05, 0.01]} & $-$0.74\,{\scriptsize[$-$0.82, $-$0.66]} \\
\quad Declaration and instruction & 0.00\,{\scriptsize[$-$0.01, 0.02]} & $-$0.74\,{\scriptsize[$-$0.82, $-$0.66]} & $-$0.06\,{\scriptsize[$-$0.10, $-$0.02]} & $-$0.84\,{\scriptsize[$-$0.92, $-$0.75]} \\
\quad Counting instruction & 0.01\,{\scriptsize[0.00, 0.02]} & $-$0.45\,{\scriptsize[$-$0.53, $-$0.36]} & $-$0.01\,{\scriptsize[$-$0.03, 0.02]} & $-$0.57\,{\scriptsize[$-$0.66, $-$0.48]} \\
\quad Placebo: display & 0.01\,{\scriptsize[0.00, 0.02]} & 0.27\,{\scriptsize[0.16, 0.38]} & $-$0.03\,{\scriptsize[$-$0.05, 0.00]} & 0.16\,{\scriptsize[0.05, 0.27]} \\
\quad Placebo: delivery & 0.01\,{\scriptsize[0.00, 0.02]} & 0.13\,{\scriptsize[0.03, 0.24]} & $-$0.04\,{\scriptsize[$-$0.07, 0.00]} & 0.00\,{\scriptsize[$-$0.11, 0.10]} \\
\quad Placebo: unrelated & 0.00\,{\scriptsize[$-$0.02, 0.02]} & 0.01\,{\scriptsize[$-$0.09, 0.11]} & $-$0.02\,{\scriptsize[$-$0.05, 0.01]} & $-$0.11\,{\scriptsize[$-$0.22, $-$0.01]} \\
\bottomrule
\end{tabular}
\end{adjustbox}
\end{table*}
\par}
\begin{table*}[!tp]
\centering
\caption{Copy weight and side-with-copy share under each note on the 500-item subset.}\label{app:tab:res-notes-w}
\small\setlength{\tabcolsep}{4pt}
\begin{adjustbox}{max width=\linewidth}
\begin{tabular}{@{}lcccc@{}}
\toprule
& Luna & Sol & Gemini & DeepSeek \\
\midrule
\multicolumn{5}{@{}l}{\emph{Copy weight $w_3$}} \\
\quad No note & 0.30\,{\scriptsize[0.27, 0.34]} & 0.08\,{\scriptsize[0.06, 0.10]} & 0.23\,{\scriptsize[0.20, 0.26]} & 0.40\,{\scriptsize[0.36, 0.44]} \\
\quad Field guide & 0.22\,{\scriptsize[0.19, 0.25]} & 0.04\,{\scriptsize[0.03, 0.06]} & 0.17\,{\scriptsize[0.14, 0.20]} & 0.37\,{\scriptsize[0.33, 0.41]} \\
\quad Declaration: records & 0.00\,{\scriptsize[0.00, 0.01]} & 0.00\,{\scriptsize[0.00, 0.00]} & 0.01\,{\scriptsize[0.00, 0.02]} & 0.08\,{\scriptsize[0.06, 0.10]} \\
\quad Declaration: relays & 0.02\,{\scriptsize[0.01, 0.03]} & 0.00\,{\scriptsize[0.00, 0.01]} & 0.01\,{\scriptsize[0.00, 0.02]} & 0.08\,{\scriptsize[0.07, 0.10]} \\
\quad Declaration: copies & 0.03\,{\scriptsize[0.02, 0.04]} & 0.00\,{\scriptsize[0.00, 0.00]} & 0.01\,{\scriptsize[0.00, 0.01]} & 0.09\,{\scriptsize[0.07, 0.11]} \\
\quad Declaration and instruction & 0.00\,{\scriptsize[0.00, 0.01]} & 0.00\,{\scriptsize[0.00, 0.00]} & 0.00\,{\scriptsize[0.00, 0.01]} & 0.03\,{\scriptsize[0.02, 0.05]} \\
\quad Counting instruction & 0.08\,{\scriptsize[0.06, 0.10]} & 0.01\,{\scriptsize[0.01, 0.02]} & 0.05\,{\scriptsize[0.04, 0.07]} & 0.18\,{\scriptsize[0.15, 0.21]} \\
\quad Placebo: display & 0.44\,{\scriptsize[0.40, 0.48]} & 0.07\,{\scriptsize[0.05, 0.09]} & 0.42\,{\scriptsize[0.38, 0.46]} & 0.54\,{\scriptsize[0.49, 0.58]} \\
\quad Placebo: delivery & 0.37\,{\scriptsize[0.33, 0.41]} & 0.07\,{\scriptsize[0.05, 0.10]} & 0.32\,{\scriptsize[0.28, 0.36]} & 0.47\,{\scriptsize[0.43, 0.51]} \\
\quad Placebo: unrelated & 0.28\,{\scriptsize[0.24, 0.31]} & 0.06\,{\scriptsize[0.04, 0.08]} & 0.21\,{\scriptsize[0.17, 0.24]} & 0.41\,{\scriptsize[0.37, 0.45]} \\
\addlinespace[3pt]
\multicolumn{5}{@{}l}{\emph{\cond{Copied minority}: answers with $\hat e>0$ (\%)}} \\
\quad No note & 21.0\,{\scriptsize[17.4, 24.6]} & 4.2\,{\scriptsize[2.6, 6.0]} & 15.2\,{\scriptsize[12.2, 18.4]} & 39.6\,{\scriptsize[35.4, 44.0]} \\
\quad Field guide & 16.2\,{\scriptsize[13.0, 19.4]} & 1.8\,{\scriptsize[0.8, 3.0]} & 12.4\,{\scriptsize[9.6, 15.4]} & 31.2\,{\scriptsize[27.2, 35.2]} \\
\quad Declaration: records & 1.0\,{\scriptsize[0.2, 2.0]} & 0.2\,{\scriptsize[0.0, 0.6]} & 0.6\,{\scriptsize[0.0, 1.4]} & 3.0\,{\scriptsize[1.6, 4.6]} \\
\quad Declaration: relays & 1.0\,{\scriptsize[0.2, 2.0]} & 0.2\,{\scriptsize[0.0, 0.6]} & 0.4\,{\scriptsize[0.0, 1.0]} & 4.8\,{\scriptsize[3.0, 6.8]} \\
\quad Declaration: copies & 1.6\,{\scriptsize[0.6, 2.8]} & 0.0 & 0.0 & 3.8\,{\scriptsize[2.2, 5.6]} \\
\quad Declaration and instruction & 0.0 & 0.0 & 0.0 & 1.4\,{\scriptsize[0.4, 2.6]} \\
\quad Counting instruction & 5.2\,{\scriptsize[3.4, 7.2]} & 0.6\,{\scriptsize[0.0, 1.4]} & 1.4\,{\scriptsize[0.4, 2.6]} & 10.4\,{\scriptsize[7.8, 13.2]} \\
\quad Placebo: display & 38.0\,{\scriptsize[33.8, 42.4]} & 6.6\,{\scriptsize[4.4, 8.8]} & 31.6\,{\scriptsize[27.4, 35.8]} & 46.5\,{\scriptsize[42.2, 50.8]} \\
\quad Placebo: delivery & 31.1\,{\scriptsize[27.1, 35.2]} & 5.2\,{\scriptsize[3.4, 7.2]} & 24.6\,{\scriptsize[21.0, 28.4]} & 39.2\,{\scriptsize[35.0, 43.6]} \\
\quad Placebo: unrelated & 21.0\,{\scriptsize[17.4, 24.6]} & 3.6\,{\scriptsize[2.0, 5.4]} & 16.2\,{\scriptsize[13.2, 19.6]} & 33.7\,{\scriptsize[29.5, 37.8]} \\
\bottomrule
\end{tabular}
\end{adjustbox}
\end{table*}
\begin{table*}[!tp]
\centering
\caption{Three notes on \cond{Claims} and \cond{Packed}, mean $\hat e$ and paired change against no note.}\label{app:tab:res-notes-cp}
\small\setlength{\tabcolsep}{4pt}
\begin{adjustbox}{max width=\linewidth}
\begin{tabular}{@{}lcccc@{}}
\toprule
& Luna & Sol & Gemini & DeepSeek \\
\midrule
\multicolumn{5}{@{}l}{\emph{\cond{Claims}}} \\
\quad No note, mean $\hat e$ & 1.07\,{\scriptsize[1.05, 1.11]} & 1.01\,{\scriptsize[1.00, 1.03]} & 1.03\,{\scriptsize[1.02, 1.06]} & 1.51\,{\scriptsize[1.44, 1.58]} \\
\quad Field guide, mean $\hat e$ & 1.07\,{\scriptsize[1.04, 1.10]} & 1.02\,{\scriptsize[1.00, 1.03]} & 1.04\,{\scriptsize[1.02, 1.07]} & 1.41\,{\scriptsize[1.34, 1.47]} \\
\quad Declaration: records, mean $\hat e$ & 1.00\,{\scriptsize[1.00, 1.01]} & 1.00\,{\scriptsize[1.00, 1.01]} & 1.00\,{\scriptsize[1.00, 1.00]} & 1.11\,{\scriptsize[1.08, 1.15]} \\
\quad Placebo: display, mean $\hat e$ & 1.15\,{\scriptsize[1.11, 1.20]} & 1.03\,{\scriptsize[1.01, 1.05]} & 1.13\,{\scriptsize[1.09, 1.17]} & 1.74\,{\scriptsize[1.66, 1.82]} \\
\quad Field guide $-$ no note & 0.00\,{\scriptsize[$-$0.04, 0.04]} & 0.00\,{\scriptsize[$-$0.02, 0.02]} & 0.01\,{\scriptsize[$-$0.02, 0.04]} & $-$0.10\,{\scriptsize[$-$0.19, $-$0.02]} \\
\quad Declaration: records $-$ no note & $-$0.07\,{\scriptsize[$-$0.10, $-$0.04]} & $-$0.01\,{\scriptsize[$-$0.03, 0.00]} & $-$0.03\,{\scriptsize[$-$0.06, $-$0.02]} & $-$0.40\,{\scriptsize[$-$0.47, $-$0.32]} \\
\quad Placebo: display $-$ no note & 0.08\,{\scriptsize[0.03, 0.13]} & 0.01\,{\scriptsize[$-$0.01, 0.04]} & 0.10\,{\scriptsize[0.05, 0.14]} & 0.23\,{\scriptsize[0.14, 0.33]} \\
\addlinespace[3pt]
\multicolumn{5}{@{}l}{\emph{\cond{Packed}}} \\
\quad No note, mean $\hat e$ & 1.68\,{\scriptsize[1.61, 1.76]} & 1.09\,{\scriptsize[1.06, 1.12]} & 1.35\,{\scriptsize[1.29, 1.41]} & 1.78\,{\scriptsize[1.70, 1.86]} \\
\quad Field guide, mean $\hat e$ & 1.56\,{\scriptsize[1.49, 1.63]} & 1.05\,{\scriptsize[1.03, 1.08]} & 1.29\,{\scriptsize[1.24, 1.35]} & 1.64\,{\scriptsize[1.56, 1.72]} \\
\quad Declaration: records, mean $\hat e$ & 1.28\,{\scriptsize[1.24, 1.33]} & 1.02\,{\scriptsize[1.01, 1.04]} & 1.11\,{\scriptsize[1.08, 1.14]} & 1.33\,{\scriptsize[1.27, 1.39]} \\
\quad Placebo: display, mean $\hat e$ & 1.68\,{\scriptsize[1.60, 1.75]} & 1.09\,{\scriptsize[1.06, 1.13]} & 1.40\,{\scriptsize[1.33, 1.47]} & 1.74\,{\scriptsize[1.66, 1.83]} \\
\quad Field guide $-$ no note & $-$0.12\,{\scriptsize[$-$0.22, $-$0.03]} & $-$0.04\,{\scriptsize[$-$0.08, 0.00]} & $-$0.06\,{\scriptsize[$-$0.13, 0.03]} & $-$0.14\,{\scriptsize[$-$0.24, $-$0.03]} \\
\quad Declaration: records $-$ no note & $-$0.40\,{\scriptsize[$-$0.48, $-$0.32]} & $-$0.06\,{\scriptsize[$-$0.10, $-$0.03]} & $-$0.24\,{\scriptsize[$-$0.31, $-$0.17]} & $-$0.45\,{\scriptsize[$-$0.54, $-$0.36]} \\
\quad Placebo: display $-$ no note & $-$0.01\,{\scriptsize[$-$0.10, 0.09]} & 0.01\,{\scriptsize[$-$0.04, 0.05]} & 0.05\,{\scriptsize[$-$0.03, 0.14]} & $-$0.03\,{\scriptsize[$-$0.14, 0.07]} \\
\bottomrule
\end{tabular}
\end{adjustbox}
\end{table*}
\begin{table*}[!tp]
\centering
\caption{Reasoning settings on the 500-item subset.}\label{app:tab:res-reasoning}
\small\setlength{\tabcolsep}{4pt}
\begin{adjustbox}{max width=\linewidth}
\begin{tabular}{@{}lcccc@{}}
\toprule
Reasoning & $w_3$ & $D_3$ & \cond{Independent}, $\hat e$ & \cond{Copied minority}, $\hat e>0$ (\%) \\
\midrule
\multicolumn{5}{@{}l}{\textbf{GPT-6 Luna}} \\
\quad off & 0.40\,{\scriptsize[0.37, 0.44]} & 0.40\,{\scriptsize[0.36, 0.45]} & 2.61\,{\scriptsize[2.57, 2.66]} & 31.4\,{\scriptsize[27.4, 35.4]} \\
\quad low (default) & 0.30\,{\scriptsize[0.27, 0.34]} & 0.61\,{\scriptsize[0.57, 0.65]} & 2.84\,{\scriptsize[2.81, 2.86]} & 21.0\,{\scriptsize[17.4, 24.6]} \\
\quad high & 0.08\,{\scriptsize[0.06, 0.10]} & 0.88\,{\scriptsize[0.86, 0.91]} & 2.93\,{\scriptsize[2.90, 2.95]} & 5.4\,{\scriptsize[3.4, 7.4]} \\
\addlinespace[3pt]
\multicolumn{5}{@{}l}{\textbf{GPT-6.1 Sol}} \\
\quad low (default) & 0.08\,{\scriptsize[0.06, 0.10]} & 0.90\,{\scriptsize[0.88, 0.92]} & 2.96\,{\scriptsize[2.95, 2.97]} & 4.2\,{\scriptsize[2.6, 6.0]} \\
\quad high & 0.01\,{\scriptsize[0.01, 0.02]} & 0.97\,{\scriptsize[0.96, 0.98]} & 2.97\,{\scriptsize[2.96, 2.97]} & 0.8\,{\scriptsize[0.2, 1.6]} \\
\addlinespace[3pt]
\multicolumn{5}{@{}l}{\textbf{Gemini 3.8 Flash}} \\
\quad low (default) & 0.23\,{\scriptsize[0.20, 0.26]} & 0.74\,{\scriptsize[0.71, 0.78]} & 2.94\,{\scriptsize[2.93, 2.95]} & 15.2\,{\scriptsize[12.2, 18.4]} \\
\quad high & 0.06\,{\scriptsize[0.04, 0.08]} & 0.91\,{\scriptsize[0.89, 0.93]} & 2.95\,{\scriptsize[2.93, 2.96]} & 4.4\,{\scriptsize[2.6, 6.2]} \\
\addlinespace[3pt]
\multicolumn{5}{@{}l}{\textbf{DeepSeek V4.1 Flash}} \\
\quad off & 0.55\,{\scriptsize[0.51, 0.59]} & 0.30\,{\scriptsize[0.25, 0.34]} & 2.69\,{\scriptsize[2.65, 2.73]} & 47.3\,{\scriptsize[43.0, 51.7]} \\
\quad low (default) & 0.40\,{\scriptsize[0.36, 0.44]} & 0.56\,{\scriptsize[0.52, 0.60]} & 2.92\,{\scriptsize[2.90, 2.94]} & 39.6\,{\scriptsize[35.4, 44.0]} \\
\quad high & 0.20\,{\scriptsize[0.16, 0.23]} & 0.78\,{\scriptsize[0.75, 0.81]} & 2.95\,{\scriptsize[2.94, 2.97]} & 13.2\,{\scriptsize[10.4, 16.2]} \\
\bottomrule
\end{tabular}
\end{adjustbox}
\end{table*}
\subsection{T2 Web evidence (Sections 4.5 and 5.1 of the main paper)}\label{app:sec:res-t2}

Table~\ref{app:tab:res-t2} gives the shift $\Delta$ of the answer toward the focal document, in log-odds, when two documents are added to the base pair, without and with the declaration. It also gives the weight $\rho$ of each copy type in units of a distinct document, and the change in shift that the declaration causes. Intervals resample the 299 questions behind the 500 items.

\begin{table*}[!tp]
\centering
\caption{Shift (log-odds) and weight $\rho$ of each added pair of documents in the web evidence.}\label{app:tab:res-t2}
\small\setlength{\tabcolsep}{4pt}
\begin{adjustbox}{max width=\linewidth}
\begin{tabular}{@{}lcccc@{}}
\toprule
Added documents & Luna & Sol & Gemini & DeepSeek \\
\midrule
\multicolumn{5}{@{}l}{\emph{Shift toward the focal document, without the declaration}} \\
\quad Duplicate & 0.22\,{\scriptsize[0.18, 0.25]} & 0.08\,{\scriptsize[0.06, 0.10]} & 0.13\,{\scriptsize[0.10, 0.17]} & 0.30\,{\scriptsize[0.25, 0.34]} \\
\quad Republished & 0.32\,{\scriptsize[0.29, 0.36]} & 0.15\,{\scriptsize[0.12, 0.18]} & 0.26\,{\scriptsize[0.21, 0.30]} & 0.45\,{\scriptsize[0.40, 0.51]} \\
\quad Summary & 0.46\,{\scriptsize[0.41, 0.50]} & 0.24\,{\scriptsize[0.20, 0.27]} & 0.41\,{\scriptsize[0.36, 0.45]} & 0.66\,{\scriptsize[0.61, 0.71]} \\
\quad Distinct & 0.71\,{\scriptsize[0.67, 0.76]} & 0.59\,{\scriptsize[0.56, 0.62]} & 0.82\,{\scriptsize[0.77, 0.86]} & 0.93\,{\scriptsize[0.87, 0.99]} \\
\addlinespace[3pt]
\multicolumn{5}{@{}l}{\emph{Weight $\rho$ in units of a distinct document, without the declaration}} \\
\quad Duplicate & 0.30\,{\scriptsize[0.25, 0.35]} & 0.14\,{\scriptsize[0.10, 0.18]} & 0.16\,{\scriptsize[0.13, 0.20]} & 0.32\,{\scriptsize[0.28, 0.36]} \\
\quad Republished & 0.45\,{\scriptsize[0.41, 0.50]} & 0.26\,{\scriptsize[0.21, 0.30]} & 0.31\,{\scriptsize[0.27, 0.36]} & 0.49\,{\scriptsize[0.44, 0.54]} \\
\quad Summary & 0.64\,{\scriptsize[0.59, 0.69]} & 0.40\,{\scriptsize[0.35, 0.46]} & 0.50\,{\scriptsize[0.45, 0.55]} & 0.71\,{\scriptsize[0.67, 0.75]} \\
\addlinespace[3pt]
\multicolumn{5}{@{}l}{\emph{Shift toward the focal document, with the declaration}} \\
\quad Duplicate & 0.06\,{\scriptsize[0.03, 0.09]} & 0.04\,{\scriptsize[0.02, 0.06]} & 0.07\,{\scriptsize[0.05, 0.10]} & 0.14\,{\scriptsize[0.10, 0.18]} \\
\quad Republished & 0.15\,{\scriptsize[0.11, 0.18]} & 0.05\,{\scriptsize[0.03, 0.07]} & 0.09\,{\scriptsize[0.06, 0.12]} & 0.25\,{\scriptsize[0.20, 0.29]} \\
\quad Summary & 0.24\,{\scriptsize[0.20, 0.28]} & 0.11\,{\scriptsize[0.08, 0.13]} & 0.19\,{\scriptsize[0.15, 0.22]} & 0.38\,{\scriptsize[0.33, 0.44]} \\
\quad Distinct & 0.66\,{\scriptsize[0.61, 0.70]} & 0.55\,{\scriptsize[0.52, 0.58]} & 0.74\,{\scriptsize[0.70, 0.78]} & 0.81\,{\scriptsize[0.75, 0.86]} \\
\addlinespace[3pt]
\multicolumn{5}{@{}l}{\emph{Weight $\rho$ in units of a distinct document, with the declaration}} \\
\quad Duplicate & 0.10\,{\scriptsize[0.05, 0.14]} & 0.07\,{\scriptsize[0.03, 0.10]} & 0.10\,{\scriptsize[0.06, 0.13]} & 0.18\,{\scriptsize[0.13, 0.22]} \\
\quad Republished & 0.22\,{\scriptsize[0.17, 0.28]} & 0.09\,{\scriptsize[0.06, 0.13]} & 0.13\,{\scriptsize[0.09, 0.17]} & 0.30\,{\scriptsize[0.25, 0.36]} \\
\quad Summary & 0.37\,{\scriptsize[0.31, 0.43]} & 0.19\,{\scriptsize[0.15, 0.24]} & 0.25\,{\scriptsize[0.20, 0.30]} & 0.47\,{\scriptsize[0.41, 0.53]} \\
\addlinespace[3pt]
\multicolumn{5}{@{}l}{\emph{Change in shift caused by the declaration}} \\
\quad Duplicate & $-$0.15\,{\scriptsize[$-$0.20, $-$0.11]} & $-$0.04\,{\scriptsize[$-$0.07, $-$0.02]} & $-$0.06\,{\scriptsize[$-$0.10, $-$0.02]} & $-$0.16\,{\scriptsize[$-$0.21, $-$0.11]} \\
\quad Republished & $-$0.18\,{\scriptsize[$-$0.22, $-$0.13]} & $-$0.10\,{\scriptsize[$-$0.13, $-$0.07]} & $-$0.16\,{\scriptsize[$-$0.21, $-$0.11]} & $-$0.21\,{\scriptsize[$-$0.28, $-$0.14]} \\
\quad Summary & $-$0.21\,{\scriptsize[$-$0.26, $-$0.17]} & $-$0.13\,{\scriptsize[$-$0.16, $-$0.10]} & $-$0.22\,{\scriptsize[$-$0.27, $-$0.17]} & $-$0.28\,{\scriptsize[$-$0.34, $-$0.22]} \\
\quad Distinct & $-$0.06\,{\scriptsize[$-$0.10, $-$0.02]} & $-$0.04\,{\scriptsize[$-$0.07, $-$0.01]} & $-$0.07\,{\scriptsize[$-$0.11, $-$0.03]} & $-$0.12\,{\scriptsize[$-$0.18, $-$0.07]} \\
\bottomrule
\end{tabular}
\end{adjustbox}
\end{table*}
\subsection{T3 Agent teams (Sections 4.6 and 5.1 to 5.3 of the main paper)}\label{app:sec:res-t3}

Table~\ref{app:tab:res-t3-protocol} describes the transcripts of each protocol without and with the reference rule. The restating share is the share of messages by agents without a sensor that state the observer's reading. Tables~\ref{app:tab:res-t3-commit} and~\ref{app:tab:res-t3-d} give, per protocol and model, the early commitment with one observer under each aggregator arm, and $\hat e$ with one observer and the discrimination $D$ without and with the rule. Table~\ref{app:tab:res-t3-arms} pools the protocols by evidence structure, including the true list and the share of episodes in which extracting first recovers the true readings. Three-observer results use complete logs, in which every observer's reading is stated at least once. Only the relay chain without the rule produces incomplete logs (Table~\ref{app:tab:res-t3-protocol}), and Table~\ref{app:tab:res-t3-d} also gives $D$ with them included.

\begin{table*}[!tp]
\centering
\caption{Transcripts of the agent teams by protocol, without and with the reference rule.}\label{app:tab:res-t3-protocol}
\small\setlength{\tabcolsep}{4pt}
\begin{adjustbox}{max width=\linewidth}
\begin{tabular}{@{}lcccc@{}}
\toprule
& Relay & Board & Discussion & Hierarchy \\
\midrule
Messages, one observer (median) & 4 & 5 & 8 & 5 \\
Messages, three observers (median) & 4 & 7 & 8 & 5 \\
Restatement count $\mu_1$, median, no rule & 4 & 5 & 7 & 3 \\
Restatement count $\mu_1$, median, reference rule & 1 & 1 & 1 & 2 \\
Restatement count $\mu_1$, mean, no rule & 3.96\,{\scriptsize[3.95, 3.98]} & 4.45\,{\scriptsize[4.36, 4.54]} & 5.85\,{\scriptsize[5.65, 6.04]} & 2.97\,{\scriptsize[2.95, 2.98]} \\
Restatement count $\mu_1$, mean, reference rule & 1.18\,{\scriptsize[1.14, 1.21]} & 1.35\,{\scriptsize[1.31, 1.40]} & 1.10\,{\scriptsize[1.08, 1.13]} & 2.31\,{\scriptsize[2.25, 2.38]} \\
Restating share (\%), no rule & 98.8 & 82.6 & 64.8 & 49.1 \\
Restating share (\%), reference rule & 5.9 & 2.7 & 1.3 & 32.9 \\
Words per message, no rule & 17.6 & 17.8 & 18.3 & 18.7 \\
Words per message, reference rule & 14.4 & 12.2 & 11.8 & 17.3 \\
Incomplete three-observer logs (\%), no rule & 25.6 & 0.0 & 0.0 & 0.0 \\
Incomplete three-observer logs (\%), reference rule & 0.0 & 0.0 & 0.0 & 0.0 \\
\bottomrule
\end{tabular}
\end{adjustbox}
\end{table*}
{\footnotesize\setlength{\tabcolsep}{4pt}
\begin{table*}[!tp]
\centering
\caption{Early commitment $P(\hat e>2)$ (\%) with one observer by protocol, aggregator arm and reference rule.}\label{app:tab:res-t3-commit}
\small\setlength{\tabcolsep}{3pt}
\begin{adjustbox}{max width=\textwidth}
\begin{tabular}{@{}lcccc@{}}
\toprule
Condition & Luna & Sol & Gemini & DeepSeek \\
\midrule
\multicolumn{5}{@{}l}{\emph{Relay}} \\
\quad no note & 11.6\,{\scriptsize[8.8, 14.4]} & 2.4\,{\scriptsize[1.2, 3.8]} & 9.4\,{\scriptsize[7.0, 12.0]} & 24.9\,{\scriptsize[21.0, 28.7]} \\
\quad declaration & 2.4\,{\scriptsize[1.2, 3.8]} & 0.8\,{\scriptsize[0.2, 1.6]} & 2.2\,{\scriptsize[1.0, 3.6]} & 6.8\,{\scriptsize[4.8, 9.2]} \\
\quad extract first & 0.0 & 0.0 & 0.0 & 0.0 \\
\quad reference rule & 0.0 & 0.0 & 0.0 & 1.4\,{\scriptsize[0.4, 2.6]} \\
\addlinespace[3pt]
\multicolumn{5}{@{}l}{\emph{Board}} \\
\quad no note & 14.0\,{\scriptsize[11.0, 17.0]} & 3.2\,{\scriptsize[1.8, 4.8]} & 13.6\,{\scriptsize[10.6, 16.6]} & 23.4\,{\scriptsize[19.6, 27.2]} \\
\quad declaration & 3.6\,{\scriptsize[2.0, 5.2]} & 0.6\,{\scriptsize[0.0, 1.4]} & 2.4\,{\scriptsize[1.2, 3.8]} & 7.0\,{\scriptsize[4.8, 9.2]} \\
\quad extract first & 0.0 & 0.0 & 0.0 & 0.0 \\
\quad reference rule & 0.4\,{\scriptsize[0.0, 1.0]} & 0.0 & 0.0 & 1.4\,{\scriptsize[0.4, 2.6]} \\
\addlinespace[3pt]
\multicolumn{5}{@{}l}{\emph{Discussion}} \\
\quad no note & 12.6\,{\scriptsize[9.8, 15.6]} & 4.0\,{\scriptsize[2.4, 5.8]} & 11.2\,{\scriptsize[8.4, 14.0]} & 25.2\,{\scriptsize[21.4, 29.2]} \\
\quad declaration & 3.2\,{\scriptsize[1.8, 4.8]} & 1.0\,{\scriptsize[0.2, 2.0]} & 1.2\,{\scriptsize[0.4, 2.2]} & 8.4\,{\scriptsize[6.0, 11.0]} \\
\quad extract first & 0.0 & 0.0 & 0.0 & 0.0 \\
\quad reference rule & 0.4\,{\scriptsize[0.0, 1.0]} & 0.0 & 0.0 & 0.8\,{\scriptsize[0.2, 1.6]} \\
\addlinespace[3pt]
\multicolumn{5}{@{}l}{\emph{Hierarchy}} \\
\quad no note & 5.4\,{\scriptsize[3.6, 7.4]} & 1.4\,{\scriptsize[0.4, 2.6]} & 5.2\,{\scriptsize[3.4, 7.2]} & 12.2\,{\scriptsize[9.4, 15.2]} \\
\quad declaration & 1.0\,{\scriptsize[0.2, 2.0]} & 0.4\,{\scriptsize[0.0, 1.0]} & 0.2\,{\scriptsize[0.0, 0.6]} & 4.2\,{\scriptsize[2.6, 6.0]} \\
\quad extract first & 0.0 & 0.0 & 0.0 & 0.0 \\
\quad reference rule & 2.8\,{\scriptsize[1.4, 4.4]} & 1.4\,{\scriptsize[0.4, 2.4]} & 2.4\,{\scriptsize[1.2, 3.8]} & 6.6\,{\scriptsize[4.6, 8.8]} \\
\addlinespace[3pt]
\multicolumn{5}{@{}l}{\emph{All protocols}} \\
\quad no note & 10.9\,{\scriptsize[9.4, 12.4]} & 2.8\,{\scriptsize[2.0, 3.6]} & 9.8\,{\scriptsize[8.5, 11.3]} & 21.4\,{\scriptsize[19.5, 23.4]} \\
\quad declaration & 2.5\,{\scriptsize[1.8, 3.3]} & 0.7\,{\scriptsize[0.4, 1.1]} & 1.5\,{\scriptsize[1.0, 2.1]} & 6.6\,{\scriptsize[5.5, 7.8]} \\
\quad extract first & 0.0 & 0.0 & 0.0 & 0.0 \\
\quad reference rule & 0.9\,{\scriptsize[0.5, 1.4]} & 0.4\,{\scriptsize[0.1, 0.6]} & 0.6\,{\scriptsize[0.3, 0.9]} & 2.5\,{\scriptsize[1.8, 3.2]} \\
\bottomrule
\end{tabular}
\end{adjustbox}
\end{table*}
\par}
\begin{table*}[!tp]
\centering
\caption{Mean $\hat e$ with one observer and discrimination $D$ by protocol, without and with the reference rule.}\label{app:tab:res-t3-d}
\small\setlength{\tabcolsep}{4pt}
\begin{adjustbox}{max width=\linewidth}
\begin{tabular}{@{}lcccc@{}}
\toprule
& Luna & Sol & Gemini & DeepSeek \\
\midrule
\multicolumn{5}{@{}l}{\emph{Relay}} \\
\quad $\hat e$, one observer, no rule & 1.36\,{\scriptsize[1.28, 1.44]} & 1.07\,{\scriptsize[1.04, 1.11]} & 1.29\,{\scriptsize[1.22, 1.36]} & 1.75\,{\scriptsize[1.64, 1.86]} \\
\quad $\hat e$, one observer, reference rule & 0.98\,{\scriptsize[0.97, 1.00]} & 0.96\,{\scriptsize[0.94, 0.97]} & 0.95\,{\scriptsize[0.93, 0.98]} & 0.99\,{\scriptsize[0.97, 1.02]} \\
\quad $D$, no rule & 0.78\,{\scriptsize[0.73, 0.83]} & 0.95\,{\scriptsize[0.93, 0.97]} & 0.86\,{\scriptsize[0.82, 0.91]} & 0.71\,{\scriptsize[0.64, 0.77]} \\
\quad $D$, reference rule & 0.92\,{\scriptsize[0.90, 0.94]} & 0.99\,{\scriptsize[0.98, 1.00]} & 0.99\,{\scriptsize[0.98, 1.01]} & 0.95\,{\scriptsize[0.93, 0.97]} \\
\quad $D$, no rule, incomplete logs included & 0.67 & 0.82 & 0.74 & 0.60 \\
\addlinespace[3pt]
\multicolumn{5}{@{}l}{\emph{Board}} \\
\quad $\hat e$, one observer, no rule & 1.41\,{\scriptsize[1.32, 1.50]} & 1.11\,{\scriptsize[1.06, 1.15]} & 1.37\,{\scriptsize[1.29, 1.47]} & 1.78\,{\scriptsize[1.66, 1.90]} \\
\quad $\hat e$, one observer, reference rule & 0.99\,{\scriptsize[0.97, 1.00]} & 1.00\,{\scriptsize[0.99, 1.02]} & 1.00\,{\scriptsize[0.98, 1.02]} & 1.03\,{\scriptsize[1.01, 1.06]} \\
\quad $D$, no rule & 0.82\,{\scriptsize[0.76, 0.87]} & 0.98\,{\scriptsize[0.95, 1.02]} & 0.89\,{\scriptsize[0.84, 0.95]} & 0.80\,{\scriptsize[0.72, 0.87]} \\
\quad $D$, reference rule & 0.95\,{\scriptsize[0.92, 0.97]} & 0.98\,{\scriptsize[0.97, 0.99]} & 0.98\,{\scriptsize[0.97, 1.00]} & 0.97\,{\scriptsize[0.95, 0.99]} \\
\quad $D$, no rule, incomplete logs included & 0.82 & 0.98 & 0.89 & 0.80 \\
\addlinespace[3pt]
\multicolumn{5}{@{}l}{\emph{Discussion}} \\
\quad $\hat e$, one observer, no rule & 1.42\,{\scriptsize[1.32, 1.52]} & 1.14\,{\scriptsize[1.08, 1.21]} & 1.39\,{\scriptsize[1.29, 1.49]} & 1.92\,{\scriptsize[1.77, 2.07]} \\
\quad $\hat e$, one observer, reference rule & 0.98\,{\scriptsize[0.97, 1.00]} & 0.99\,{\scriptsize[0.98, 0.99]} & 0.98\,{\scriptsize[0.97, 1.00]} & 0.99\,{\scriptsize[0.97, 1.01]} \\
\quad $D$, no rule & 0.90\,{\scriptsize[0.83, 0.97]} & 0.95\,{\scriptsize[0.90, 0.99]} & 0.90\,{\scriptsize[0.84, 0.96]} & 0.94\,{\scriptsize[0.83, 1.05]} \\
\quad $D$, reference rule & 0.94\,{\scriptsize[0.92, 0.95]} & 0.98\,{\scriptsize[0.97, 0.99]} & 0.99\,{\scriptsize[0.98, 1.00]} & 0.97\,{\scriptsize[0.95, 0.99]} \\
\quad $D$, no rule, incomplete logs included & 0.90 & 0.95 & 0.90 & 0.94 \\
\addlinespace[3pt]
\multicolumn{5}{@{}l}{\emph{Hierarchy}} \\
\quad $\hat e$, one observer, no rule & 1.16\,{\scriptsize[1.12, 1.20]} & 1.03\,{\scriptsize[1.01, 1.06]} & 1.15\,{\scriptsize[1.11, 1.19]} & 1.37\,{\scriptsize[1.31, 1.43]} \\
\quad $\hat e$, one observer, reference rule & 1.10\,{\scriptsize[1.06, 1.13]} & 1.03\,{\scriptsize[1.01, 1.05]} & 1.08\,{\scriptsize[1.05, 1.12]} & 1.22\,{\scriptsize[1.17, 1.27]} \\
\quad $D$, no rule & 0.93\,{\scriptsize[0.90, 0.96]} & 0.98\,{\scriptsize[0.97, 1.00]} & 0.95\,{\scriptsize[0.92, 0.99]} & 0.95\,{\scriptsize[0.91, 1.00]} \\
\quad $D$, reference rule & 0.94\,{\scriptsize[0.91, 0.97]} & 0.98\,{\scriptsize[0.96, 0.99]} & 0.97\,{\scriptsize[0.95, 0.99]} & 0.98\,{\scriptsize[0.94, 1.03]} \\
\quad $D$, no rule, incomplete logs included & 0.93 & 0.98 & 0.95 & 0.95 \\
\bottomrule
\end{tabular}
\end{adjustbox}
\end{table*}
\begin{table*}[!tp]
\centering
\caption{Mean $\hat e$ by evidence structure, aggregator arm and reference rule, pooled over protocols.}\label{app:tab:res-t3-arms}
\small\setlength{\tabcolsep}{4pt}
\begin{adjustbox}{max width=\linewidth}
\begin{tabular}{@{}lcccc@{}}
\toprule
& Luna & Sol & Gemini & DeepSeek \\
\midrule
\multicolumn{5}{@{}l}{\emph{One observer ($e^\ast=1$), mean $\hat e$}} \\
\quad no note & 1.34\,{\scriptsize[1.29, 1.38]} & 1.09\,{\scriptsize[1.06, 1.11]} & 1.30\,{\scriptsize[1.26, 1.34]} & 1.70\,{\scriptsize[1.64, 1.76]} \\
\quad declaration & 1.04\,{\scriptsize[1.02, 1.07]} & 1.02\,{\scriptsize[1.00, 1.03]} & 1.03\,{\scriptsize[1.01, 1.05]} & 1.19\,{\scriptsize[1.16, 1.23]} \\
\quad extract first & 0.97\,{\scriptsize[0.96, 0.98]} & 0.99\,{\scriptsize[0.98, 0.99]} & 0.98\,{\scriptsize[0.98, 0.99]} & 0.95\,{\scriptsize[0.93, 0.96]} \\
\quad reference rule & 1.01\,{\scriptsize[1.00, 1.02]} & 0.99\,{\scriptsize[0.99, 1.00]} & 1.01\,{\scriptsize[0.99, 1.02]} & 1.06\,{\scriptsize[1.04, 1.07]} \\
\quad true list & 1.00\,{\scriptsize[1.00, 1.00]} & 1.00\,{\scriptsize[1.00, 1.00]} & 1.00\,{\scriptsize[1.00, 1.00]} & 1.00\,{\scriptsize[1.00, 1.00]} \\
\quad extraction success (\%) & 98.0\,{\scriptsize[97.4, 98.7]} & 99.2\,{\scriptsize[98.8, 99.6]} & 98.9\,{\scriptsize[98.4, 99.3]} & 96.5\,{\scriptsize[95.7, 97.4]} \\
\addlinespace[3pt]
\multicolumn{5}{@{}l}{\emph{Three observers ($e^\ast=3$), complete logs, mean $\hat e$}} \\
\quad no note & 3.06\,{\scriptsize[3.02, 3.10]} & 3.02\,{\scriptsize[3.00, 3.05]} & 3.11\,{\scriptsize[3.07, 3.15]} & 3.42\,{\scriptsize[3.35, 3.49]} \\
\quad declaration & 2.88\,{\scriptsize[2.86, 2.91]} & 2.95\,{\scriptsize[2.94, 2.96]} & 2.96\,{\scriptsize[2.95, 2.98]} & 3.09\,{\scriptsize[3.04, 3.15]} \\
\quad extract first & 2.85\,{\scriptsize[2.83, 2.87]} & 2.92\,{\scriptsize[2.91, 2.93]} & 2.91\,{\scriptsize[2.89, 2.92]} & 2.87\,{\scriptsize[2.85, 2.89]} \\
\quad reference rule & 2.88\,{\scriptsize[2.86, 2.90]} & 2.96\,{\scriptsize[2.95, 2.97]} & 2.97\,{\scriptsize[2.96, 2.98]} & 2.99\,{\scriptsize[2.97, 3.02]} \\
\quad true list & 2.90\,{\scriptsize[2.86, 2.92]} & 2.96\,{\scriptsize[2.95, 2.96]} & 2.95\,{\scriptsize[2.92, 2.96]} & 2.94\,{\scriptsize[2.91, 2.96]} \\
\quad extraction success (\%) & 94.8\,{\scriptsize[93.7, 95.9]} & 96.8\,{\scriptsize[96.0, 97.6]} & 96.0\,{\scriptsize[95.0, 96.9]} & 92.8\,{\scriptsize[91.4, 94.1]} \\
\bottomrule
\end{tabular}
\end{adjustbox}
\end{table*}

Table~\ref{app:tab:res-t3-mu} and Figure~\ref{app:fig:res-t3} break the one-observer episodes without a note down by restatement count. Table~\ref{app:tab:res-t3-slope} gives the slope of $\hat e$ on $\mu_1-1$ with protocol fixed effects for the worker model of the main runs (Luna with reasoning off) and for a second worker model (DeepSeek with reasoning off), and Table~\ref{app:tab:res-t3-w2} compares their restatement counts on the configurations that both ran.

\begin{table*}[!tp]
\centering
\caption{One-observer episodes without a note by restatement count, pooled over protocols.}\label{app:tab:res-t3-mu}
\small\setlength{\tabcolsep}{4pt}
\begin{adjustbox}{max width=\linewidth}
\begin{tabular}{@{}lcccc@{}}
\toprule
& Luna & Sol & Gemini & DeepSeek \\
\midrule
\multicolumn{5}{@{}l}{\emph{Mean $\hat e$}} \\
\quad $\mu_1=2$ ($n=162$) & 1.03\,{\scriptsize[0.99, 1.08]} & 1.00\,{\scriptsize[0.97, 1.03]} & 1.04\,{\scriptsize[1.00, 1.08]} & 1.09\,{\scriptsize[1.03, 1.14]} \\
\quad $\mu_1=3$ ($n=542$) & 1.17\,{\scriptsize[1.12, 1.21]} & 1.04\,{\scriptsize[1.01, 1.06]} & 1.15\,{\scriptsize[1.11, 1.19]} & 1.37\,{\scriptsize[1.31, 1.43]} \\
\quad $\mu_1=4$ ($n=561$) & 1.35\,{\scriptsize[1.27, 1.42]} & 1.08\,{\scriptsize[1.04, 1.11]} & 1.27\,{\scriptsize[1.21, 1.34]} & 1.74\,{\scriptsize[1.64, 1.85]} \\
\quad $\mu_1=5$ ($n=418$) & 1.46\,{\scriptsize[1.36, 1.56]} & 1.11\,{\scriptsize[1.06, 1.17]} & 1.46\,{\scriptsize[1.36, 1.57]} & 1.92\,{\scriptsize[1.78, 2.06]} \\
\quad $\mu_1=6+$ ($n=313$) & 1.61\,{\scriptsize[1.47, 1.77]} & 1.22\,{\scriptsize[1.12, 1.33]} & 1.54\,{\scriptsize[1.40, 1.69]} & 2.24\,{\scriptsize[2.03, 2.46]} \\
\addlinespace[3pt]
\multicolumn{5}{@{}l}{\emph{Early commitment $P(\hat e>2)$ (\%)}} \\
\quad $\mu_1=2$ ($n=162$) & 1.2\,{\scriptsize[0.0, 3.2]} & 0.6\,{\scriptsize[0.0, 2.0]} & 0.6\,{\scriptsize[0.0, 2.0]} & 0.6\,{\scriptsize[0.0, 2.0]} \\
\quad $\mu_1=3$ ($n=542$) & 5.5\,{\scriptsize[3.6, 7.5]} & 1.5\,{\scriptsize[0.5, 2.7]} & 5.5\,{\scriptsize[3.6, 7.6]} & 12.9\,{\scriptsize[10.2, 15.8]} \\
\quad $\mu_1=4$ ($n=561$) & 11.6\,{\scriptsize[9.0, 14.2]} & 2.7\,{\scriptsize[1.4, 4.1]} & 8.9\,{\scriptsize[6.6, 11.4]} & 24.6\,{\scriptsize[20.9, 28.4]} \\
\quad $\mu_1=5$ ($n=418$) & 14.8\,{\scriptsize[11.5, 18.4]} & 3.1\,{\scriptsize[1.6, 4.9]} & 16.3\,{\scriptsize[12.8, 19.9]} & 27.0\,{\scriptsize[22.8, 31.3]} \\
\quad $\mu_1=6+$ ($n=313$) & 18.9\,{\scriptsize[14.6, 23.2]} & 5.8\,{\scriptsize[3.3, 8.5]} & 15.3\,{\scriptsize[11.4, 19.4]} & 33.9\,{\scriptsize[28.6, 39.2]} \\
\bottomrule
\end{tabular}
\end{adjustbox}
\end{table*}
\begin{figure*}[!tp]
\centering
\suppplot{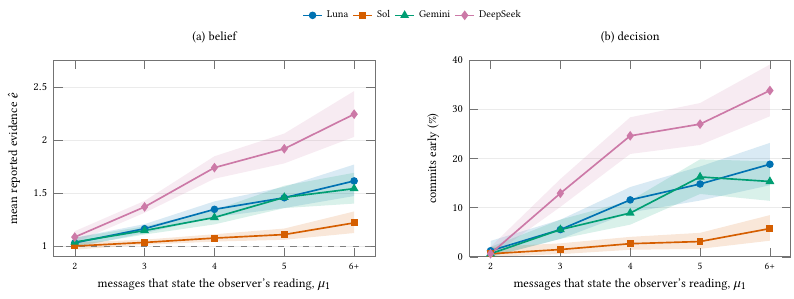}
\caption{Mean reported evidence (a) and early commitment (b) with one observer and no note against the restatement count, pooled over protocols, with 95\% intervals. The dashed line marks the oracle.}\label{app:fig:res-t3}
\Description{Two line charts against the number of messages that state the observer's reading, one line per model with a shaded interval band. Reported evidence and the share of episodes that would commit early both rise with the count.}
\end{figure*}
\begin{table*}[!tp]
\centering
\caption{Slope of $\hat e$ on the restatement count, with protocol fixed effects.}\label{app:tab:res-t3-slope}
\small\setlength{\tabcolsep}{4pt}
\begin{adjustbox}{max width=\linewidth}
\begin{tabular}{@{}lcccc@{}}
\toprule
Worker model & Luna & Sol & Gemini & DeepSeek \\
\midrule
Luna, reasoning off (500 configurations) & 0.121\,{\scriptsize[0.092, 0.151]} & 0.042\,{\scriptsize[0.022, 0.064]} & 0.113\,{\scriptsize[0.083, 0.145]} & 0.219\,{\scriptsize[0.178, 0.261]} \\
DeepSeek, reasoning off (100 configurations) & 0.100\,{\scriptsize[0.018, 0.184]} & 0.091\,{\scriptsize[0.033, 0.169]} & 0.158\,{\scriptsize[0.055, 0.277]} & 0.236\,{\scriptsize[0.091, 0.373]} \\
\bottomrule
\end{tabular}
\end{adjustbox}
\end{table*}
\begin{table*}[!tp]
\centering
\caption{Mean restatement count $\mu_1$ with one observer on the 100 configurations run with both worker models.}\label{app:tab:res-t3-w2}
\small\setlength{\tabcolsep}{4pt}
\begin{adjustbox}{max width=\linewidth}
\begin{tabular}{@{}lcc@{}}
\toprule
Protocol & Luna, reasoning off & DeepSeek, reasoning off \\
\midrule
Relay & 3.96\,{\scriptsize[3.92, 3.99]} & 3.97\,{\scriptsize[3.93, 4.00]} \\
Board & 4.54\,{\scriptsize[4.33, 4.73]} & 4.62\,{\scriptsize[4.45, 4.77]} \\
Discussion & 5.69\,{\scriptsize[5.23, 6.14]} & 6.65\,{\scriptsize[6.31, 6.97]} \\
Hierarchy & 2.98\,{\scriptsize[2.95, 3.00]} & 2.96\,{\scriptsize[2.92, 2.99]} \\
\bottomrule
\end{tabular}
\end{adjustbox}
\end{table*}
\subsection{Copy weight on one scale (Section 4.6 of the main paper)}\label{app:sec:res-synth}

Table~\ref{app:tab:res-synth} collects the three measures of the weight of a copy in units of an independent source: the copy weight $w_3$ on controlled logs, the slope of $\hat e$ on the restatement count in agent teams, and $\rho$ for each copy type in web evidence.

\begin{table*}[!tp]
\centering
\caption{Weight of a copy in units of an independent source, measured on the three testbeds.}\label{app:tab:res-synth}
\small\setlength{\tabcolsep}{4pt}
\begin{adjustbox}{max width=\linewidth}
\begin{tabular}{@{}lcccc@{}}
\toprule
& Luna & Sol & Gemini & DeepSeek \\
\midrule
T1 Controlled logs: copy weight $w_3$ & 0.27\,{\scriptsize[0.25, 0.30]} & 0.06\,{\scriptsize[0.05, 0.07]} & 0.21\,{\scriptsize[0.19, 0.24]} & 0.42\,{\scriptsize[0.39, 0.45]} \\
T3 Agent teams: slope on $\mu_1$ & 0.12\,{\scriptsize[0.09, 0.15]} & 0.04\,{\scriptsize[0.02, 0.06]} & 0.11\,{\scriptsize[0.08, 0.14]} & 0.22\,{\scriptsize[0.18, 0.26]} \\
T2 Web evidence: $\rho$, Duplicate & 0.30\,{\scriptsize[0.25, 0.35]} & 0.14\,{\scriptsize[0.10, 0.18]} & 0.16\,{\scriptsize[0.13, 0.20]} & 0.32\,{\scriptsize[0.28, 0.36]} \\
T2 Web evidence: $\rho$, Republished & 0.45\,{\scriptsize[0.41, 0.50]} & 0.26\,{\scriptsize[0.21, 0.30]} & 0.31\,{\scriptsize[0.27, 0.36]} & 0.49\,{\scriptsize[0.44, 0.54]} \\
T2 Web evidence: $\rho$, Summary & 0.64\,{\scriptsize[0.59, 0.69]} & 0.40\,{\scriptsize[0.35, 0.46]} & 0.50\,{\scriptsize[0.45, 0.55]} & 0.71\,{\scriptsize[0.67, 0.75]} \\
\bottomrule
\end{tabular}
\end{adjustbox}
\end{table*}
\subsection{Tests}\label{app:sec:res-tests}

Table~\ref{app:tab:res-tests} lists every pre-specified contrast with its estimate and its $p$ value after Holm correction within the family and model. Test W is the Wilcoxon signed-rank test on paired differences, or on differences from the oracle value, with zero differences dropped. M is the exact McNemar test on emitted commitment (a binomial test on the discordant pairs). P is Page's trend test, whose estimate is the change from \cond{Single} to $m=8$. B is the cluster bootstrap over configurations. Estimates are mean differences in readings, in log-odds (web evidence), in shares (actions) or in messages ($\mu_1$). In the agent teams the unit of a test is the configuration, and contrasts pooled over protocols use each configuration's mean over protocols. Table~\ref{app:tab:res-equiv} tests the null controls for equivalence with \cond{Single}: a control is equivalent if the 90\% paired bootstrap interval of the mean difference lies within $\pm0.10$ readings.

{\scriptsize\setlength{\tabcolsep}{3pt}
\begin{table*}[!tp]
\centering
\caption{Pre-specified tests with Holm-adjusted $p$ values.}\label{app:tab:res-tests}
\footnotesize\setlength{\tabcolsep}{3pt}
\begin{adjustbox}{max width=\textwidth}
\begin{tabular}{@{}>{\raggedright\arraybackslash}p{0.38\linewidth}ccccccccc@{}}
\toprule
Contrast & Test & \multicolumn{2}{c}{Luna} & \multicolumn{2}{c}{Sol} & \multicolumn{2}{c}{Gemini} & \multicolumn{2}{c}{DeepSeek} \\
\cmidrule(l{2pt}r{2pt}){3-4}\cmidrule(l{2pt}r{2pt}){5-6}\cmidrule(l{2pt}r{2pt}){7-8}\cmidrule(l{2pt}r{2pt}){9-10}
& & est. & $p$ & est. & $p$ & est. & $p$ & est. & $p$ \\
\midrule
\multicolumn{10}{@{}l}{\textbf{F1 Decomposition} (confirmatory)} \\
\quad \cond{Copies} $-$ \cond{Single} & W & 0.55 & $<$.001 & 0.12 & $<$.001 & 0.42 & $<$.001 & 0.85 & $<$.001 \\
\quad \cond{Packed} $-$ \cond{Trace} & W & 0.65 & $<$.001 & 0.09 & $<$.001 & 0.35 & $<$.001 & 0.75 & $<$.001 \\
\quad \cond{Pointers} $-$ \cond{Single} & W & 0.00 & .176 & 0.00 & 1.000 & 0.00 & .419 & 0.01 & .030 \\
\quad \cond{Padding} $-$ \cond{Single} & W & 0.00 & .576 & 0.00 & 1.000 & 0.00 & .600 & 0.01 & .212 \\
\quad \cond{Claims} $-$ \cond{Pointers} & W & 0.07 & $<$.001 & 0.01 & .020 & 0.03 & $<$.001 & 0.53 & $<$.001 \\
\quad \cond{Copies} $-$ \cond{Padding} & W & 0.55 & $<$.001 & 0.12 & $<$.001 & 0.43 & $<$.001 & 0.84 & $<$.001 \\
\quad \cond{Copies} $-$ \cond{Copies} (one relay) & W & $-$0.07 & .176 & 0.00 & 1.000 & 0.05 & .419 & 0.06 & .124 \\
\quad \cond{Independent} $-$ \cond{Copies} & W & 1.28 & $<$.001 & 1.84 & $<$.001 & 1.51 & $<$.001 & 1.07 & $<$.001 \\
\addlinespace[3pt]
\multicolumn{10}{@{}l}{\textbf{F2 Representation} (confirmatory)} \\
\quad \cond{Copies} $-$ \cond{Copies} (no labels) & W & $-$0.15 & $<$.001 & $-$0.08 & $<$.001 & $-$0.10 & .003 & $-$0.13 & .005 \\
\quad \cond{Copies} (prose) $-$ \cond{Copies} (prose, no labels) & W & $-$0.11 & .005 & $-$0.09 & $<$.001 & $-$0.08 & .048 & $-$0.15 & $<$.001 \\
\quad \cond{Copies} $-$ \cond{Copies} (prose) & W & 0.24 & $<$.001 & 0.06 & $<$.001 & 0.18 & $<$.001 & 0.30 & $<$.001 \\
\quad \cond{Copies} (no labels) $-$ \cond{Copies} (prose, no labels) & W & 0.28 & $<$.001 & 0.06 & .062 & 0.20 & $<$.001 & 0.27 & $<$.001 \\
\quad \cond{Independent} $-$ \cond{Independent} (no labels) & W & $-$0.01 & 1.000 & 0.00 & 1.000 & 0.00 & 1.000 & 0.01 & .517 \\
\quad \cond{Independent} (prose) $-$ \cond{Independent} (prose, no labels) & W & 0.01 & 1.000 & 0.00 & 1.000 & 0.00 & 1.000 & 0.00 & 1.000 \\
\quad \cond{Independent} $-$ \cond{Independent} (prose) & W & 0.00 & 1.000 & 0.01 & .656 & 0.01 & 1.000 & 0.02 & .743 \\
\quad \cond{Independent} (no labels) $-$ \cond{Independent} (prose, no labels) & W & 0.02 & 1.000 & 0.01 & 1.000 & 0.01 & 1.000 & 0.01 & 1.000 \\
\quad \cond{Single} $-$ \cond{Single} (prose) & W & 0.00 & 1.000 & 0.00 & 1.000 & 0.00 & .641 & $-$0.01 & .434 \\
\addlinespace[3pt]
\multicolumn{10}{@{}l}{\textbf{F3 Dose} (confirmatory)} \\
\quad \cond{Single} to \cond{Copies}, 8 statements, trend & P & 1.03 & $<$.001 & 0.30 & $<$.001 & 0.93 & $<$.001 & 1.53 & $<$.001 \\
\quad \cond{Single} to \cond{Independent}, $m=8$, trend & P & 4.73 & $<$.001 & 5.74 & $<$.001 & 5.58 & $<$.001 & 5.53 & $<$.001 \\
\quad \cond{Packed}, $m=8$ $-$ \cond{Packed} & W & 0.67 & $<$.001 & 0.14 & $<$.001 & 0.47 & $<$.001 & 0.71 & $<$.001 \\
\quad \cond{Copies}, 8 statements $-$ \cond{Padding}, $m=8$ & W & 1.02 & $<$.001 & 0.30 & $<$.001 & 0.93 & $<$.001 & 1.52 & $<$.001 \\
\quad \cond{Padding}, $m=8$ $-$ \cond{Single} & W & 0.00 & .233 & 0.00 & .317 & 0.00 & .04995 & 0.01 & .013 \\
\addlinespace[3pt]
\multicolumn{10}{@{}l}{\textbf{F4 Conflict} (confirmatory)} \\
\quad \cond{Copied tie} $-$ \cond{Tie} & W & 0.51 & $<$.001 & 0.11 & $<$.001 & 0.35 & $<$.001 & 0.86 & $<$.001 \\
\quad \cond{Copied tie} against 0 & W & 0.50 & $<$.001 & 0.11 & $<$.001 & 0.35 & $<$.001 & 0.86 & $<$.001 \\
\quad \cond{Copied minority} against $-$1 & W & 0.48 & $<$.001 & 0.10 & .003 & 0.36 & $<$.001 & 0.87 & $<$.001 \\
\addlinespace[3pt]
\multicolumn{10}{@{}l}{\textbf{F5 Notes} (confirmatory)} \\
\quad Field guide $-$ no note, \cond{Copies} & W & $-$0.17 & .002 & $-$0.07 & .125 & $-$0.12 & .144 & $-$0.07 & 1.000 \\
\quad Field guide $-$ no note, \cond{Independent} & W & $-$0.01 & 1.000 & 0.00 & 1.000 & 0.00 & 1.000 & $-$0.02 & 1.000 \\
\bottomrule
\end{tabular}
\end{adjustbox}
\end{table*}

\begin{table*}[!tp]
\centering
\ContinuedFloat
\caption{Pre-specified tests with Holm-adjusted $p$ values. (continued).}
\footnotesize\setlength{\tabcolsep}{3pt}
\begin{adjustbox}{max width=\textwidth}
\begin{tabular}{@{}>{\raggedright\arraybackslash}p{0.38\linewidth}ccccccccc@{}}
\toprule
Contrast & Test & \multicolumn{2}{c}{Luna} & \multicolumn{2}{c}{Sol} & \multicolumn{2}{c}{Gemini} & \multicolumn{2}{c}{DeepSeek} \\
\cmidrule(l{2pt}r{2pt}){3-4}\cmidrule(l{2pt}r{2pt}){5-6}\cmidrule(l{2pt}r{2pt}){7-8}\cmidrule(l{2pt}r{2pt}){9-10}
& & est. & $p$ & est. & $p$ & est. & $p$ & est. & $p$ \\
\midrule
\multicolumn{10}{@{}l}{\textbf{F5 Notes} (confirmatory)} \\
\quad Field guide $-$ no note, \cond{Copied minority} & W & $-$0.10 & .425 & $-$0.06 & .333 & $-$0.06 & 1.000 & $-$0.17 & .021 \\
\quad Field guide $-$ no note, \cond{Single} & W & 0.00 & 1.000 & 0.00 & 1.000 & 0.00 & 1.000 & 0.00 & 1.000 \\
\quad Declaration: records $-$ no note, \cond{Copies} & W & $-$0.60 & $<$.001 & $-$0.15 & $<$.001 & $-$0.44 & $<$.001 & $-$0.65 & $<$.001 \\
\quad Declaration: records $-$ no note, \cond{Independent} & W & 0.01 & 1.000 & $-$0.01 & 1.000 & 0.00 & 1.000 & $-$0.06 & .036 \\
\quad Declaration: records $-$ no note, \cond{Copied minority} & W & $-$0.47 & $<$.001 & $-$0.11 & $<$.001 & $-$0.33 & $<$.001 & $-$0.77 & $<$.001 \\
\quad Declaration: records $-$ no note, \cond{Single} & W & 0.00 & 1.000 & 0.00 & 1.000 & 0.00 & 1.000 & 0.00 & 1.000 \\
\quad Declaration: relays $-$ no note, \cond{Copies} & W & $-$0.56 & $<$.001 & $-$0.15 & $<$.001 & $-$0.44 & $<$.001 & $-$0.64 & $<$.001 \\
\quad Declaration: relays $-$ no note, \cond{Independent} & W & 0.01 & 1.000 & $-$0.01 & 1.000 & $-$0.01 & 1.000 & $-$0.02 & 1.000 \\
\quad Declaration: relays $-$ no note, \cond{Copied minority} & W & $-$0.46 & $<$.001 & $-$0.11 & $<$.001 & $-$0.34 & $<$.001 & $-$0.71 & $<$.001 \\
\quad Declaration: relays $-$ no note, \cond{Single} & W & 0.00 & 1.000 & 0.00 & 1.000 & 0.00 & 1.000 & 0.01 & 1.000 \\
\quad Placebo: display $-$ no note, \cond{Copies} & W & 0.28 & $<$.001 & $-$0.02 & 1.000 & 0.39 & $<$.001 & 0.27 & $<$.001 \\
\quad Placebo: display $-$ no note, \cond{Independent} & W & 0.01 & 1.000 & $-$0.01 & 1.000 & 0.01 & 1.000 & $-$0.03 & 1.000 \\
\quad Placebo: display $-$ no note, \cond{Copied minority} & W & 0.38 & $<$.001 & 0.05 & 1.000 & 0.40 & $<$.001 & 0.16 & .136 \\
\quad Placebo: display $-$ no note, \cond{Single} & W & $-$0.01 & 1.000 & 0.00 & 1.000 & 0.00 & 1.000 & 0.01 & 1.000 \\
\quad Placebo: delivery $-$ no note, \cond{Copies} & W & 0.13 & .299 & 0.00 & 1.000 & 0.18 & .011 & 0.13 & .287 \\
\quad Placebo: delivery $-$ no note, \cond{Independent} & W & $-$0.01 & 1.000 & $-$0.01 & 1.000 & 0.00 & 1.000 & $-$0.04 & 1.000 \\
\quad Placebo: delivery $-$ no note, \cond{Copied minority} & W & 0.21 & $<$.001 & 0.02 & 1.000 & 0.21 & $<$.001 & 0.00 & 1.000 \\
\quad Placebo: delivery $-$ no note, \cond{Single} & W & 0.00 & 1.000 & 0.00 & 1.000 & 0.00 & 1.000 & 0.01 & 1.000 \\
\quad Placebo: unrelated $-$ no note, \cond{Copies} & W & $-$0.05 & 1.000 & $-$0.03 & 1.000 & $-$0.05 & 1.000 & 0.01 & 1.000 \\
\quad Placebo: unrelated $-$ no note, \cond{Independent} & W & $-$0.01 & 1.000 & 0.00 & 1.000 & 0.00 & 1.000 & $-$0.02 & 1.000 \\
\quad Placebo: unrelated $-$ no note, \cond{Copied minority} & W & $-$0.01 & 1.000 & $-$0.02 & 1.000 & 0.03 & 1.000 & $-$0.11 & .656 \\
\quad Placebo: unrelated $-$ no note, \cond{Single} & W & $-$0.01 & 1.000 & 0.00 & 1.000 & 0.00 & 1.000 & 0.00 & 1.000 \\
\quad Counting instruction $-$ no note, \cond{Copies} & W & $-$0.46 & $<$.001 & $-$0.13 & $<$.001 & $-$0.36 & $<$.001 & $-$0.45 & $<$.001 \\
\quad Counting instruction $-$ no note, \cond{Independent} & W & 0.03 & .122 & $-$0.01 & 1.000 & 0.01 & 1.000 & $-$0.01 & 1.000 \\
\quad Counting instruction $-$ no note, \cond{Copied minority} & W & $-$0.34 & $<$.001 & $-$0.09 & $<$.001 & $-$0.29 & $<$.001 & $-$0.57 & $<$.001 \\
\quad Counting instruction $-$ no note, \cond{Single} & W & 0.00 & 1.000 & 0.00 & 1.000 & 0.00 & 1.000 & 0.01 & 1.000 \\
\quad Placebo: display $-$ Placebo: unrelated, \cond{Copies} & W & 0.33 & $<$.001 & 0.01 & 1.000 & 0.43 & $<$.001 & 0.26 & $<$.001 \\
\bottomrule
\end{tabular}
\end{adjustbox}
\end{table*}

\begin{table*}[!tp]
\centering
\ContinuedFloat
\caption{Pre-specified tests with Holm-adjusted $p$ values. (continued).}
\footnotesize\setlength{\tabcolsep}{3pt}
\begin{adjustbox}{max width=\textwidth}
\begin{tabular}{@{}>{\raggedright\arraybackslash}p{0.38\linewidth}ccccccccc@{}}
\toprule
Contrast & Test & \multicolumn{2}{c}{Luna} & \multicolumn{2}{c}{Sol} & \multicolumn{2}{c}{Gemini} & \multicolumn{2}{c}{DeepSeek} \\
\cmidrule(l{2pt}r{2pt}){3-4}\cmidrule(l{2pt}r{2pt}){5-6}\cmidrule(l{2pt}r{2pt}){7-8}\cmidrule(l{2pt}r{2pt}){9-10}
& & est. & $p$ & est. & $p$ & est. & $p$ & est. & $p$ \\
\midrule
\multicolumn{10}{@{}l}{\textbf{F5 Notes} (confirmatory)} \\
\quad Placebo: delivery $-$ Placebo: unrelated, \cond{Copies} & W & 0.19 & .001 & 0.03 & 1.000 & 0.22 & $<$.001 & 0.12 & .889 \\
\quad Declaration: records $-$ Declaration: relays, \cond{Copies} & W & $-$0.04 & .052 & 0.00 & 1.000 & 0.00 & 1.000 & $-$0.02 & 1.000 \\
\addlinespace[3pt]
\multicolumn{10}{@{}l}{\textbf{F6 Actions} (secondary)} \\
\quad Commit: \cond{Copies} $-$ \cond{Single}, no note & M & 0.22 & $<$.001 & 0.06 & $<$.001 & 0.24 & $<$.001 & 0.35 & $<$.001 \\
\quad Commit: Declaration: records $-$ no note, \cond{Copies} & M & $-$0.22 & $<$.001 & $-$0.06 & $<$.001 & $-$0.24 & $<$.001 & $-$0.25 & $<$.001 \\
\quad Commit: Declaration: records $-$ no note, \cond{Independent} & M & $-$0.01 & 1.000 & 0.00 & 1.000 & $-$0.01 & .688 & 0.00 & 1.000 \\
\quad Commit: Declaration: records $-$ no note, \cond{Copied minority} & M & 0.01 & 1.000 & 0.01 & .754 & 0.01 & .821 & $-$0.01 & 1.000 \\
\addlinespace[3pt]
\multicolumn{10}{@{}l}{\textbf{F7 Reasoning} (secondary)} \\
\quad Reasoning off $-$ low, \cond{Copies} & W & 0.20 & $<$.001 & -- & -- & -- & -- & 0.29 & $<$.001 \\
\quad Reasoning off $-$ low, \cond{Independent} & W & $-$0.22 & $<$.001 & -- & -- & -- & -- & $-$0.23 & $<$.001 \\
\quad Reasoning off $-$ low, \cond{Copied minority} & W & 0.29 & $<$.001 & -- & -- & -- & -- & 0.18 & .001 \\
\quad Reasoning high $-$ low, \cond{Copies} & W & $-$0.45 & $<$.001 & $-$0.12 & $<$.001 & $-$0.34 & $<$.001 & $-$0.41 & $<$.001 \\
\quad Reasoning high $-$ low, \cond{Independent} & W & 0.09 & $<$.001 & 0.01 & .061 & 0.01 & .315 & 0.03 & .035 \\
\quad Reasoning high $-$ low, \cond{Copied minority} & W & $-$0.37 & $<$.001 & $-$0.09 & $<$.001 & $-$0.24 & $<$.001 & $-$0.57 & $<$.001 \\
\addlinespace[3pt]
\multicolumn{10}{@{}l}{\textbf{T2-F1 Copies against no change} (confirmatory)} \\
\quad Duplicate against 0 & W & 0.22 & $<$.001 & 0.08 & $<$.001 & 0.13 & $<$.001 & 0.30 & $<$.001 \\
\quad Republished against 0 & W & 0.32 & $<$.001 & 0.15 & $<$.001 & 0.26 & $<$.001 & 0.45 & $<$.001 \\
\quad Summary against 0 & W & 0.46 & $<$.001 & 0.24 & $<$.001 & 0.41 & $<$.001 & 0.66 & $<$.001 \\
\addlinespace[3pt]
\multicolumn{10}{@{}l}{\textbf{T2-F2 Copies against distinct documents} (secondary)} \\
\quad Duplicate $-$ Distinct & W & $-$0.50 & $<$.001 & $-$0.51 & $<$.001 & $-$0.68 & $<$.001 & $-$0.63 & $<$.001 \\
\quad Republished $-$ Distinct & W & $-$0.39 & $<$.001 & $-$0.44 & $<$.001 & $-$0.56 & $<$.001 & $-$0.48 & $<$.001 \\
\quad Summary $-$ Distinct & W & $-$0.26 & $<$.001 & $-$0.35 & $<$.001 & $-$0.41 & $<$.001 & $-$0.27 & $<$.001 \\
\addlinespace[3pt]
\multicolumn{10}{@{}l}{\textbf{T2-F3 Declaration effect} (confirmatory)} \\
\quad Declaration, Duplicate & W & $-$0.15 & $<$.001 & $-$0.04 & .011 & $-$0.06 & .009 & $-$0.16 & $<$.001 \\
\quad Declaration, Republished & W & $-$0.18 & $<$.001 & $-$0.10 & $<$.001 & $-$0.16 & $<$.001 & $-$0.21 & $<$.001 \\
\quad Declaration, Summary & W & $-$0.21 & $<$.001 & $-$0.13 & $<$.001 & $-$0.22 & $<$.001 & $-$0.28 & $<$.001 \\
\quad Declaration, Distinct & W & $-$0.06 & .029 & $-$0.04 & .012 & $-$0.07 & .002 & $-$0.12 & $<$.001 \\
\addlinespace[3pt]
\multicolumn{10}{@{}l}{\textbf{T3-F1 One observer against the oracle} (confirmatory)} \\
\quad Relay: $\hat e$ against 1, one observer & W & 0.36 & $<$.001 & 0.07 & $<$.001 & 0.29 & $<$.001 & 0.75 & $<$.001 \\
\quad Board: $\hat e$ against 1, one observer & W & 0.41 & $<$.001 & 0.11 & $<$.001 & 0.37 & $<$.001 & 0.78 & $<$.001 \\
\quad Discussion: $\hat e$ against 1, one observer & W & 0.42 & $<$.001 & 0.14 & $<$.001 & 0.39 & $<$.001 & 0.92 & $<$.001 \\
\quad Hierarchy: $\hat e$ against 1, one observer & W & 0.16 & $<$.001 & 0.03 & $<$.001 & 0.15 & $<$.001 & 0.37 & $<$.001 \\
\addlinespace[3pt]
\multicolumn{10}{@{}l}{\textbf{T3-F2 Discrimination against the oracle} (secondary)} \\
\quad Relay: $D$ against 1 & W & $-$0.21 & $<$.001 & $-$0.05 & $<$.001 & $-$0.13 & $<$.001 & $-$0.30 & $<$.001 \\
\bottomrule
\end{tabular}
\end{adjustbox}
\end{table*}

\begin{table*}[!tp]
\centering
\ContinuedFloat
\caption{Pre-specified tests with Holm-adjusted $p$ values. (continued).}
\footnotesize\setlength{\tabcolsep}{3pt}
\begin{adjustbox}{max width=\textwidth}
\begin{tabular}{@{}>{\raggedright\arraybackslash}p{0.38\linewidth}ccccccccc@{}}
\toprule
Contrast & Test & \multicolumn{2}{c}{Luna} & \multicolumn{2}{c}{Sol} & \multicolumn{2}{c}{Gemini} & \multicolumn{2}{c}{DeepSeek} \\
\cmidrule(l{2pt}r{2pt}){3-4}\cmidrule(l{2pt}r{2pt}){5-6}\cmidrule(l{2pt}r{2pt}){7-8}\cmidrule(l{2pt}r{2pt}){9-10}
& & est. & $p$ & est. & $p$ & est. & $p$ & est. & $p$ \\
\midrule
\multicolumn{10}{@{}l}{\textbf{T3-F2 Discrimination against the oracle} (secondary)} \\
\quad Board: $D$ against 1 & W & $-$0.18 & $<$.001 & $-$0.02 & $<$.001 & $-$0.11 & $<$.001 & $-$0.20 & $<$.001 \\
\quad Discussion: $D$ against 1 & W & $-$0.10 & $<$.001 & $-$0.05 & $<$.001 & $-$0.10 & $<$.001 & $-$0.06 & $<$.001 \\
\quad Hierarchy: $D$ against 1 & W & $-$0.07 & $<$.001 & $-$0.02 & $<$.001 & $-$0.05 & $<$.001 & $-$0.05 & $<$.001 \\
\addlinespace[3pt]
\multicolumn{10}{@{}l}{\textbf{T3-F3 Aggregator arms} (confirmatory)} \\
\quad Extract first $-$ no note, one observer & W & $-$0.36 & $<$.001 & $-$0.10 & $<$.001 & $-$0.32 & $<$.001 & $-$0.75 & $<$.001 \\
\quad Declaration $-$ no note, one observer & W & $-$0.29 & $<$.001 & $-$0.07 & $<$.001 & $-$0.27 & $<$.001 & $-$0.51 & $<$.001 \\
\quad Extract first $-$ no note, three observers & W & $-$0.21 & $<$.001 & $-$0.10 & $<$.001 & $-$0.20 & $<$.001 & $-$0.55 & $<$.001 \\
\quad Declaration $-$ no note, three observers & W & $-$0.18 & $<$.001 & $-$0.07 & $<$.001 & $-$0.15 & $<$.001 & $-$0.32 & $<$.001 \\
\addlinespace[3pt]
\multicolumn{10}{@{}l}{\textbf{T3-F4 Slope} (secondary)} \\
\quad Slope on $\mu_1$ against 0 & B & 0.12 & $<$.001 & 0.04 & $<$.001 & 0.11 & $<$.001 & 0.22 & $<$.001 \\
\addlinespace[3pt]
\multicolumn{10}{@{}l}{\textbf{T3-F5 Reference rule} (confirmatory)} \\
\quad Relay: $\mu_1$, rule $-$ no rule & W & $-$2.79 & $<$.001 & $-$2.79 & $<$.001 & $-$2.79 & $<$.001 & $-$2.79 & $<$.001 \\
\quad Relay: $\hat e$ with one observer, rule $-$ no rule & W & $-$0.38 & $<$.001 & $-$0.12 & $<$.001 & $-$0.33 & $<$.001 & $-$0.75 & $<$.001 \\
\quad Relay: $D$, rule $-$ no rule & W & 0.13 & $<$.001 & 0.05 & .012 & 0.12 & $<$.001 & 0.25 & $<$.001 \\
\quad Board: $\mu_1$, rule $-$ no rule & W & $-$3.10 & $<$.001 & $-$3.10 & $<$.001 & $-$3.10 & $<$.001 & $-$3.10 & $<$.001 \\
\quad Board: $\hat e$ with one observer, rule $-$ no rule & W & $-$0.42 & $<$.001 & $-$0.10 & $<$.001 & $-$0.37 & $<$.001 & $-$0.74 & $<$.001 \\
\quad Board: $D$, rule $-$ no rule & W & 0.13 & $<$.001 & $-$0.01 & 1.000 & 0.09 & .004 & 0.17 & $<$.001 \\
\quad Discussion: $\mu_1$, rule $-$ no rule & W & $-$4.74 & $<$.001 & $-$4.74 & $<$.001 & $-$4.74 & $<$.001 & $-$4.74 & $<$.001 \\
\quad Discussion: $\hat e$ with one observer, rule $-$ no rule & W & $-$0.44 & $<$.001 & $-$0.15 & $<$.001 & $-$0.40 & $<$.001 & $-$0.93 & $<$.001 \\
\quad Discussion: $D$, rule $-$ no rule & W & 0.03 & .328 & 0.04 & .160 & 0.09 & .021 & 0.03 & .173 \\
\quad Hierarchy: $\mu_1$, rule $-$ no rule & W & $-$0.65 & $<$.001 & $-$0.65 & $<$.001 & $-$0.65 & $<$.001 & $-$0.65 & $<$.001 \\
\quad Hierarchy: $\hat e$ with one observer, rule $-$ no rule & W & $-$0.06 & .096 & 0.00 & 1.000 & $-$0.07 & .019 & $-$0.15 & $<$.001 \\
\quad Hierarchy: $D$, rule $-$ no rule & W & 0.01 & .328 & $-$0.01 & 1.000 & 0.02 & .021 & 0.03 & .203 \\
\bottomrule
\end{tabular}
\end{adjustbox}
\end{table*}
\par}
\begin{table*}[!tp]
\centering
\caption{Equivalence of the null controls with \cond{Single}: mean paired difference with its 90\% interval.}\label{app:tab:res-equiv}
\small\setlength{\tabcolsep}{4pt}
\begin{adjustbox}{max width=\linewidth}
\begin{tabular}{@{}lcccccccc@{}}
\toprule
Control & \multicolumn{2}{c}{Luna} & \multicolumn{2}{c}{Sol} & \multicolumn{2}{c}{Gemini} & \multicolumn{2}{c}{DeepSeek} \\
\cmidrule(l{2pt}r{2pt}){2-3}\cmidrule(l{2pt}r{2pt}){4-5}\cmidrule(l{2pt}r{2pt}){6-7}\cmidrule(l{2pt}r{2pt}){8-9}
& difference & eq. & difference & eq. & difference & eq. & difference & eq. \\
\midrule
\cond{Pointers} $-$ \cond{Single} & 0.004\,{\scriptsize[$-$0.001, 0.010]} & yes & 0.000 & yes & 0.001\,{\scriptsize[0.000, 0.003]} & yes & 0.013\,{\scriptsize[0.004, 0.022]} & yes \\
\cond{Trace} $-$ \cond{Single} & 0.005\,{\scriptsize[0.001, 0.009]} & yes & 0.003\,{\scriptsize[0.001, 0.007]} & yes & 0.007\,{\scriptsize[0.002, 0.013]} & yes & 0.008\,{\scriptsize[0.001, 0.015]} & yes \\
\cond{Padding} $-$ \cond{Single} & $-$0.003\,{\scriptsize[$-$0.009, 0.002]} & yes & 0.000 & yes & $-$0.002\,{\scriptsize[$-$0.006, 0.001]} & yes & 0.005\,{\scriptsize[$-$0.002, 0.012]} & yes \\
\cond{Single} (prose) $-$ \cond{Single} & $-$0.002\,{\scriptsize[$-$0.006, 0.002]} & yes & 0.000\,{\scriptsize[0.000, 0.001]} & yes & 0.002\,{\scriptsize[0.000, 0.004]} & yes & 0.007\,{\scriptsize[0.001, 0.013]} & yes \\
\cond{Padding}, $m=8$ $-$ \cond{Single} & 0.002\,{\scriptsize[$-$0.003, 0.006]} & yes & 0.001\,{\scriptsize[0.000, 0.002]} & yes & 0.003\,{\scriptsize[0.001, 0.005]} & yes & 0.009\,{\scriptsize[0.003, 0.017]} & yes \\
\bottomrule
\end{tabular}
\end{adjustbox}
\end{table*}
\FloatBarrier
\clearpage
\onecolumn
\section{Prompt templates and examples}\label{app:sec:promptcollection}
The following templates and examples give the complete system and user texts for T1, T2 and T3. Blue braces mark fields filled for each item or configuration.

\subsection{T1: System and user prompts}\label{app:sec:t1prompts}
System text. Paragraphs are separated by a blank line. The note paragraph is present only under an arm other than \arm{No note}.
\begin{Verbatim}
You are the aggregation component of \ph{team}. The team needs to know whether \ph{statement} (state \ph{POS}) or not (state \ph{NEG}). Before any sensor readings are taken, the two states are equally likely.

Each sensor takes exactly one reading, which is either \ph{POS} or \ph{NEG}. A sensor with reliability r reports the true state with probability r and the other state with probability 1 - r. Different sensors make their errors independently of one another.

\ph{Note on the message log: ... (arm text, Table \ref{app:tab:arms})}

You will receive the message log that the team's message bus delivered to you. Based on the log, report your probability that the true state is \ph{POS}.
\end{Verbatim}
For the action channel, the last paragraph reads \texttt{You will receive the message log that the team\textquotesingle{}s message bus delivered to you, followed by a decision that you must make.} For the policy control, it reads \texttt{You will be told your current probability that the true state is \{POS\}, followed by a decision that you must make.}

User text, belief channel (with a single record the header reads \texttt{1 message}):
\begin{Verbatim}
Message log (\ph{n} messages, in order of delivery):
\ph{one record per line}

Question: What is your probability that the true state is \ph{POS}?
Reply with a single line of the form PROBABILITY: <number>, where <number> is a decimal between 0 and 1.
\end{Verbatim}
User text, action channel: the same message log, a blank line, then the block below. For the policy control, the log is replaced by \texttt{Your current probability that the true state is \{POS\} is \{p:.3f\}.}
\begin{Verbatim}
Decision: choose exactly one of the following actions.
COMMIT \ph{POS}: declare that the true state is \ph{POS}.
COMMIT \ph{NEG}: declare that the true state is \ph{NEG}.
QUERY: pay a cost of 0.05 to obtain one additional reading from a new sensor with reliability \ph{r}, whose errors are independent of all other sensors; after seeing that reading, you must declare \ph{POS} or \ph{NEG}.
A correct declaration earns 1 point and an incorrect declaration earns -1 point. If you choose QUERY, its cost is subtracted from the points you earn afterwards. Choose the action with the highest expected number of points.
Reply with a single line of the form ACTION: <action>, where <action> is COMMIT \ph{POS}, COMMIT \ph{NEG} or QUERY.
\end{Verbatim}
The reliability $r$ of the new sensor is $0.717$, $0.800$, $0.870$, $0.925$ and $0.966$ for $q=0.60,\dots,0.80$. It is chosen so that the commit boundary $r-c/2$ equals the two-reading posterior up to rounding, which is $e_b\approx2$ in evidence units. The oracle therefore queries after one reading however often it is restated, and it commits after three agreeing independent readings.

The policy prompts take seven stated posteriors per reliability level:
\begin{itemize}
\item the one-reading posterior $q$,
\item the boundary shifted by $-0.10$, $-0.05$, $-0.02$, $+0.02$ and $+0.05$,
\item the three-reading posterior.
\end{itemize}
Each is mirrored for the negative polarity and used in all five domains, giving 350 prompts, each asked twice.

{\small
\begin{longtable}{@{}>{\raggedright\arraybackslash}p{0.2\linewidth}>{\raggedright\arraybackslash}p{0.76\linewidth}@{}}
\caption{Arm texts, inserted as the third paragraph of the system text. \arm{No note} has no such paragraph. \arm{Declaration: records}, \arm{Declaration: relays} and the three placebos have 47 to 49 tokens (o200k\_base). The placebos are true and irrelevant to the posterior, and they differ in how directly they point at the forwarded records. Block T1 notes uses the seven arms from \arm{Field guide} to \arm{Counting instruction}. \arm{Declaration: copies} and \arm{Declaration and instruction} are used only in T1 additional notes.}\label{app:tab:arms}\\
\toprule
Arm & Text \\
\midrule
\endfirsthead
\toprule
Arm & Text \\
\midrule
\endhead
No note & \emph{no third paragraph} \\
Field guide & \texttt{Note on the message log: Each line is one JSON record. report\_id identifies the record. sender is the agent or sensor that sent it. forwarded\_from is the report\_id of the record that this record was forwarded from, or null if the record was not forwarded. sensor\_id, reliability, reading\_id and measured\_at describe the reading that the record carries, and value is that reading.} \\
Declaration: records & \texttt{Note on the message log: Every record whose forwarded\_from field is not null is a lossless copy of the reading it carries. A forwarded record involves no new sensing and no independent verification, and it adds no private evidence of any kind.} \\
Declaration: relays & \texttt{Note on the message log: Relay agents have no sensors of their own. When a relay forwards a record, it passes on the same single reading unchanged; it does not measure again, does not check the value, and adds no information.} \\
Placebo: display & \texttt{Note on the message log: Every record whose forwarded\_from field is not null is rendered within the JSON records section. That section uses quoted field names, lists one record per line, and follows a fixed display order; it appears before the question.} \\
Placebo: delivery & \texttt{Note on the message log: Every record whose forwarded\_from field is not null was delivered over the team\textquotesingle{}s encrypted radio link. The link uses the standard channel, and the message bus logged each delivery at the moment it reached the base station.} \\
Placebo: unrelated & \texttt{Note on the message log: All times in the log are local station time in 24-hour format, and the log is exported once every hour. Your answer will be read by an automated parser, so please follow the required answer format exactly.} \\
Counting instruction & \texttt{Note on the message log: Count each distinct sensor reading exactly once, however many times it appears in the log.} \\
Declaration: copies & \texttt{Note on the message log: A forwarded record is not a second observation. It restates a reading that a sensor already reported, so it carries no new measurement, no independent check of that reading, and no extra evidence about the true state.} \\
Declaration and instruction & the text of \arm{Declaration: records} followed by one space and the sentence of \arm{Counting instruction} without its opening clause. \\
\bottomrule
\end{longtable}
}

\subsection{T1: Record templates}\label{app:sec:t1records}
Records are numbered \texttt{r1}, \texttt{r2}, \dots\ in display order. A JSON record is one line with the keys in the order shown:
\begin{Verbatim}
\{"report_id": "r1", "sender": "OD-34", "forwarded_from": null, "sensor_id": "OD-34", "reliability": 0.70, "reading_id": "rd-7852", "measured_at": "13:04", "value": "STORM"\}
\{"report_id": "r2", "sender": "Marten", "forwarded_from": "r1", "sensor_id": "OD-34", "reliability": 0.70, "reading_id": "rd-7852", "measured_at": "13:04", "value": "STORM"\}
\end{Verbatim}
The first is an original report, sent by the sensor itself. The second is a forward by a relay: it carries the same sensor, reading and time. Other record types differ from these as follows.
\begin{itemize}
\item \emph{Reference}: a forward whose \texttt{value} is \texttt{"see r1"}.
\item \emph{Equality claim}: a forward whose \texttt{value} is \texttt{"same as r1"}.
\item \emph{Packed record}: one record sent by the last relay, with \texttt{forwarded\_from} null and a trace of the earlier hops, for example \texttt{"relay\_trace": [\{"hop": 1, "by": "OD-34", "value": "STORM"\}, \{"hop": 2, "by": "Marten", "value": "STORM"\}]}. Hop 1 is the sensor and hop $k$ is relay $k-1$, so the value is stated once per hop plus once at the top. The \emph{packed reference} version omits \texttt{value} inside the hops.
\item \emph{Heartbeat}: \texttt{\{"report\_id", "sender", "forwarded\_from": null, "type": "heartbeat", "link": "ok", "battery\_pct", "queue\_depth", "uptime\_h", "firmware"\}}. Filler fields are added in the order \texttt{signal\_dbm}, \texttt{temp\_c}, \texttt{last\_sync}, \texttt{clock\_drift\_ms}, cycling over the heartbeats, until the user text is within 3 tokens (o200k\_base, as counted by \texttt{tiktoken}) of the matching forwarded log. If that is not enough, the last heartbeat gets a short \texttt{note} such as \texttt{"routine link check"}.
\item \emph{No labels}: the \texttt{forwarded\_from} key is dropped.
\item \emph{Prose}: \texttt{[r2] Marten: Forwarded from r1. Sensor OD-34 (reliability 0.70), reading rd-7852 at 13:04: STORM.} An original report begins \texttt{Original report.} instead. Without labels, this first sentence is dropped.
\end{itemize}

\subsection{T1: A complete episode}\label{app:sec:t1episode}
Item \texttt{T1-D1-70-P-01} (D1, $q=0.70$, primary reading \texttt{STORM}), form \cond{Copies}, arm \arm{No note}, belief channel. The system text is the template of Section~\ref{app:sec:t1prompts} filled for D1. The user text is:
\begin{Verbatim}
Message log (3 messages, in order of delivery):
\{"report_id": "r1", "sender": "OD-34", "forwarded_from": null, "sensor_id": "OD-34", "reliability": 0.70, "reading_id": "rd-7852", "measured_at": "13:04", "value": "STORM"\}
\{"report_id": "r2", "sender": "Marten", "forwarded_from": "r1", "sensor_id": "OD-34", "reliability": 0.70, "reading_id": "rd-7852", "measured_at": "13:04", "value": "STORM"\}
\{"report_id": "r3", "sender": "Stoat", "forwarded_from": "r1", "sensor_id": "OD-34", "reliability": 0.70, "reading_id": "rd-7852", "measured_at": "13:04", "value": "STORM"\}

Question: What is your probability that the true state is STORM?
Reply with a single line of the form PROBABILITY: <number>, where <number> is a decimal between 0 and 1.
\end{Verbatim}
A reply ending in \texttt{PROBABILITY: 0.78} gives $\hat p=0.78$. For this item $L=0.847$, so $\hat e=\logit(0.78)/0.847=1.49$ readings and the copy weight on this item is $(1.49-1)/2\approx0.25$. The oracle value is $e^\ast=1$ ($p^\ast=0.70$).
\subsection{T2: Evidence, paraphrase and stance-check prompts}\label{app:sec:t2prompts}
System text, with the second paragraph present only in the declaration conditions:
\begin{Verbatim}
You answer yes-or-no questions using only the documents you are given. The documents come from web search results, and each one shows its source.

Note on the documents: Some documents may be exact duplicates, republished copies or summaries of another document in the list. Such a copy repeats its original and is not an independent source.

Report your probability that the answer to the question is Yes.
\end{Verbatim}
User text:
\begin{Verbatim}
Question: \ph{question}

Documents (\ph{n}, in the order they were retrieved):

[Document 1]
Source: \ph{url}
Title: \ph{title}
\ph{attribution line, only for republished copies and summaries}
\ph{text}

[Document 2]
...

Based only on these documents, what is your probability that the answer to the question is Yes?
Reply with a single line of the form PROBABILITY: <number>, where <number> is a decimal between 0 and 1.
\end{Verbatim}
Paraphrase (system \texttt{You rewrite web documents for a research study. Follow the instructions exactly.}):
\begin{Verbatim}
Rewrite the document below as a news-style summary written by a different outlet. Keep every factual claim, number and conclusion, and keep the document's overall position on the question "\ph{question}". Do not add new information and do not mention that the text is a rewrite. Use roughly the same length as the original.

Document:
\ph{text of F}
\end{Verbatim}
Stance check (system \texttt{You classify documents for a research study. Follow the instructions exactly.}):
\begin{Verbatim}
Question: \ph{question}

Document:
\ph{summary}

Does this document argue that the answer to the question is Yes or No? Reply with a single line of the form ANSWER: YES, ANSWER: NO or ANSWER: NEITHER.
\end{Verbatim}
The stance reply is read from the last line matching the pattern below, case-insensitively.
\begin{Pattern}
ANSWER\s*:\s*(YES|NO|NEITHER)
\end{Pattern}
A summary whose stance differs from $F$'s is regenerated, up to two more times. If it still differs, the two \cond{Summary} conditions of that item are missing.
\subsection{T3: Worker prompts}\label{app:sec:t3worker}
\begin{Verbatim}
You are \ph{name}, a member of \ph{team}. The team needs to know whether \ph{statement} (state \ph{POS}) or not (state \ph{NEG}).

\ph{ROLE}

Messages you write are posted on the team's message bus, which also delivers them to the team's aggregation component. Keep each message short and factual (at most 60 words), and write plain text without lists or formatting.
\end{Verbatim}
Under the reference rule the system text ends with one more paragraph:
\begin{Verbatim}
Team rule: every message stays on the team's message bus, so each reading should be stated in full only once. When you mention a reading that already appears in a message you have seen, including your own earlier message, do not repeat its value and do not say whether you agree with it. Refer to it instead, in the form "see <agent>'s report on <sensor>".
\end{Verbatim}
The role sentence depends on the agent:
\begin{itemize}
\item observer: \texttt{You operate sensor \{sensor\_id\} (reliability \{q\}), which takes exactly one reading. Its reading \{reading\_id\}, taken at \{time\}, is \{VALUE\}.}
\item agent without a sensor: \texttt{You do not operate any sensor and cannot take readings yourself.}
\end{itemize}

The user text of a turn is one of the following. A quoted message is shown in double quotes.
\begin{itemize}
\item \texttt{You are the first agent in a relay chain. Write your message to \{next\}, the next agent in the chain.}
\item \texttt{You are agent \{pos\} of 4 in a relay chain. You received this message from \{prev\}:} followed on the next line by the quoted message, then a blank line and \texttt{Write your message to \{next\}, the next agent in the chain.} The last agent is told \texttt{You are the last agent in a relay chain.} and asked to \texttt{Write your message to the team\textquotesingle{}s aggregation component.}
\item \texttt{Team board so far:}, one line \texttt{\{sender\}: \{text\}} per post (or \texttt{(empty)}), a blank line, \texttt{Post an update to the team board.}
\item \texttt{Team discussion so far:}, the same line format (or \texttt{(no messages yet)}), a blank line, \texttt{It is your turn to speak. Write your message to the team.}
\item \texttt{Your sub-team lead is \{lead\}. Write your report to \{lead\}.}
\item \texttt{You lead a sub-team. Report from your sub-team member \{member\}:}, the quoted report, a blank line, \texttt{Write a summary of your sub-team\textquotesingle{}s information for the team\textquotesingle{}s aggregation component.}
\item \texttt{You are the overall team lead. Messages posted so far:}, one line \texttt{\{sender\} to \{to\}: \{text\}} per message, a blank line, \texttt{Write the final team summary for the team\textquotesingle{}s aggregation component.}
\end{itemize}
Before a reply is passed on or logged, runs of whitespace are collapsed to single spaces and one pair of enclosing quotes is removed. Nothing else is changed. An empty reply is requested again, up to three attempts. If a turn still fails, the episode restarts from turn 1, at most twice, and the number of restarts is recorded.

\subsection{T3: Aggregator prompts}\label{app:sec:t3agg}
The bus log is rendered as \texttt{Message log (\{n\} messages, in order of delivery):} (\texttt{1 message} for a single one) followed by one line \texttt{[m\{i\}] \{sender\} to \{to\}: \{text\}} per message. The recipient is an agent name, \texttt{team board}, \texttt{team} or \texttt{aggregator}.

The system text contains, in order:
\begin{enumerate}
\item the two world paragraphs of T1,
\item the roster,
\item the arm note, under \emph{declaration} only,
\item a closing paragraph.
\end{enumerate}
The roster with one observer reads: \texttt{The team has four agents. \{A1\} operates sensor \{S1\} (reliability \{q\}). \{A2\}, \{A3\} and \{A4\} do not operate any sensor and cannot take readings.} With three observers: \texttt{The team has four agents. \{A1\} operates sensor \{S1\} (reliability \{q\}), \{A2\} operates sensor \{S2\} (reliability \{q\}) and \{A3\} operates sensor \{S3\} (reliability \{q\}). \{A4\} does not operate any sensor and cannot take readings.}

Episodes without a rule are scored with \emph{no note}, \emph{declaration} and \emph{extract first} under each of the four low-effort settings. Episodes under the reference rule are scored with \emph{no note} only, since the rule changes the log and not the aggregator.
\begin{itemize}
\item \emph{No note}. The closing paragraph is \texttt{You will receive the log of the team\textquotesingle{}s message bus. Based on the log, report your probability that the true state is \{POS\}.} The user text is the log followed by the T1 probability question.
\item \emph{Declaration}. As \emph{no note}, with the note \texttt{Note on the message log: Only agents that operate a sensor can report a new reading. A message from an agent without a sensor can only repeat readings that others reported; it involves no new measurement and no independent verification.}
\item \emph{Extract first}, a two-step call to the same model.
  \begin{itemize}
  \item Step 1 closes the system text with \texttt{You will receive the log of the team\textquotesingle{}s message bus, followed by a task.} Its user text is the log followed by \texttt{Task: list every distinct sensor reading reported in this log. Write one line per sensor in the form READING: <sensor\_id> = <value>, where <value> is \{POS\} or \{NEG\}. List each sensor once, however many messages mention its reading. If the log reports no reading, write the single line READING: NONE.} The reply is read line by line after deleting \texttt{*} and \texttt{\textasciigrave}, upper-casing, and replacing \texttt{\_} and \texttt{-} by spaces. A line matching
\begin{Pattern}
READING\s*:\s*([A-Z]{2})\s*(\d{2})\s*=\s*(<POS>|<NEG>)
\end{Pattern}
  gives the sensor \texttt{XX-NN} and its value. Lines for sensors outside the roster are dropped, and a sensor listed twice keeps its first value. \texttt{READING: NONE} gives an empty list. A reply with neither a valid line nor \texttt{NONE} is requested again.
  \item Step 2 contains only the two world paragraphs and \texttt{You will receive the distinct sensor readings reported by the team. Based on them, report your probability that the true state is \{POS\}.} Its user text is \texttt{Distinct sensor readings reported by the team:}, one line \texttt{Sensor \{sensor\_id\} (reliability \{q\}): \{VALUE\}.} per parsed reading (or \texttt{(no readings reported)}), a blank line, and the T1 question.
  \end{itemize}
\item \emph{True list}. Step 2 given the true readings. It is run once per configuration and structure.
\end{itemize}

\subsection{T3: Annotation prompt}\label{app:sec:t3annotprompt}
Sol (setting \texttt{sol-high}) annotates each transcript once. The system text is \texttt{You annotate message logs for a research study. Follow the instructions exactly.} The user text is \texttt{Team roster: \{roster\}}, a blank line, the bus log, a blank line and:
\begin{Verbatim}
For each message, list the sensor readings whose value the message states. A message states a reading's value if the value appears in the message, or if the message asserts that the value is the same as a reading reported elsewhere (for example "same reading as \ph{A1} reported"). A message that only points to another report without asserting the value (for example "see \ph{A1}'s report on \ph{S1}") does not state it. Include the sender's own reading if the sender reports it. Use one line per message, in the form
m<k>: <sensor_id>=<value>; <sensor_id>=<value>
or, if the message states no reading,
m<k>: none
where <value> is \ph{POS} or \ph{NEG}. Identify each sensor by the identifier in the roster. At the end of a line add " | CLAIM" if the message says or implies that a reading was confirmed, verified or double-checked, or that more sources support it than the readings actually reported in this log. Write exactly one line per message, in order, and nothing else.
\end{Verbatim}

\Needspace{0.42\textheight}
\subsection{T3: A complete episode}\label{app:sec:t3episode}
Configuration \texttt{T3-D1-70-P-01} (agents Raven, Otter, Lynx, Marten; sensor OD-85; $q=0.70$; \texttt{STORM}), protocol Relay, one observer. The four worker turns produce, for example:
\begin{Verbatim}
[m1] Raven to Otter: Sensor OD-85 reading rd-9300 at 12:55: STORM. Passing this to Otter.
[m2] Otter to Lynx: Raven reports that sensor OD-85 read STORM at 12:55. Forwarding to Lynx.
[m3] Lynx to Marten: Update from the chain: sensor OD-85 indicates STORM (reading rd-9300). Please pass it on.
[m4] Marten to aggregator: Final relay: the team's sensor OD-85 reported STORM at 12:55, as passed along by Raven, Otter and Lynx.
\end{Verbatim}
The annotation marks all four messages as stating \texttt{OD-85=STORM}, so $\mu_1=4$. Each model then makes four aggregator calls on this log: \emph{no note}, \emph{declaration}, and the two steps of \emph{extract first}. Under the reference rule the same chain might read:
\begin{Verbatim}
[m1] Raven to Otter: Sensor OD-85 reading rd-9300 at 12:55: STORM. Passing this to Otter.
[m2] Otter to Lynx: See Raven's report on OD-85. Forwarding to Lynx.
[m3] Lynx to Marten: See Raven's report on OD-85; passing the reference on to Marten.
[m4] Marten to aggregator: For the aggregation component: see Raven's report on OD-85, relayed by Otter and Lynx.
\end{Verbatim}
Here only \texttt{m1} states the reading, so $\mu_1=1$.
\clearpage
\endgroup
\end{document}